\documentclass{article} % For LaTeX2e

\usepackage[dvipsnames]{xcolor}

\usepackage{iclr2024_conference,times}
\usepackage{amsmath,amsfonts,bm}

\def\figref#1{figure~\ref{#1}}
\def\secref#1{section~\ref{#1}}
\def\eqref#1{equation~\ref{#1}}
\def\algref#1{algorithm~\ref{#1}}
\def\Algref#1{Algorithm~\ref{#1}}

\def\1{\bm{1}}

\DeclareMathAlphabet{\mathsfit}{\encodingdefault}{\sfdefault}{m}{sl}
\SetMathAlphabet{\mathsfit}{bold}{\encodingdefault}{\sfdefault}{bx}{n}

\DeclareMathOperator*{\argmin}{arg\,min}

\def\algref#1{Algorithm~\ref{#1}}
\def\Algref#1{Algorithm~\ref{#1}}

\def\cB{{\mathcal{B}}}

\def\cD{{\mathcal{D}}}
\def\cE{{\mathcal{E}}}
\def\cF{{\mathcal{F}}}
\def\cG{{\mathcal{G}}}

\def\cM{{\mathcal{M}}}
\def\cN{{\mathcal{N}}}
\def\cO{{\mathcal{O}}}

\def\cS{{\mathcal{S}}}

\def\cW{{\mathcal{W}}}
\def\cX{{\mathcal{X}}}
\def\cY{{\mathcal{Y}}}
\def\cZ{{\mathcal{Z}}}

\def\sE{{\mathbb{E}}}

\newcommand{\defref}[1]{definition~\ref{#1}}
\newcommand{\tabref}[1]{table~\ref{#1}}

\newcommand{\thmref}[1]{theorem~\ref{#1}}

\newcommand{\propref}[1]{proposition~\ref{#1}}

\def\ellfg{{\ell^{\mathrm{fg}}}}
\def\ellcg{{\ell^{\mathrm{cg}}}}

\def\datafg{\mathcal{D}^{\mathrm{fg}}_n}
\def\datacg{\mathcal{D}^{\mathrm{cg}}_m}
\def\dataaug{\cD^{\mathrm{fg,aug}}_{n}}
\def\alg{\mathcal{A}}

\usepackage{hyperref}
\usepackage{url}

\usepackage{graphicx}
\usepackage{amsmath}
\usepackage{amssymb}
\usepackage{mathtools}
\usepackage{amsthm}
\usepackage{bbm}
\usepackage{multirow}
\usepackage{algorithm}
\usepackage{algorithmic}
\usepackage{multirow}
\usepackage{subfig}
\usepackage{xspace}
\usepackage{wrapfig}
\usepackage{hyperref}
\usepackage{xspace}
\usepackage{blindtext}

\newtheorem{prop}{Proposition}

\newtheorem{theorem}[prop]{Theorem}

\theoremstyle{definition}
\newtheorem{defn}[prop]{Definition}

\title{Enhancing Instance-Level Image Classification with Set-Level Labels}

\author{Renyu Zhang \\
Department of Computer Science\\
University of Chicago\\
\texttt{zhangr@uchicago.edu} \\
\And
Aly A. Khan \\
Department of Pathology and Family Medicine\\
University of Chicago \\
\texttt{aakhan@uchicago.edu} \\
\AND
Yuxin Chen \\
Department of Computer Science\\
University of Chicago\\
\texttt{chenyuxin@uchicago.edu} \\
\And
Robert L. Grossman \\
Department of Computer Science and Medicine\\
University of Chicago\\
\texttt{rgrossman1@uchicago.edu}\\
\AND
}

\newif\iffinal
\finaltrue

\iclrfinalcopy % Uncomment for camera-ready version, but NOT for submission.
\begin{document}

\maketitle

\begin{abstract}
Instance-level image classification tasks have traditionally relied on single-instance labels to train models, e.g., few-shot learning and transfer learning. However, set-level coarse-grained labels that capture relationships among instances can provide richer information in real-world scenarios. In this paper, we present a novel approach to enhance instance-level image classification by leveraging set-level labels. We provide a theoretical analysis of the proposed method, including recognition conditions for fast excess risk rate, shedding light on the theoretical foundations of our approach. We conducted experiments on two distinct categories of datasets: natural image datasets and histopathology image datasets. Our experimental results demonstrate the effectiveness of our approach, showcasing improved classification performance compared to traditional single-instance label-based methods. Notably, our algorithm achieves 13\% improvement in classification accuracy compared to the strongest baseline on the histopathology image classification benchmarks. Importantly, our experimental findings align with the theoretical analysis, reinforcing the robustness and reliability of our proposed method. This work bridges the gap between instance-level and set-level image classification, offering a promising avenue for advancing the capabilities of image classification models with set-level coarse-grained labels.
\end{abstract}

\section{Introduction}\label{sec_intro}

A large amount of labeled data is typically required in traditional machine learning approaches, e.g., few-shot learning (FSL) and transfer learning (TL), to learn a robust model. However, procuring sufficient labeled data for each task is often challenging or infeasible in real-world scenarios. In this paper, we consider a novel problem setting where similar to FSL, we have a limited number of fine-grained labels in the target domain but in the source domain, we have a large amount of coarse-grained set-level labels which are easier to obtain and relevant to fine-grained labels. %\YC{can bring up a few real-world applications here to show practical relevance.} 
For example, in a digital library, there are coarse-grained set-level labels indicating the general content of photo albums, such as ``beach vacation'', ``nature landscapes'', or ``picnic''. However, within each of these albums, there are numerous individual images, each with its own unique details and characteristics that are not explicitly labeled. In the downstream task, for instance, we care about the object classification such as ``tree'', ``beach'', or ``mountain''. Similarly, in the medical domain, it is often useful to predict fine-grained labels of tissues, while only set-level annotations of histopathology slides are available for training at scale. We seek to enhance the downstream classification tasks with the coarse-grained set-level labels.

An effective approach to addressing the overreliance on abundant training data is FSL---a paradigm that has gained significant attention in recent years %, with the goal to adapt models to new tasks when labeled data is limited 
\citep{vinyals2016matching, wang2016learning, triantafillou2017few, finn2017model, snell2017prototypical, sung2018learning, wang2018low, oreshkin2018tadam, rusu2018meta, ye2018learning, lee2019meta, li2019finding}. %Traditional machine learning methods often require a substantial amount of labeled data to train a robust model. However, collecting a sufficient amount of labeled data in digital pathology can be challenging or even impractical in some scenarios. 
FSL pretrains a model that can quickly adapt to new tasks using only a few labeled examples. Recent studies \citep{chen2019closer, tian2020rethinking, shakeri2022fhist, yang2022towards} have shown that pretraining, coupled with finetuning on a new task, outperforms more sophisticated episodic training methods. This involves initially training a base model on a diverse set of tasks using abundant labeled data from a source domain, and subsequently finetuning the model using only a small number of labeled examples specific to the target task. Despite their promising performance, existing FSL models typically depend on finely labeled source data for predicting fine-grained labels. 

\begin{figure*}[t]%
\centering
  \subfloat[]{\includegraphics[width=0.6\textwidth]{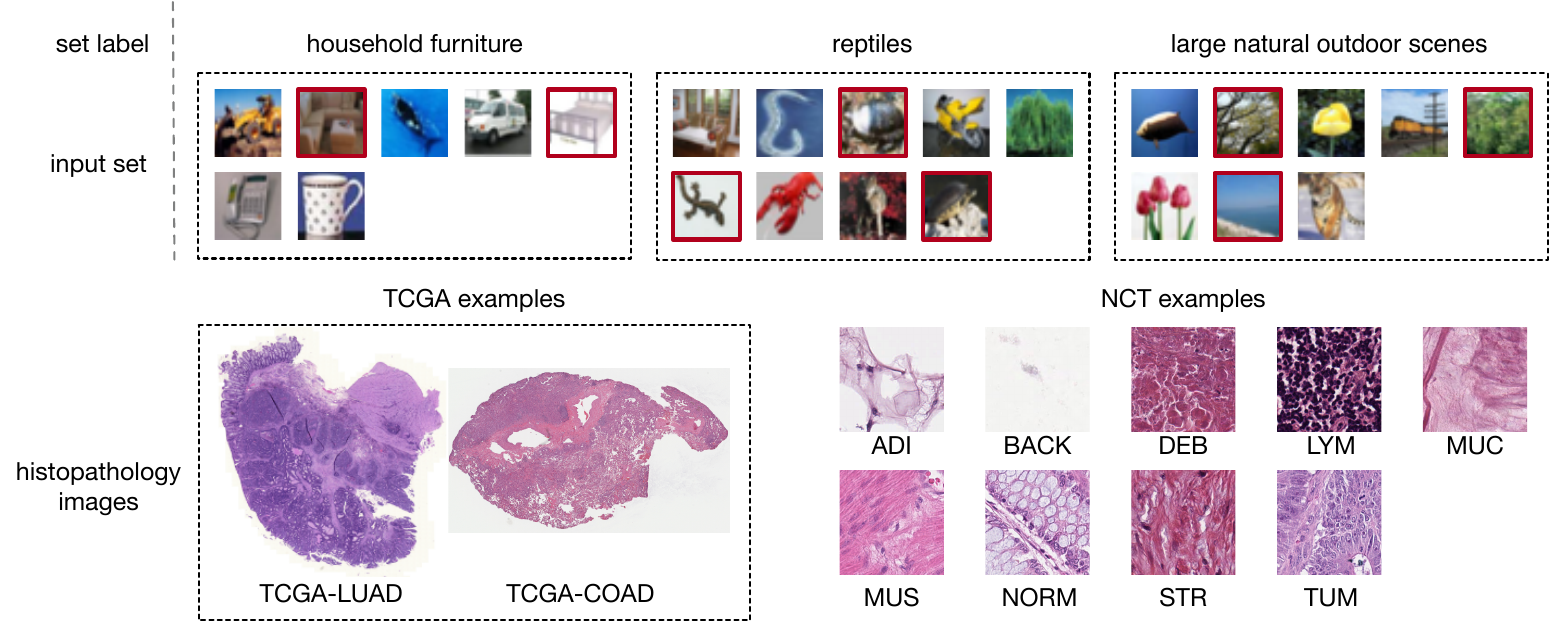}\label{fig:wsi_tile_example}}
  \hfill
  \subfloat[]{\includegraphics[width=0.38\textwidth]{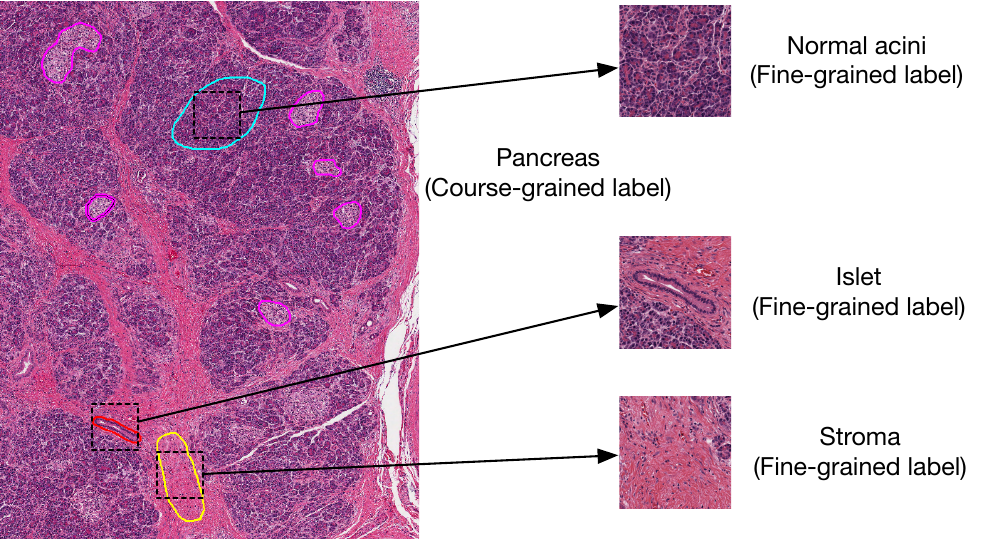}\label{fig:hierarchy}}
  \caption{(a) A collection of image sets sampled from CIFAR-100 are in the upper row. The coarse-grained label of a set is the most frequent superclass of images inside the set. WSI examples from TCGA and patches from NCT dataset are in the lower row. (b) Hierarchy of coarse- and fine-grained labels for histopathology images. %\yuxin{connecting to the main text in intro (organs, functional groups)}
  }
\end{figure*}

%\YC{Change the \figref{} macro to be first letter capital? Currently it's inconsistent with the table reference format}
% all have some initial in the ref, table, figure, section
As an illustrative example, %\YC{we consider histopathology image classification, where} 
we consider histopathology image classification where acquiring a substantial number of fine-grained labels for individual patches (e.g., tissue labels shown in the lower row of \figref{fig:wsi_tile_example}) is challenging. Conversely, a wealth of coarse-grained labels (e.g. the site of origin of the tumors associated with whole slide images (WSIs) from TCGA shown on the left-hand side of the upper row of \figref{fig:wsi_tile_example}) are easily available. This motivates us to leverage these abundant %and cost-efficient 
coarse-grained labels and hierarchical relationships, such as between organs and tissues (as depicted in \figref{fig:hierarchy}), to enhance representation learning. Tissues consist of cellular assemblies with shared functionalities, while organs are comprised of multiple tissues. This hierarchical relationship serves as a conceptual foundation for our representation learning and provides significant contextual information for facilitating representation learning. By using coarse-grained information within this hierarchy, our goal is to learn efficiently fine-grained tissue representations within WSIs. %Our hope is that our approach applies more generally in the case where hierarchical structures are present. 
Another example is shown in the upper row of \figref{fig:wsi_tile_example}. We emulate a programmatic labeler that use heuristics such as keywords, regular expression, or knowledge bases to solicit sets of images. The coarse-grained labels, e.g., the most frequent superclass of images in the set, can be used to facilitate representation learning for downstream tasks such as instance-level image classification.

\paragraph{Our contribution} In this paper, we introduce \underline{F}ine-gr\underline{A}ined representation learning from \underline{C}oarse-gra\underline{I}ned \underline{L}ab\underline{E}ls (FACILE), a novel generic representation learning framework that uses easily accessible coarse-grained annotations to quickly adapt to new fine-grained tasks. Distinct from existing practices in FSL and TL, our approach utilizes coarse-grained labels in the source domain. This sets our methodology apart from conventional FSL and TL techniques, which typically rely on meticulously labeled source data to train models. 

We provide an initial theoretical analysis to motivate the empirical success of FACILE and examine the convergence rate for the excess risk of downstream tasks under a novel Lipschitzness condition on the loss function concerning the fine-grained labels. Our study reveals that the availability of coarse-grained labels can lead to a substantial acceleration in the excess risk rate for fine-grained label prediction tasks, achieving a fast rate of $\cO(1/n)$, where $n$ represents the number of fine-grained data points. This analysis highlights the significant potential for leveraging coarse-grained labels to enhance the learning process in fine-grained label prediction tasks. 

In our experiments, we thoroughly investigate the effectiveness of FACILE through a series of extensive experiments on natural image datasets and histopathology image datasets. For natural image datasets, we sample input sets from training data from CIFAR-100 and use the unique superclass number and most frequent superclass as coarse-grained labels.  The generated datasets are used to evaluate different models. We also evaluate models by finetuning the fully connected layer appended to ViT-B/16 \citep{dosovitskiy2020image} of CLIP \citep{radford2021learning} in an anomaly detection dataset based on CUB200 \citep{he2019fine}. For histopathology applications, we leverage two large datasets with coarse-grained labels to pretrain our models. Subsequently, we evaluate the performance of these trained models on a diverse collection of histopathology datasets. Our algorithm achieves strong performance on 4 downstream datasets. Notably, when tested on LC25000 \citep{lc25000}, our model achieves roughly 90\% average ACC with 1,000 randomly sampled tasks which only have 5 fine-grained labeled data points for each of the 5 classes, outperforms the strongest baseline by roughly 13\% with logistic regression fine-grained classifier. We further evaluate various models by fine-tuning the fully connected layer appended to ViT-B/14 \citep{dosovitskiy2020image} of DINO V2 \citep{oquab2023dinov2}. These models can leverage the capability of ``foundation'' models and enhance the model performance on target tasks. Our experiments provide compelling evidence of the efficacy and generalizability of FACILE across various datasets, highlighting its potential as a robust representation learning framework.

\section{Fine-Grained Representation Learning from Coarse-Grained Labels}\label{sec_method}

% \subsection{Notation}

\paragraph{Notation} Our model pretrains on a collection of samples, denoted by $\{(s_i,w_i)\}_{i=1}^{m}$. Each $s_i$ is an input set of instances $\{x_j\}_{j=1}^{a}$, where $a$ represents the variable input set size.
$\{w_i\}$ are the set-level coarse-grained labels. The space of all instances is $\cX$ and the space of all instance labels, which we call fine-grained labels, is $\cY$. The space of pretraining data is $\cS\times \cW$, where $\cS=\left\{\{x_1, \ldots,x_a\}: x_j \in \cX \mathrm{\:for\:} \forall j \in [a]\right\}$ and $\cW$ denotes the space of coarse-grained labels. %We receive $(S,W)$ from product space $\cS\times \cW$ and $(X,Y)$, whose size may be small, from a product space $\cX\times \cY$. 
We receive $(X, Y)$ from product space $\cX\times \cY$ and corresponding $(S, W)$ from product space $\cS\times \cW$.
The goal is to predict the fine-grained labels ${y\in } \cY$ from the instance features ${x\in}\cX$. %The model could benefit from the information on the coarse-grained labels. 

% model
\subsection{The FACILE Algorithm}
\begin{wrapfigure}{r}{0.35\textwidth}
    \centering
    \vspace{-2mm}
    \includegraphics[width=0.25\textwidth]{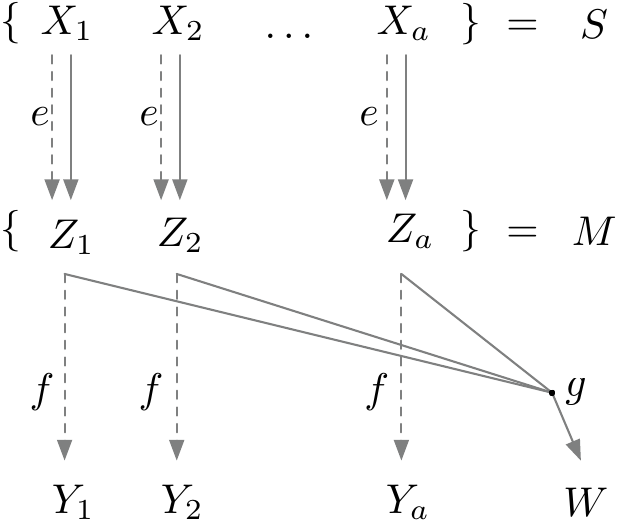}
    \caption{Schema of the FACILE model. The dotted lines represent the flow of fine-grained data, and the solid lines denote the flow of coarse-grained labels}
    \label{fig:schema}
\end{wrapfigure}

We study the model in an FSL setting where we have three datasets: (1) pretraining coarse-grained datasets $\datacg=\{(s_i,w_i)\}_{i=1}^{m}$ sampled i.i.d. from $P_{S, W}$ (2) fine-grained support dataset $\datafg=\{(x_i,y_i)\}_{i=1}^{n}$ sampled i.i.d., from $P_{X, Y}$, and (3) query set $\cD^{\mathrm{query}}$. The support set $\datafg$ contains $c$ classes and $k$ samples $x$ in each class (i.e., $n\equiv kc$). We assume a latent space $\cZ$ for embedding $Z$. We define instance feature maps $\cE =\{e: \cX \rightarrow \cZ\}$, set-input functions $\cG=\{g: \cM \rightarrow \cW\}$ where $\cM=\{\{z_1,\ldots,z_a\}:z_j \in \cZ \mathrm{\:for\:} j \in [a]\}$, and fine-grained label predictors $\cF=\{f: \cZ \rightarrow \cY\}$. 
%The instance feature map $e$ can also apply to $S$ and return the set of features in $\cZ$ for all instances. 
The corresponding set-input feature map of an instance feature map $e$ is defined as $\phi^{e}:\cS\rightarrow \cM$.
We assume the class of $f$ is parameterized and identify $f$ with parameter vectors for theoretical analysis. We then learn feature map $e$, fine-grained label predictor $f$, and predict fine-grained label with $f\circ e$. The schema of our model is illustrated in \figref{fig:schema}.

% loss
We assume two loss functions: $\ellfg: \cY\times \cY \rightarrow \mathbb{R}$ for fine-grained label prediction and $\ellcg: \cW\times \cW \rightarrow \mathbb{R}$ for coarse-grained label prediction. $\ellfg$ measures the loss of the fine-grained label predictor. We assume this loss is differentiable in its first argument. $\ellcg$ measures the loss of pretraining with coarse-grained labels. For theoretical analysis, we are interested in two particular cases of $\ellcg$: i) $\ellcg (w, w^{\prime})=\mathbbm{1}\left\{w \neq w^{\prime}\right\}$ when $\mathcal{W}$ is a categorical space; and ii) $\ellcg (w, w^{\prime})=\left\|w-w^{\prime}\right\|$ (for some norm $\left\|\cdot\right\|$ on $\mathcal{W}$) when $\mathcal{W}$ is a continuous space. We can also measure the loss of a feature map $e$ by $\ell^{\mathrm{cg}}_{e}=\ellcg (g_{e}\circ \phi^{e}(s), w)$, where $g_{e} \in \argmin _{g}\sE_{P_{S,W}} \ellcg (g\circ \phi^{e}(S), W)$. We assume there is an unknown ``good'' embedding $M=\phi^{e_0}(S)\in \cM$, by which a set-input function $g_{e_0}$ 
%and additive noise $N$ can determine $W$, i.e., $g_{e_0}(M)+N=g_{e_0}\circ e_0(S)+N=W$. 
can determine $W$, i.e., $g_{e_0}(M)=g_{e_0}\circ \phi^{e_0}(S)=W$. The strict assumption of equality can be relaxed by incorporating an additive error term into our risk bounds of $g_{e_0}\circ \phi^{e_0}$.

%Note that the strong equality assumption can be relaxed via an additive error term in our risk bounds that capture the risk of $g_{e_0}\circ \phi^{e_0}$

% objective
Our primary goal is to learn an instance label predictor or fine-grained label predictor $\hat{f} \circ \hat{e}$ that achieves low risk $\sE_{P_{X,Y}}[\ellfg(\hat{f}\circ \hat{e}(X), Y)]$ and we can bound the excess risk:
\begin{equation}
    \sE_{P_{X,Y}}[\ellfg(\hat{f}\circ \hat{e}(X),Y)-\ellfg(f^{*}\circ e^{*}(X),Y)] 
\end{equation}
where %$e^{*}$ and $f^{*}$ are defined 
$e^{*}\in \argmin_{e\in \cE}\sE _{P_{S,W}}\ell^{\mathrm{cg}}_{e}(S,W)$ and $f^{*} \in \argmin _{f\in \cF}\sE_{P_{X,Y}}[\ellfg(f\circ e^{*}(X), Y)]$. %\note{define space of embedding functions, predictor function space}

\begin{figure*}[t]
\centering
\includegraphics[width=.9\textwidth]{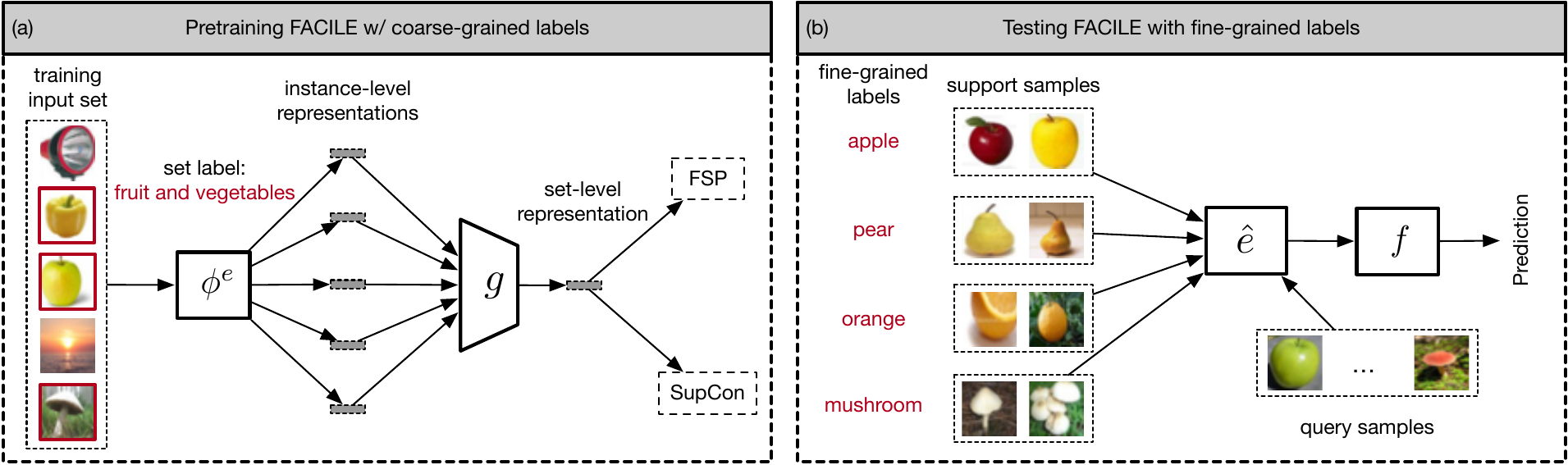}
\caption{An overview of the FACILE algorithm. (a) Pretraining step of FACILE with coarse-grained labels. The input is a set of images and the target is set-level coarse-grained label. $e$ is an instance feature map and $\phi^{e}$ is the corresponding set-input feature map. $g$ is the set-input model. We can instantiate the $\alg(\ellcg, \datacg, \cE)$ with any supervised learning algorithms, e.g., fully supervised pretraining (FSP) with cross-entropy loss and the SupCon model. (b) Fine-grained learning of FACILE with fine-grained labels. The learned instance feature map $\hat{e}$ extracts instance-level features from patches of the support set and query set. $f$ is the fine-grained label predictor.}
\label{fig:architecture}
\end{figure*}

\begin{algorithm}[!t]
\caption{FACILE algorithm}
\label{facile_alg}
\begin{algorithmic}[1]
 \STATE {\bfseries Input:} loss functions $\ellfg$, $\ellcg$, predictors $\cE$, $\cG$, $\cF$, datasets $\datacg$ and $\datafg$ 
\STATE {{obtain feature map $\hat{e} \leftarrow \alg(\ellcg, \datacg, \cE)$} }
\STATE create dataset $\dataaug=\{(z_i,y_i):z_i=\hat{e}(x_i),(x_i,y_i)\in \datafg\}_{i=1}^{n}$ 
%, where $z_i=\hat{e}(x_i)$ and $x_i\in \datafg$
\STATE obtain fine-grained label predictor { $\hat{f}\circ \hat{e}$, where 
$\hat{f} \leftarrow \alg(\ellfg, \dataaug, \cF)$}
\STATE {\bfseries Return:} $\hat{f}\circ \hat{e}$
\end{algorithmic}
\end{algorithm}
% %{{obtain feature map %$\hat{e} \leftarrow \alg_m(\ellcg, P_{S,W}, \cE)$} }
% % \STATE define distribution $\hat{P}(Z,Y)=P(Z,Y)\mathbbm{1}\{Z=\hat{e}(X)\}$ \yuxin{remove}
% %$\hat{f} \leftarrow \alg(\ellfg, \hat{P}_{Z,Y}, \cF)$ 
%change to P_XY
% describe it as generic alg
The pseudocode for FACILE is provided in \algref{facile_alg}, and we further illustrate the FACILE algorithm in \figref{fig:architecture}. Given an input set $s_i$ comprising instances ${x_1, \ldots, x_a}$, the feature map $e$ is employed to extract instance-level features for all the instances within the input set. Subsequently, a set-input model $g$ is utilized to generate set-level features based on the instance-level features. 
%Our FACILE model is pretrained on coarse-grained labels in conjunction with either the FSP or the SupCon model \citep{khosla2020supervised}. 
Our FACILE framework is designed to be a generic algorithm that is compatible with any supervised learning method in its pretraining stage. We chose SupCon (Supervised Contrastive Learning) \citep{khosla2020supervised} and FSP as they are representative of the two main approaches within supervised learning: contrastive and non-contrastive (traditional supervised) learning, respectively.
During testing, we extract the pretrained feature map $\hat{e}$ and finetune a classifier $f$ using the generated embeddings from $\hat{e}$ and the fine-grained labels of the support set. The performance of the classifier $\hat{f}$ is then reported for the query set. Note that \Algref{facile_alg} is generic since the two learning steps can use any supervised learning algorithm. 
%Similar to \citet{robinson2020strength}, our analysis treats the case where $\alg(\ellcg,  \datacg, \cE)$ is empirical risk minimization (ERM) and is agnostic to the choice of $\alg(\ellfg, \datafg, \cF)$. 

\subsection{Theoretical Analysis}

%In this section, we describe the theoretical motivation of our algorithm. 
We denote the underlying distribution of $\datacg$ as $P_{S,W}$ and the underlying distribution of $\datafg$ as $P_{X,Y}$. We assume the joint distribution of $Z$ and $Y$ is $P_{Z,Y}$. After we learn the feature map $\hat{e}$, we can define a new distribution $\hat{P}_{Z,Y}=P(Z,Y)\mathbbm{1}\{Z=\hat{e}(X)\}$, where $\mathbbm{1}$ is the indicator function. The $\dataaug$ is i.i.d. samples from $\hat{P}_{Z,Y}$. In order to include the underlying distribution of $\datacg$, and $\datafg$ into analysis, with a slight abuse of notation we use $\alg_m(\ellcg, P_{S,W}, \cE)$ to denote $\alg(\ellcg, \datacg, \cE)$ and use $\alg_n(\ellfg, \hat{P}_{Z,Y}, \cF)$ to denote $\alg(\ellfg, \dataaug, \cF)$.

%The two learning algorithms are described as follows. 
\begin{defn}\label{coarse-grained_learning}
(Coarse-grained learning; pretraining) Let $\mathrm{Rate}_{m}(\ellcg, P_{S,W}, \cE; \delta)$ (abbreviated to $\mathrm{Rate}_{m}(\ellcg, P_{S,W}, \cE)$) be the rate of $\alg_m(\ellcg, P_{S,W}, \cE)$ which takes $\ellcg$, $\cE$ and $m$ i.i.d. observations from $P_{S,W}$ as input, and return a feature map $\hat{e}\in \cE$ such that
\begin{equation*}
    \sE_{P_{S,W}}\ell^{\mathrm{cg}}_{\hat{e}}(S,W) \le \mathrm{Rate}_{m}(\ellcg, P_{S,W}, \cE; \delta)
\end{equation*}
with probability at least $1-\delta$.
\end{defn}

\begin{defn}\label{fine-grained_learning}
(Fine-grained learning; downstream task learning) Let $\mathrm{Rate}_n(\ellfg, P_{Z,Y}, \cF; \delta)$ (abbreviated to $\mathrm{Rate}_n(\ellfg, P_{Z,Y}, \cF)$) be the excess risk rate of $\alg_n(\ellfg, P_{Z,Y}, \cF)$ which take $\ellfg$, $\cF$, and $n$ i.i.d. observations from a distribution $P_{Z,Y}$ as input, and returns a fine-grained predictor $\hat{f}\in \cF$ such that
% \begin{equation*}
    $\sE_{P_{Z,Y}}\left[ \ell^{\mathrm{fg}}_{\hat{f}}(Z,Y) - \ell^{\mathrm{fg}}_{f^{*}}(Z,Y) \right] \le \mathrm{Rate}_n(\ellcg,P_{Z,Y}, \cF; \delta)$
% \end{equation*}
with probability at least $1-\delta$.
\end{defn}

Next, we introduce our relative Lipschitz assumption and the central condition for quantifying task relatedness. The Lipschitz property requires that small perturbations to the feature map $e$ that do not harm the pretraining task, do not affect the loss of downstream task much either. 
\begin{defn}\label{defn:l_lipschitz}
We say that $f$ is $L$-Lipschitz relative to $\cE$ if for all $s\in \cS$, $x\in s$, $y\in \cY$, and $e,e^{\prime}\in \cE$,
\begin{equation*}
    |\ellfg(f\circ e(x), y)-\ellfg(f\circ e^{\prime}(x), y)| \le L\ellcg(g_{e}\circ \phi^{e}(s), g_{e^{\prime}}\circ \phi^{e^{\prime}}(s) )
\end{equation*}
The function class $\cF$ is $L$-Lipschitz relative to $\cE$, if every $f\in \cF$ is $L$-Lipschitz relative to $\cE$.
\end{defn}

Definition~\ref{defn:l_lipschitz} generalizes the definition of $L$-Lipschitzness in \citet{robinson2020strength} to bound the downstream loss deviation through the loss of the set label predictions. In the special case where $s=\{x\}$, and $g$ is a classifier for the pretraining labels, our Lipschitz condition reduces to the Lipschitzness definition of \citet{robinson2020strength}. 

The central condition is well-known to yield fast rates for supervised learning \citep{van2015fast}. Please refer to \defref{defn:central_condition} for the definition of central condition. %However, it is not clear that our surrogate problem $(\ellfg,\hat{P}_{Z,Y}, \cF)$ would continue to satisfy the central condition\YC{This sentence is in conflict with the following sentence.}. 
We show that our surrogate problem $(\ellfg, \hat{P}_{Z,Y}, \cF)$ satisfies a central condition in \propref{prop:surrogate_weak}. 

\begin{theorem}\label{theorem:main}
Suppose that $(\ellfg, P_{Z,Y}, \cF)$ satisfies the central condition, $\cF$ is $L$-Lipschitz relative to $\cE$, $\ellfg$ is bounded by $B>0$, $\cF$ is $L^{\prime}$-Lipschitz in its $d$-dimensional parameters in the $l_2$ norm, $\cF$ is contained in the Euclidean ball of radius $R$, and $\cY$ is compact. We also assume that $\mathrm{Rate}_{m}(\ellcg, P_{S,W}, \cE)=\mathcal{O}\left(1 / m^\alpha\right)$. Then when $\alg_n(\ellfg, \hat{P}_{Z,Y}, \cF)$ is ERM we obtain excess risk $\sE_{P_{X,Y}}\left[ \ell^{\mathrm{fg}}_{\hat{f}\circ \hat{e}}(X, Y) -\ell^{\mathrm{fg}}_{f^{*}\circ e^{*}}(X,Y) \right]$ bound with probability at least $1-\delta$ by %follows,
% \begin{equation*}
    $\mathcal{O}\left(\frac{d\alpha \beta \log R L^{\prime}n + \log \frac{1}{\delta}}{n} + \frac{B+2L}{n^{\alpha \beta}}\right)$
% \end{equation*}
if $m=\Omega (n^{\beta})$ and $\ellcg(w, w^{\prime})=\mathbbm{1}\{w\ne w^{\prime}\}$.
\end{theorem}

For a typical scenario where $\mathrm{Rate}_{m}(\ellcg,P_{S,W}, \cE)=\cO(1/\sqrt{m})$, we can obtain fast rates with $m=\Omega(n^2)$. Similarly, in the scenario where $\alg_m(\ellcg, P_{S,W}, \cE)$ achieves fast rate, i.e., $\mathrm{Rate}_{m}(\ellcg,P_{S,W}, \cE)=\cO(1/m)$, one can obtains fast rates when $m=\Omega(n)$. More generally, if $\alpha \beta \ge 1$, we observe fast rates.

We prove our theorem by first showing that the excess risk of $\hat{f}\circ \hat{e}$ can be bounded by $2L\mathrm{Rate}_m(\ellcg, P_{S, W}, \cE) + \mathrm{Rate}_{n}(\ellfg, \hat{P}_{Z, Y}, \cF)$ in \propref{prop:excess_risk_split}. 
Then, we show that $(\ellfg,\hat{P}_{Z,Y},\cF)$ also satisfies the weak central condition in \propref{prop:surrogate_weak}. Thus, $\mathrm{Rate}_{n}(\ellfg, \hat{P}_{Z,Y}, \cF)$ is also bounded by \propref{prop:central_condition4fast_rate}. We refer interested readers to \secref{app:excess_risk_proof} for full details of the proof. 

In the next section, we first aim to empirically study the relationship between generalization error, coarse-grained dataset size, and fine-grained dataset size that our theoretical analysis predicts in \secref{sec:most_freq} and \secref{sec:hist_image}. We also demonstrate the exceptional efficacy of the proposed algorithm compared to baseline models on natural image datasets and histopathology image datasets.

%\YC{It would be useful to put a forward pointer to the relevant discussion on this fast rate in the experimental results section.}
% sucky description; delete
%While our theoretical analysis focuses on specific $\ellcg$ loss functions and $\ellfg$ may not be bounded by $B$ in the case of cross-entropy loss, our experimental findings demonstrate the exceptional efficacy of the proposed algorithm for cross-entropy loss. 
%Despite the theoretical limitations, our experiments strongly indicate that the algorithm performs exceptionally well in the context of cross-entropy loss, further validating its effectiveness in practical applications. 

\section{Empirical Study} 

\subsection{Baseline Models and Algorithm Instantiation}
% avoid overlap with related work

We consider two sets of baseline models: self-supervised models \citep{bachman2019learning,he2020momentum,chen2020simple,caron2020unsupervised,grill2020bootstrap,chen2021exploring} and weakly supervised models \citep{donahue2014decaf,sun2017revisiting,zeiler2014visualizing,robinson2020strength}. 

\paragraph{Self-supervised models} %For our large amount of 
Given pretraining data $(S,W)$, self-supervised learning models ignore the labels $W$ and learn $\hat{e}$ from $S$. Then, we can test $\hat{e}$ with a new task, which consists of a support set and a query set. A new model that leverages the learned $\hat{e}$ is finetuned on the support set and tested on the query set. We performed two self-supervised learning models in two categories, e.g., SimCLR \citep{chen2020simple} for contrastive learning and SimSiam \citep{chen2021exploring} for non-contrastive learning. Details of these self-supervised learning algorithms can be found in \secref{contrastive_non_contrastive}.

\paragraph{Weakly supervised models} 
We assign each instance, from the pretraining dataset, a label of the input set to which it belongs. We train feature map $\hat{e}$ appended with a linear classifier on the pretraining dataset. We call this model FSP-Patch, where FSP stands for fully supervised pretraining and the model is trained with the assigned instance-level labels. For a new task with a support set and a query set, we use the $\hat{e}$ to extract features for both sets, train a classifier on the support set features, and test the classifier on the query set features.

%We instantiate $\alg_m(\ellcg, P_{S,W}, \cE)$ with supervised learning and the SupCon model \citep{khosla2020supervised}, which are denoted as FACILE-FSP and FACILE-SupCon respectively. 

%The ResNet18 \citep{he2016deep} is used as feature maps $\cE$. 
Following previous work in FSL \citep{tian2020rethinking,chen2019closer}, we use $l _2$-normalized features for downstream tasks. 
% evaluation
Unless otherwise specified, we evaluate methods with 1,000 randomly sampled meta-tasks from each dataset. All meta-tasks use 15 samples per class as the query set. The average F1/ACC and 95\% confidence interval (CI) are reported.
We follow the test setting of \citet{yang2022towards} and use NearestCentroid (NC), LogisticRegression (LR), and RidgeClassifier (RC). 

\subsection{Pretrain with Unique Class Number of Input Sets}

\label{sec:uniq_class}
In order to show the advantages of using the coarse-grained labels, we introduce a new task of pretraining with the unique class number of input sets. Inspired by \citet{lee2019set}, we use the CIFAR-100 \citep{krizhevsky2009learning} dataset, which contains 100 classes grouped into 20 superclasses. We generate input sets by sampling between 6 and 10 images from CIFAR-100 training data. The targets of the input sets are the unique superclass number of the input sets. In our downstream tasks, we perform few-shot classifications of fine-grained classes. Despite being distinct from the downstream fine-grained labels, the coarse-grained labels offer useful information for learning useful representations for downstream tasks.

The ResNet18 \citep{he2016deep} is used as feature maps $\hat{e}$. For FACILE-FSP, we pretrain the feature map $\hat{e}$ from these input sets and targets with $\ell _{1}$ loss. The features of CIFAR-100 test images are extracted with $\hat{e}$. Training settings of SimSiam, SimCLR and FACILE-FSP can be found in \secref{app:pretrain_uniqc_class_and_most_freq}. We then test $\hat{e}$ in a few-shot manner. We random sample 5 classes, 5 examples from each class, for each meta-test dataset. The fine-grained label predictor $\hat{f}$ is trained on the support examples and tested on the query examples. The performance of these models is reported in \tabref{tab:pretrain_uniq_class_and_most_frequent}. We can see that FACILE-FSP outperforms self-supervised learning models by a large margin.

% \begin{table}[ht]
%     \centering
%     \scalebox{0.8}{
%     \begin{tabular}{c|c|c|c}
%     \hline
%         pretraining method & NC & LR & RC \\
%     \hline
%        SimCLR  & $76.07 \pm 0.97$ & $75.88 \pm 1.01$ & $75.50 \pm 1.02$ \\
%        SimSiam & $ 78.15 \pm 0.93$ & $ 79.44 \pm 0.92 $ & $ 79.03 \pm 0.95 $\\
%        FACILE-FSP & $\mathbf{86.25\pm 0.79}$ & $\mathbf{85.42 \pm 0.82}$ & $\mathbf{85.84 \pm 0.81}$\\
%         \hline
%     \end{tabular}
%     }
%     \caption{Pretraining on input sets from CIFAR-100. Testing with 5-shot 5-way meta-test sets; average F1 and CI are reported.}
%     \label{tab:pretrain_uniq_class}
% \end{table}
\vspace{-2mm}
\subsection{Pretrain with Most Frequent Class Label}
\vspace{-2mm}
\label{sec:most_freq}
We sample input sets randomly from training data of CIFAR-100. The targets are the most frequent superclass of the input sets. If there is a tie in an input set, we choose a random top frequent superclass as the target of the input set. Training settings are similar to \secref{sec:uniq_class} and can be found in \secref{app:pretrain_uniqc_class_and_most_freq}. 
The performances of all models are reported in \tabref{tab:pretrain_uniq_class_and_most_frequent}. We can see that FACILE-FSP obtains better results compared to other models.

% \begin{table}[ht]
%     \centering
%     \scalebox{0.99}{
%     \begin{tabular}{c|c|c|c}
%     \hline
%         pretraining method & NC & LR & RC \\
%     \hline
%        SimCLR  & $75.91 \pm 1.00$ & $75.82\pm 1.01$ & $75.91\pm 1.02$ \\
%        SimSiam & $78.80\pm 0.93$ & $79.44\pm 0.95$ & $79.43\pm 0.93$\\
%        FSP-Patch & $73.21 \pm 0.97$ & $73.92 \pm 0.98$ & $73.40 \pm 0.98$ \\
%        FACILE-SupCon & $79.54 \pm 0.92$ & $79.54 \pm 0.96$ & $79.12 \pm 0.95$ \\
%        FACILE-FSP & $\mathbf{82.04 \pm 0.84}$ & $\mathbf{81.70 \pm 0.91}$ & $\mathbf{81.75 \pm 0.90}$ \\
%         \hline
%     \end{tabular}
%     }
%     \caption{Pretraining on input sets from CIFAR-100. Testing with 5-shot 5-way meta-test sets; average F1 and CI are reported.}
%     \label{tab:pretrain_most_frequent}
% \end{table}

\begin{table}[ht]
    \centering
    \scalebox{0.75}{
    \begin{tabular}{c|ccc|ccc}
    \hline
    \multirow{2}{*}{pretraining method} & \multicolumn{3}{c}{unique superclass number} & \multicolumn{3}{c}{most frequent superclass} \\
           & NC & LR & RC & NC & LR & RC \\
    \hline
       SimCLR & $76.07 \pm 0.97$ & $75.88 \pm 1.01$ & $75.50 \pm 1.02$ & $75.91 \pm 1.00$ & $75.82\pm 1.01$ & $75.91\pm 1.02$ \\
       SimSiam & $ 78.15 \pm 0.93$ & $ 79.44 \pm 0.92 $ & $ 79.03 \pm 0.95 $ & $78.80\pm 0.93$ & $79.44\pm 0.95$ & $79.43\pm 0.93$\\
       FSP-Patch & N/A & N/A & N/A & $73.21 \pm 0.97$ & $73.92 \pm 0.98$ & $73.40 \pm 0.98$ \\
       FACILE-SupCon & N/A & N/A & N/A & $79.54 \pm 0.92$ & $79.54 \pm 0.96$ & $79.12 \pm 0.95$ \\
       FACILE-FSP & $\mathbf{86.25\pm 0.79}$ & $\mathbf{85.42 \pm 0.82}$ & $\mathbf{85.84 \pm 0.81}$ & $\mathbf{82.04 \pm 0.84}$ & $\mathbf{81.70 \pm 0.91}$ & $\mathbf{81.75 \pm 0.90}$ \\
        \hline
    \end{tabular}
    }
    \caption{Pretraining on input sets from CIFAR-100. Testing with 5-shot 5-way meta-test sets; average F1 and CI are reported.}
    \label{tab:pretrain_uniq_class_and_most_frequent}
\end{table}

Note that the excess risk bound of the form $b=C/n^{\gamma}$ implies a log-linear relationship $\log b = \log C - \gamma \log n$ between the error and the number of fine-grained labels. We can visually interpret the learning rate $\gamma$. We study two cases: when the number of coarse-grained labels $m$ grows linearly with the number of fine-grained labels, and when the number of coarse-grained labels $m$ grows quadratically with the number of fine-grained labels. 

\begin{wrapfigure}{r}{0.35\textwidth}
  \begin{center}
    \vspace{-4mm}
    \includegraphics[width=0.33\textwidth]%,trim={.2cm 1cm 1cm 3.0cm}]
    {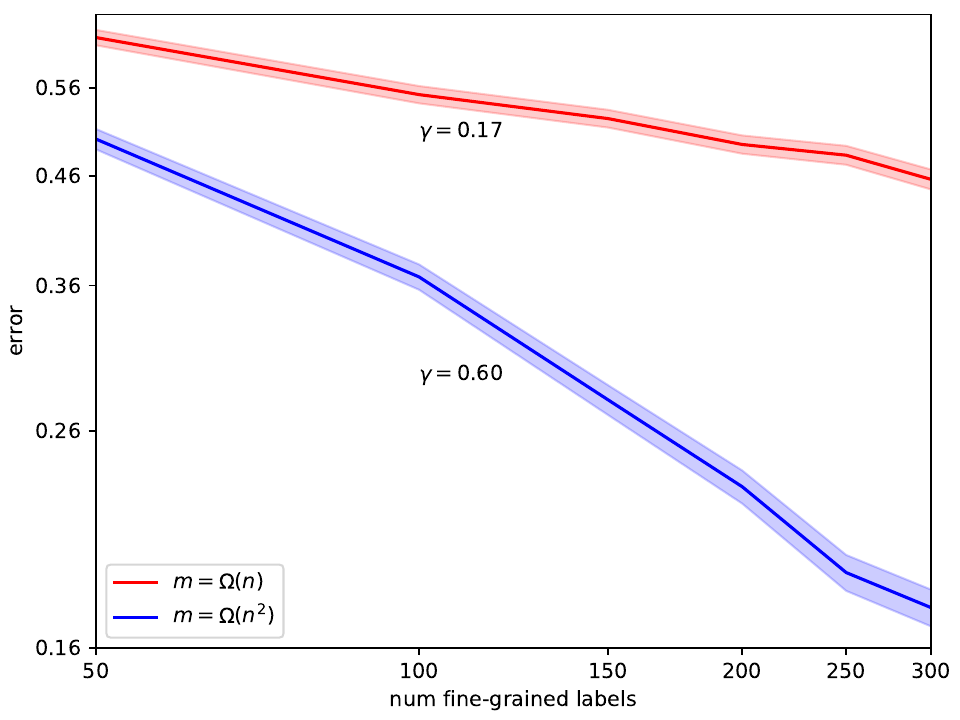}
  \end{center}
  \vspace{-3mm}
  \caption{Generalization error (with two growth rates) of FACILE-FSP on CIFAR-100 test dataset as a function of the number of coarse-grained labels $m$. %The FACILE-FSP trains on the input sets sampled from CIFAR-100 dataset with $m$ coarse-grained labels. 
  %We show the error curves with two growth rates of $m$.
  }
  \label{fig:excess_risk_freq_class}
  \vspace{-4mm}
\end{wrapfigure}

In order to show the generalization error rate of FACILE-FSP w.r.t. fine-grained label number on CIFAR-100 test dataset, we randomly sample 5 classes (i.e., 5-way testing) for each task. We then sample $n/5$ fine-grained examples in each class for the support set and sample $15$ examples for each class for the query set. The curves are shown in \figref{fig:excess_risk_freq_class}. The figure shows the log-linear relationship of FACILE-FSP's generalization error on downstream tasks w.r.t. fine-grained label number. This visualization effectively captures how coarse-grained label number $m$ impacts the model's generalization capabilities.

% \begin{figure*}[t]%
% \centering
% \includegraphics[width=0.5\textwidth]{Figs/most_freq/excess_risk.pdf}
% \caption{Generalization error on CIFAR-100 test dataset.The FACILE-FSP trains on the CIFAR-100 dataset with $m$ coarse-grained labels. We show the error curves with two growth rates of $m$.}
% \label{fig:excess_risk_freq_class}
% \end{figure*}

\subsection{Evaluation on Histopathology Images}
\label{sec:hist_image}
% \subsubsection{Datasets and Data Extraction} 
\paragraph{Datasets and data extraction} 
We pretrain our models using two independent sources of WSIs. First, we downloaded data from The Cancer Genome Atlas (TCGA) from the NCI Genomic Data Commons (GDC) \citep{heath2021gdc}. Two collections of non-overlapping patches with different patch sizes, i.e., $224\times 224$ and $1,000\times 1,000$ at 20X magnification. Background patches with high or low intensity were removed.  Because the number of patches generated with size $224\times 224$ at 20X magnification is very large, at most $1,000$ randomly selected patches are kept for each slide. The names of the tumors/organs, from which slides are collected, are used as coarse-grained labels. Second, we downloaded all clinical slides from the Genotype-Tissue Expression (GTEx) project \citep{lonsdale2013genotype}, which provides a resource for studying human gene expression and regulation in relation to genetic variation. We extracted non-overlapping patches with size $1,000\times1,000$ at 20X magnification and patches with intensity larger than 0.1 and smaller than 0.85 are kept. For these slides, we used the organs from which the tissues were extracted as coarse-grained labels. Examples and class distributions for the two datasets can be found in \secref{app:datasets}.

We test models on 3 public datasets: LC \citep{lc25000}, PAIP \citep{kim2021paip}, NCT \citep{kather_jakob_nikolas_2018_1214456} and 1 private dataset PDAC. Details of these datasets are deferred to \secref{app:datasets}. Note that the TCGA and GTEx have meticulously categorized an extensive array of cancer types and organs, covering a diverse range of tissues as outlined in the LC, PAIP, and NCT. The strategic use of WSI-level labels is rooted in their potential to enrich tissue-level classification. While these labels may appear broad, they encapsulate a wealth of underlying heterogeneity inherent to different cancer regions and tissue types.

% \subsubsection{Main Results}
%The models are trained at two scales: $224\times 224$ and $1,000\times 1,000$ at 20X magnification. 

\paragraph{Pretrain on TCGA with patch size $224\times 224$}
We first train models on TCGA patches with size $224\times 224$ at 20X magnification. After the models are trained, we test the feature map in these models on LC, PAIP, and NCT. Full details about FACILE-FSP, FACILE-SupCon, and baseline models' training settings can be found in \secref{app:pretrain_tcga_gtex}.

Latent augmentation (LA) has been shown to improve FSL performance for histopathology images \citep{yang2022towards}. We use faiss \citep{johnson2019billion} to perform k-means clustering. Following the setting of \cite{yang2022towards}, the number of prototypes in the base dictionary is 16. Each sample is augmented 100 times by LA. We refer readers to \secref{app:la} for details of LA.

\paragraph{Main results} The test result is shown in \tabref{tab:lc_paip_nct}. In order to show the performance improvement over models pretrained on natural image datasets, we report the performance of the FSP model pretrained on ImageNet. We can see from \tabref{tab:lc_paip_nct} that our model FACILE-FSP performs the best, with a large margin compared to other models. The contrastive learning model SimCLR performs worse than non-contrastive learning model SimSiam. A possible reason could be the small batch size we used for SimCLR. SimSiam maintains high performance even with small batch sizes. FSP-Patch achieves better performance compared to self-supervised learning models and the ImageNet pretrained model, which shows the usefulness of the coarse-grained labels for downstream tasks. More experiment results about test ACC on LC, PAIP and NCT datasets can be found in \secref{app:acc_on_lc_paip_nct}. Test result with larger shot number is in \secref{app:larger_shot}. We further pretrain models on GTEx and TCGA with patch size $1,000\times 1,000$ and test the models on our private dataset PDAC. We refer readers to \secref{app:pdac_result} for experiment results on PDAC dataset.

\begin{table}[ht]
    \centering
    \scalebox{0.8}{
    \begin{tabular}{c|c|c|c|c|c}
    \hline
        pretraining method & NC & LR & RC & LR+LA & RC+LA \\
    \hline
    \multicolumn{6}{c}{1-shot 5-way test on LC dataset} \\
    \hline
    ImageNet (FSP) & $63.26 \pm 1.46$ & $63.13 \pm 1.41$ & $63.24 \pm 1.40$ & $64.51 \pm 1.41$ & $64.95 \pm 1.39$ \\
    SimSiam & $65.83 \pm 1.32$ & $66.52 \pm 1.31$ & $66.24 \pm 1.32$ & $67.21 \pm 1.29$ & $67.83 \pm 1.33$ \\
    SimCLR  & $64.57 \pm 1.36$ & $63.85 \pm 1.37$ & $64.16 \pm 1.37$ & $65.78 \pm 1.33$ & $66.81 \pm 1.40$ \\
    FSP-Patch  & $66.73 \pm 1.29$ & $66.25 \pm 1.29$ & $66.59 \pm 1.28$ & $68.01 \pm 1.24$ & $68.28 \pm 1.26$ \\
    FACILE-SupCon & $67.75 \pm 1.31$ & $66.11\pm 1.37$ & $65.92 \pm 1.40$ & $68.57\pm 1.30$ & $69.79 \pm 1.33$ \\
    FACILE-FSP  & $\mathbf{75.09 \pm 1.30}$ & $\mathbf{73.57 \pm 1.29}$ & $\mathbf{73.16 \pm 1.33}$ & $\mathbf{74.03 \pm 1.28}$ & $\mathbf{72.88 \pm 1.34}$ \\
    \hline
    \multicolumn{6}{c}{5-shot 5-way test on LC dataset} \\
    \hline
    ImageNet (FSP) & $82.82 \pm 0.75$ & $80.13 \pm 0.82$ & $80.23 \pm 0.83$ & $84.70 \pm 0.70$ & $84.42 \pm 0.74$ \\
    SimSiam  & $85.12 \pm 0.68$ & $82.69 \pm 0.75$  & $82.80 \pm 0.76$  & $87.45 \pm 0.63$  & $87.50 \pm 0.66$ \\
    SimCLR & $83.45 \pm 0.77$  & $81.93 \pm 0.83$  & $81.40 \pm 0.89$  & $85.69 \pm 0.73$  & $84.93 \pm 0.79$ \\ 
    FSP-Patch  &  $84.96 \pm 0.64$ & $84.10 \pm 0.69$  & $84.45 \pm 0.68$  &  $86.31 \pm 0.65$ &  $86.29 \pm 0.68$ \\
    FACILE-SupCon & $84.77 \pm  0.67$ & $82.49 \pm  0.82$ & $82.10 \pm 0.86$ & $86.88 \pm  0.64$ & $86.92 \pm  0.65$ \\
    FACILE-FSP  & $\mathbf{90.67 \pm 0.54}$  & $\mathbf{89.18 \pm 0.61}$  & $\mathbf{89.02 \pm 0.63}$  &  $\mathbf{90.03 \pm 0.59}$  & $\mathbf{88.71 \pm 0.67}$  \\
    \hline
        \multicolumn{6}{c}{1-shot 3-way test on PAIP dataset} \\
    \hline
    ImageNet (FSP) & $45.96 \pm 1.22$ & $47.82 \pm 1.29$ & $47.43 \pm 1.29$ & $46.38 \pm 1.24$ & $44.90 \pm 1.24$ \\
    SimSiam & $46.43 \pm 1.21$  & $47.93 \pm 1.24$  & $47.74 \pm 1.23$  & $47.20 \pm 1.21$  &  $46.31 \pm 1.22$ \\
    SimCLR  & $44.51 \pm 1.16$ & $46.44 \pm 1.14$ & $45.59 \pm 1.15$ & $45.40 \pm 1.14$ &  $45.04 \pm 1.16$ \\
    FSP-Patch  &  $48.85 \pm 1.21$ & $49.44 \pm 1.26$  &  $50.27 \pm 1.22$ &  $49.76 \pm 1.20$ &  $48.44 \pm 1.21$ \\
    FACILE-SupCon & $\mathbf{52.55 \pm 1.25}$ & $\mathbf{52.10\pm 1.36}$ & $\mathbf{52.40 \pm  1.35}$ & $\mathbf{50.66\pm 1.28}$ & $\mathbf{50.27 \pm  1.28}$ \\
    FACILE-FSP  & $48.81 \pm 1.21$ &  $50.08 \pm 1.24$ & $50.75 \pm 1.23$ & $50.03 \pm 1.23$  &  $49.41 \pm 1.20$ \\
    \hline
    \multicolumn{6}{c}{5-shot 3-way test on PAIP dataset} \\
    \hline
    ImageNet (FSP) & $60.73 \pm 1.02$ & $61.21 \pm 1.12$ & $61.04 \pm 1.11$ & $61.66 \pm 0.91$ & $59.30 \pm 0.93$ \\
    SimSiam  & $62.88 \pm 0.97$  & $62.59 \pm 1.08$ & $63.48 \pm 1.04$ & $65.01 \pm 0.88$  &  $63.22 \pm 0.89$  \\
    SimCLR & $60.99 \pm 0.93$ & $61.38 \pm 1.00$ & $61.62 \pm 1.02$ & $62.39 \pm 0.91$ & $61.29 \pm 0.90$ \\
    FSP-Patch  &  $64.45 \pm 0.92$ & $64.60 \pm 0.98$  &  $64.49 \pm 0.99$ &  $64.08 \pm 0.89$ &  $62.79 \pm 0.89$ \\
    FACILE-SupCon & $65.59 \pm  0.96$ & $65.82\pm 1.06$ & $66.72 \pm  1.01$ & $67.88\pm 0.86$ & $66.25 \pm 0.87$ \\
    FACILE-FSP  & $\mathbf{66.61 \pm 0.91}$  & $\mathbf{67.57 \pm 0.95}$  & $\mathbf{67.78 \pm 0.95}$ &  $\mathbf{68.24 \pm 0.85}$ &  $\mathbf{67.20 \pm 0.86}$ \\
    \hline
        \multicolumn{6}{c}{1-shot 9-way test on NCT dataset} \\
    \hline
    ImageNet (FSP) & $57.35 \pm 1.68$ & $56.39 \pm 1.64$ & $56.08 \pm 1.64$ & $57.78 \pm 1.66$ & $55.85 \pm 1.64$ \\
    SimSiam & $63.60 \pm 1.62$ & $64.43 \pm 1.54$ & $64.79 \pm 1.53$ & $65.26 \pm 1.56$ & $65.39 \pm 1.53$ \\
    SimCLR & $59.73 \pm 1.57$ & $59.61 \pm 1.57$ & $59.34 \pm 1.56$ & $60.57 \pm 1.57$ & $60.99 \pm 1.53$ \\
    FSP-Patch  & $60.08 \pm 1.46$ & $61.55 \pm 1.50$ & $62.32 \pm 1.50$  & $61.99 \pm 1.42$ &  $60.62 \pm 1.38$ \\
    FACILE-SupCon & $64.56 \pm  1.52$ & $63.79 \pm  1.53$ & $63.94 \pm  1.52$ & $64.96 \pm 1.51$ & $65.08 \pm 1.47$ \\
    FACILE-FSP  &  $\mathbf{67.76 \pm 1.31}$ & $\mathbf{68.52 \pm 1.30}$ & $\mathbf{68.55 \pm 1.28}$ & $\mathbf{68.33 \pm 1.28}$ &  $\mathbf{67.72 \pm 1.28}$\\
    \hline
    \multicolumn{6}{c}{5-shot 9-way test on NCT dataset} \\
    \hline
    ImageNet (FSP) & $74.59 \pm 1.11$ & $73.21 \pm 1.13$ & $74.60 \pm 1.07$ & $76.68 \pm 1.04$ & $74.39 \pm 1.09$ \\
    SimSiam  & $79.97 \pm 1.05$ & $79.81 \pm 1.03$ & $80.84 \pm 0.98$ & $83.45 \pm 0.92$ & $83.61 \pm 0.90$ \\
    SimCLR & $76.80 \pm 1.09$ & $76.95 \pm 1.07$ & $78.25 \pm 1.03$ & $80.54 \pm 0.97$ & $81.13 \pm 0.95$ \\
    FSP-Patch  & $79.50 \pm 0.94$ & $79.54 \pm 0.95$ & $81.00 \pm 0.88$ & $82.42 \pm 0.81$ & $81.33 \pm 0.79$ \\
    FACILE-SupCon  & $79.97 \pm 0.96$ & $79.73 \pm 0.98$ & $80.58 \pm 0.93$ & $83.79 \pm 0.82$ & $84.03 \pm 0.79$ \\
    FACILE-FSP  & $\mathbf{86.45 \pm 0.62}$ & $\mathbf{87.74 \pm 0.59}$ & $\mathbf{87.97 \pm 0.58}$ & $\mathbf{88.00 \pm 0.59}$ & $\mathbf{86.92 \pm 0.61}$ \\
    \hline
    \end{tabular}
    }
    \caption{Test result on LC, PAIP, and NCT dataset; average F1 and CI are reported.}
    \label{tab:lc_paip_nct}
    %\vspace{-8mm}
\end{table}

\begin{wrapfigure}{r}{0.35\textwidth}
  \begin{center}
    \vspace{-2mm}
    \includegraphics[width=0.33\textwidth]%,trim={.2cm 1cm 1cm 3.0cm}]
    {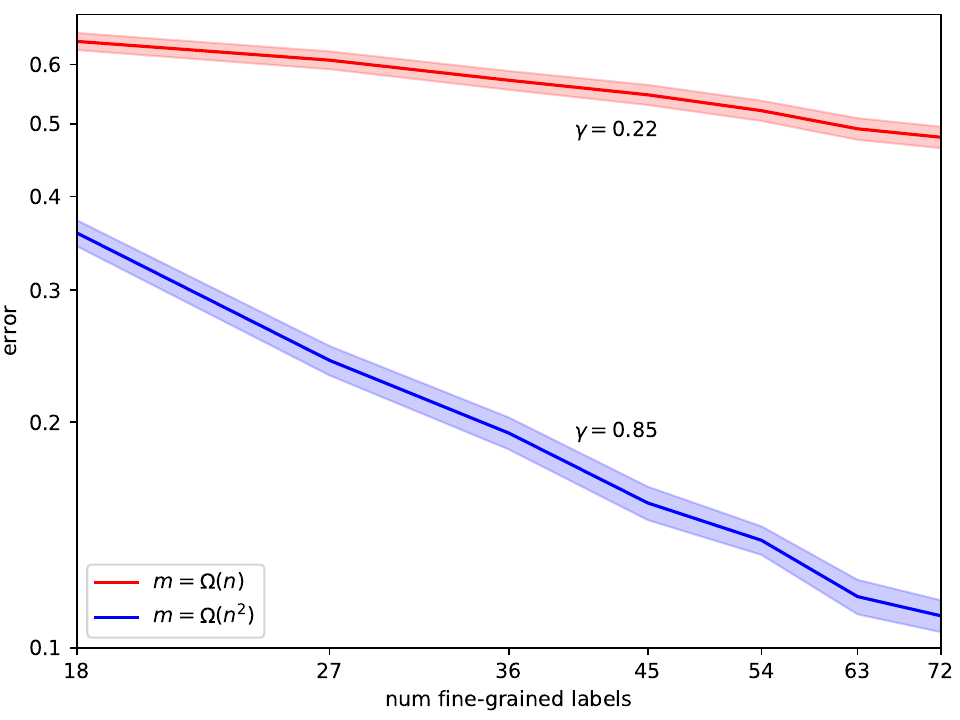}
  \end{center}
  \vspace{-2mm}
  \caption{Generalization error on NCT dataset. The FACILE-FSP trains on TCGA dataset with $m$ coarse-grained labels. We show the error curve with two growth rates of $m$.}
  \label{fig:excess_risk_tcga}
  %\vspace{-6mm}
\end{wrapfigure}

We show the generalization error of FACILE-FSP w.r.t. fine-grained label number in \figref{fig:excess_risk_tcga}. The figure reveals a pronounced log-linear relationship. A larger growth rate of coarse-grained labels implies a faster rate of excess risk. 

%\subsubsection{Benefits of pretraining on Large Pathology Datasets}
\paragraph{Benefits of pretraining on Large Pathology Datasets}
In order to show the benefits of pretraining on large pathology datasets, we pretrain different models on the NCT training dataset and test the performance on the LC dataset, following the setting of \citet{yang2022towards}. Instead of separating the mixture-domain and out-domain tasks, we directly report the average F1 and CI of LR models over all 5 classes of the LC dataset. Training details of the models can be found in \secref{app:benefits_of_large_dataset}. The test result on the LC dataset is shown in \tabref{tab:pretrain_nct}. 

We can see from \tabref{tab:pretrain_nct} the best model pretrained on NCT, i.e., FSP with strong augmentation, performs worse than our model FACILE-FSP in \tabref{tab:lc_paip_nct}. Our method get roughly 13\% improvement compared to \citet{yang2022towards} on the LC dataset. The large margin between the two best models pretrained on two different datasets shows the importance of pretraining with a large number of coarse-grained labels. More results on LC and PAIP can be found in \figref{tab:pretrain_nct_lc_paip}. Note that FSP with strong augmentation achieves much better performance than FSP with simple augmentation. It shows the importance of strong augmentation for pretraining. SimSiam model, trained with a batch size of 55, maintains competitive performance to MoCo v3 which needs a large batch size. 

\begin{table}[ht]
    \centering
    \scalebox{0.75}{
    \begin{tabular}{c|c|c|c|c|c}
    \hline
        pretraining method & NC & LR & RC & LR+LA & RC+LA \\
    \hline
       SimSiam  & $76.21\pm 0.87$ & $74.05\pm 1.10$ & $74.59\pm 1.10$ & $77.87\pm 0.87$ & $76.03\pm 0.94$\\
       %SimCLR & $82.09 \pm 0.78$ & $78.54 \pm 1.07$ & $78.19 \pm 1.13$ & $83.36 \pm 0.76$ & $82.35 \pm 0.88$ \\
        MoCo v3 (\citep{yang2022towards}) & $72.82\pm 1.25$ & $70.29\pm 1.43$ & $71.31\pm 1.40$ & $78.72\pm 1.00$ & $79.71\pm 0.95$\\ 
        FSP (simple aug; \citep{yang2022towards}) & $56.44\pm 1.50$ & $52.27\pm 1.81$ & $55.62 \pm 1.74$ & $63.47\pm 1.37$ & $63.47\pm 1.46$ \\
        FSP (strong aug) & $\mathbf{83.53 \pm 0.79}$ & $\mathbf{80.81 \pm 1.01}$ & $\mathbf{80.27 \pm 1.08}$ & $\mathbf{85.57 \pm 0.77}$ & $\mathbf{84.06 \pm 0.89}$ \\
        SupCon & $81.51 \pm 0.85$ & $78.77 \pm 1.03$ & $78.65 \pm 1.08$ & $83.51 \pm 0.84$ & $83.31 \pm 0.91$ \\
        \hline
    \end{tabular}
    }
    \caption{Pretraining on NCT and 5-shot 5-way testing on LC dataset; %with 5-shot 5-way support sets
    average F1 and CI are reported.}
    \label{tab:pretrain_nct}
    %\vspace{-5mm}
\end{table}

\paragraph{More experiments and ablation study} We refer interested readers to \secref{app:ablation} for ablation studies about set size, training procedures, and set-input models. We provide additional insights through finetuning experiments. In Appendix \ref{app:finetune_clip}, we detail the finetuning of the ViT-B/16 \citep{dosovitskiy2020image} from CLIP \citep{radford2021learning} using CUB200-based anomaly detection data \citep{he2019fine}. Similarly, Appendix \ref{app:finetune_dino_v2} discusses finetuning the ViT-B/14 model of DINO V2 \citep{oquab2023dinov2} on TCGA dataset. These experiments extend our analysis to specialized tasks, showcasing the adaptability of FACILE to foundation models.
\section{Related Work}

\paragraph{Weakly supervised learning}
The concept of weakly supervised learning is introduced as a means to alleviate the annotation bottleneck in the training of machine learning models. There has been a large body of existing work in learning with only weak labels. A comprehensive survey about weakly supervised learning is provided in \citet{zhou2018brief,zhang2022survey}. We study a novel form of weak supervision which is provided by set-level coarse-grained labels. Among weakly supervised learning methods, \citet{robinson2020strength} studied the generalization properties of weakly supervised learning and proposed a generic learning algorithm that can learn from weak and strong labels and can be proved to achieve a fast rate. The authors consider a different setting where each instance has a weak label and a strong label, and the strong label predictor learns to predict the strong labels from the instances and their corresponding embeddings learned with weak labels. We consider the setting where we have some coarse-grained labels of some sets, rather than instances and the downstream classifiers only use the learned embeddings to train and test on the downstream tasks. %\YC{this paragraph is missing one or two sentences drawing the connection and difference with our approach.} 

% \paragraph{FSL with H\&E images}
% \citet{medela2019few} used a triplet loss to pretrain an encoder and finetuned SVM to adapt to the target domain for few-shot learning. \citet{sikaroudi2020supervision} explored deep neural networks and triplet loss for representation learning of histopathology images and investigated the notion of similarity and dissimilarity in WSIs. \citet{teh2020learning} learned transferable features with ProxyNCA \citep{movshovitz2017no} from weakly labeled data, which were collected from various parts of the body and organized by non-medical experts. \citet{yang2022towards} introduced latent augmentation (LA) to contrastive learning and showed competitive performance on downstream few-shot tasks of histopathology images compared to SOTA FSL methods.

\paragraph{Multiple-instance learning for WSIs}
WSI classification and regression are formulated based on multiple-instance learning (MIL) \citep{campanella2019clinical, xu2022risk, ilse2018attention, sharma2021cluster, hashimoto2020multi, shao2021transmil, yao2020whole, lu2021data, lu2021ai, chen2021multimodal, li2021dual, chen2021whole, myronenko2021accounting, xiang2022exploring, javed2022additive}. These MIL models employ two procedures: i) feature extraction for patches cropped from a WSI and ii) aggregation of features from the same WSI. ImageNet pretrained backbones, self-supervised backbones pretrained on histopathology images, or backbones finetuned during training are used to extract features from patches. Deep attention pooling, graph neural networks, or sequence models, adapted for WSIs, are used for feature aggregation. In this paper, we 
consider a different problem setting where we enhance patch-level classification with related set-level labels. In the application of histopathology images, line 2 of our generic algorithm can be instantiated with any MIL models that have the backbones with trainable modules to extract patch-level features, e.g., \citet{ilse2018attention}. A complete comparison of MIL models for WSIs is out of the scope of this paper.
\vspace{-2mm}
\paragraph{Learning from coarsely-labeled data}
Another related line of research is %that is different from our setting but related is \
\citet{phoo2021coarsely}, where the authors assume a taxonomy of classes with two levels, i.e., a set of fine-grained classes that are more challenging to annotate and a set of coarse-grained classes that are easier to annotate. In our paper, we do not assume a taxonomy of classes for the coarse-grained and fine-grained labels. The coarse-grained and fine-grained labels are closely related in a more sophisticated way. Besides, the inputs that are fed to models to predict the coarse-grained or fine-grained labels are different, i.e., set input for coarse-grained labels and instances for fine-grained labels.

\vspace{-2mm}
\section{Conclusion and Discussion}\label{disc}
\vspace{-2mm}

\paragraph{Summary} We introduce FACILE, a representation learning framework that leverages coarse-grained labels for model training and enhances model performance for downstream tasks. Our theoretical analysis highlights the significant potential of leveraging set-level coarse-grained labels to benifit the learning process of fine-grained label prediction tasks. To demonstrate the effectiveness of FACILE, we conduct pretraining on CIFAR-100-based datasets and two large public histopathology datasets using coarse-grained labels and evaluate our model on a diverse collection of datasets with fine-grained labels. 

\paragraph{Limitation and future work}
In this paper, we consider a novel problem setting where we enhance downstream fine-grained label classification with easily available coarse-grained labels and propose a generic algorithm that contains two supervised learning steps. As an initial step towards addressing this issue, it's important to recognize that the separate utilization of loosely related coarse-grained labels and fine-grained labels can be inefficient. Besides, the pretraining of our proposed algorithm could be expensive given the possible large amount of coarse-grained data and the nature of set-input models. We are motivated to explore methods for efficient and robust representation learning with coarse-grained labels. Specifically, we are interested in investigating methods of selecting a subset of the coarse-grained dataset to accelerate pretraining and make representation learning robust.

\bibliography{iclr2024_conference}
\bibliographystyle{iclr2024_conference}

% easy method for extracting main paper
%\end{document}

\appendix
\appendix
\onecolumn

\section{Training Details}
\subsection{Pretrain with Unique Class Number and Most Frequent Class of Input Sets}
\label{app:pretrain_uniqc_class_and_most_freq}
In our study, an epoch refers to going through all the input sets in the dataset once. SimSiam is trained for 2,000 epochs using a batch size of 512. SGD is employed with a learning rate of 0.1, weight decay of 1e-4, and momentum of 0.9. The training process incorporates a cosine scheduler. Similarly, SimCLR is trained for 2,000 epochs with a batch size of 256 and a temperature of 0.07. SGD is used with a learning rate of 0.05, weight decay of 1e-4, and momentum of 0.9. The training also utilizes a cosine scheduler. 

We train FSP-Patch for 800 epochs with a batch size of 256. The SGD is used with a weight decay of 1e-4, momentum of 0.9, and cosine scheduler. 

FACILE-FSP is trained for 800 epochs with a batch size of 64. SGD is used with a learning rate of 0.0125, weight decay of 1e-4, and momentum of 0.9. $\ell_1$ loss is optimized for pretraining with unique class numbers of input sets. For FACILE-SupCon, we train the model with 2,000 epochs and a batch size of 256. An additional temperature parameter is set to 0.07. The SGD is used with a learning rate of 0.05, weight decay of 1e-4, and momentum of 0.9. 

\subsection{Finetune ViT-B/16 of CLIP with CUB200}
\label{app:finetune_clip_training}
SimSiam is trained for 400 epochs using a batch size of 64. SGD is used with an initial learning rate of 0.0125, weight decay of 1e-4, and momentum of 0.9. The cosine scheduler is used for the optimizer. SimCLR is also trained for 400 epochs with a batch size of 64. An additional temperature parameter is set to 0.07. SGD is used with a learning rate of 0.0125, weight decay of 1e-4, and momentum of 0.9. The training also uses a cosine scheduler.

FACILE-FSP is trained for 200 epochs with a batch size of 64. SGD is used with a learning rate of 0.0125, weight decay of 1e-4, and momentum of 0.9. For FACILE-SupCon, we train the model with 800 epochs and a batch size of 64. An additional temperature parameter is set to 0.07. The SGD is used with an initial learning rate of 0.0125, weight decay of 1e-4, and momentum of 0.9. Both models' training utilized a cosine annealing scheduler.

\subsection{Pretrain ResNet18 with TCGA and GTEx Dataset}
\label{app:pretrain_tcga_gtex}
In SimSiam, SimCLR, and FSP-Patch models, the data loader samples one patch for each slide. In FACILE-FSP and FACILE-SupCon, the data loader samples a set of $a$ patches for each slide. 

SimSiam is trained for 5,000 epochs using a batch size of 55. SGD is employed with a learning rate of 0.01, weight decay of 1e-4, and momentum of 0.9. The training process incorporates a cosine scheduler. Similarly, SimCLR is trained for 5,000 epochs with a batch size of 32. An additional temperature parameter is set to 0.07. SGD is used with a learning rate of 0.006, weight decay of 1e-4, and momentum of 0.9. The training also utilizes a cosine scheduler.

FSP-Patch is trained for 1,000 epochs with a batch size of 64. We employ SGD with a learning rate of 0.05, weight decay of 1e-4, and momentum of 0.9. The training process includes the utilization of a cosine scheduler.

FACILE-FSP is trained for 1,000 epochs with batch size 16. The input set size is 5 by default. We employ SGD with a learning rate of 0.0125, weight decay of 1e-4, and momentum of 0.9. The training process includes the utilization of a cosine scheduler. Set Transformer with 3 inducing points and 4 attention heads is used for the set-input model $g$. Similarly, for our FACILE-SupCon model, we use the same input set size and set-input model. The batch size is 8. An additional temperature parameter is set to 0.07. We use SGD with a learning rate of 0.0016, weight decay of 1e-4, and momentum of 0.9. We use an MLP as a projection head with two fc layers, a hidden dimension of 512, and an output dimension of 512.

\subsection{Finetune ViT-B/14 of DINO V2 with TCGA}
\label{app:finetune_dino_v2_training}
SimSiam is trained for 400 epochs with a batch size of 64, utilizing Stochastic Gradient Descent (SGD) with an initial learning rate of 0.0125, a weight decay of 1e-4, and a momentum of 0.9. A cosine scheduler was employed. SimCLR underwent a similar training regimen for 400 epochs and a batch size of 64, with an additional temperature parameter set at 0.07 and identical SGD parameters, including the use of a cosine scheduler for learning rate adjustments.

FSP-Patch also completed 400 epochs of training with a batch size of 64. The model employed SGD with a learning rate of 0.0125, a weight decay of 1e-4, and a momentum of 0.9, along with a cosine scheduler to modulate the learning rate.

For FACILE-FSP, training spanned 200 epochs with a batch size of 64, using SGD with the same learning rate, weight decay, and momentum settings. FACILE-SupCon extended its training to 800 epochs with a batch size of 64, including an additional temperature setting of 0.07 and the same SGD configuration. Both FACILE-FSP and FACILE-SupCon models utilized a cosine annealing scheduler.

\section{Additional Result}
\label{app:additional_result}

\subsection{Finetune CLIP Model with Anomaly Detection Dataset}
\label{app:finetune_clip}
In this experiment, we sought to enhance model performance with coarse-grained labels of the anomaly detection datasets \citep{zaheer2017deep,lee2019set}. A total of 11,788 input sets of size 10 are constructed from the CUB200 \citep{he2019fine} training dataset by including one example that lacks an attribute common to the other examples in the input set. The coarse-grained labels are the positions of the anomalies. This setup creates a challenging scenario for models, as they must identify the outlier among otherwise similar instances. In downstream tasks, we evaluate the finetuned feature encoder composed of the fixed CLIP \citep{radford2021learning} image encoder ViT-B/16 and appended fully-connected layer on the classification of species of the CUB200 test dataset. The batch normalization \citep{ioffe2015batch} and ReLU are applied to the fully-connected layer. 

% explain why coarse-grained labels help for downstream task
Following this experiment setup, the rationale behind utilizing coarse-grained labels is grounded in their potential to enhance model discernment in downstream tasks. By training the model to identify anomalies in sets where one item diverges from the rest, we essentially teach it to focus on subtle differences and critical attribute features. This enhanced focus is particularly beneficial for fine-grained classification tasks in the CUB200 test dataset, where distinguishing between closely related species requires the model to recognize and prioritize minute, yet significant, differences. 

The model training approach in this experiment centered around the CLIP image encoder, enhanced with an additional fully-connected layer. FACILE-FSP and FACILE-SupCon incorporate this setup, utilizing the CLIP-based feature encoder and focusing on finetuning the fully-connected layer through the FACILE pretraining step. In contrast, the SimSiam approach leverages the CLIP image encoder as a backbone while finetuning the projector and predictor components. Similarly, the SimCLR method also uses the CLIP encoder as its foundation but focuses solely on finetuning the projector. These varied strategies reflect our efforts to optimize the feature encoder for accurately identifying anomalies and improving classification performance in related tasks. The training details can be found in \secref{app:finetune_clip_training}.

\begin{table}[ht]
    \centering
    \scalebox{0.99}{
    \begin{tabular}{c|c|c|c}
    \hline
        pretraining method & NC & LR & RC \\
    \hline
       CLIP (ViT-B/16) & $83.84 \pm 1.10$ & $ 81.01 \pm 1.23 $ &  $ 82.75 \pm 1.17 $ \\
       SimCLR  & $ 84.03 \pm 1.08$ & $83.49 \pm 1.14$ & $ 86.30 \pm 1.03 $ \\
       SimSiam & $ 84.02 \pm 1.10 $ & $ 83.90 \pm 1.13 $ & $ 85.68 \pm 1.07 $\\
       FACILE-SupCon & $ 87.49 \pm 0.99 $ & $ 86.57 \pm 1.07 $ & $ 88.01 \pm 0.99 $ \\
       FACILE-FSP & $\mathbf{  88.74 \pm 0.94}$ & $\mathbf{ 88.45 \pm 0.96 }$ & $\mathbf{ 88.36 \pm 0.95 }$ \\
        \hline
    \end{tabular}
    }
    \caption{Pretraining on input sets from CUB200. Testing with 5-shot 20-way meta-test sets; average F1 and CI are reported.}
    \label{tab:pretrain_anomaly_detection}
\end{table}

% We can see from \tabref{tab:pretrain_anomaly_detection} that all models can benefit from the the data of target domain. FACILE-SupCon and FACILE-FSP get better performance compared other baseline models.

\tabref{tab:pretrain_anomaly_detection} clearly demonstrates that all models tested benefit from incorporating data from the target domain. Notably, both FACILE-SupCon and FACILE-FSP exhibit superior performance compared to other baseline models. This observation underscores the effectiveness of our models in leveraging coarse-grained labels to enhance their anomaly detection capabilities.

\subsection{Pretrain ResNet18 with TCGA}
\subsubsection{ACC on LC, PAIP, and NCT Datasets}
\label{app:acc_on_lc_paip_nct}
We pretrain the models on TCGA datasets with patches size $224\times 224$ at 20X magnification. Then, these pretrained models are tested on LC, PAIP, and NCT datasets. The average ACC and CI on the LC, PAIP, and NCT datasets are shown in \tabref{tab:lc_paip_nct_acc}.
\begin{table}[ht]
    \centering
    \scalebox{0.85}{
    \begin{tabular}{c|c|c|c|c|c}
    \hline
        pretraining method & NC & LR & RC & LR+LA & RC+LA \\
    \hline
    \multicolumn{6}{c}{1-shot 5-way test on LC dataset} \\
    \hline
    ImageNet (FSP) & $65.64 \pm 0.49$ & $66.06 \pm 0.46$ & $65.92 \pm 0.48$ & $66.60 \pm 0.48$ & $67.09 \pm 0.47$ \\
    SimSiam  & $68.88 \pm 0.51$ & $68.53 \pm 0.48$ & $68.27 \pm 0.48$ & $68.81 \pm 0.49$ & $70.24 \pm 0.47$ \\
    SimCLR  & $66.41 \pm 0.48$ & $66.52 \pm 0.46$ & $66.10 \pm 0.46$ & $67.70 \pm 0.45$ & $68.71 \pm 0.46$ \\
    FSP-Patch  & $68.56 \pm 0.46$ & $68.51 \pm 0.45$ & $68.68 \pm 0.46$ & $69.38 \pm 0.46$ & $69.63 \pm 0.46$ \\
    FACILE-SupCon  & $70.10 \pm 0.50$ & $69.15 \pm 0.47$ & $68.70 \pm 0.46$ & $70.35 \pm 0.47$ & $71.43 \pm 0.48$ \\
    FACILE-FSP   & $\mathbf{76.76 \pm 0.47}$ & $\mathbf{75.94 \pm 0.44}$ & $\mathbf{75.61 \pm 0.43}$ & $\mathbf{75.97 \pm 0.45}$ & $\mathbf{74.76 \pm 0.43}$\\
    \hline
    \multicolumn{6}{c}{5-shot 5-way test on LC dataset} \\
    \hline
    ImageNet (FSP)  & $82.79 \pm 0.32$ & $81.31 \pm 0.31$ & $81.13 \pm 0.30$ & $84.50 \pm 0.30$ & $84.73 \pm 0.28$ \\ 
    SimSiam   & $85.12 \pm 0.30$ & $83.39 \pm 0.32$ & $83.85 \pm 0.30$ & $87.74 \pm 0.27$ & $87.90 \pm 0.26$ \\ 
    SimCLR  & $83.75 \pm 0.30$ & $82.38 \pm 0.30$ & $82.32 \pm 0.31$ & $86.12 \pm 0.28$ & $85.40 \pm 0.30$ \\ 
    FSP-Patch   & $85.15 \pm 0.29$ & $84.38 \pm 0.31$ & $85.01 \pm 0.29$ & $86.71 \pm 0.28$ & $86.24 \pm 0.27$ \\ 
    FACILE-SupCon  & $85.04 \pm 0.30$ & $83.39 \pm 0.29$ & $83.44 \pm 0.30$ & $86.89 \pm 0.28$ & $87.25 \pm 0.29$  \\ 
    FACILE-FSP   & $\mathbf{90.76 \pm 0.24}$ & $\mathbf{89.67 \pm 0.24}$ & $\mathbf{89.22 \pm 0.25}$ & $\mathbf{90.54 \pm 0.23}$ & $\mathbf{88.98 \pm 0.26}$ \\ 
    \hline
        pretraining method & NC & LR & RC & LR+LA & RC+LA \\
    \hline
    \multicolumn{6}{c}{1-shot 3-way test on PAIP dataset} \\
    \hline
    ImageNet (FSP)  & $48.44 \pm 0.65$ & $50.34 \pm 0.65$ & $50.21 \pm 0.62$ & $48.90 \pm 0.62$ & $47.51 \pm 0.59$ \\
    SimSiam  & $49.42 \pm 0.65$ & $50.25 \pm 0.65$ & $49.76 \pm 0.65$ & $49.51 \pm 0.62$ & $49.09 \pm 0.63$ \\
    SimCLR   & $47.39 \pm 0.59$ & $48.35 \pm 0.59$ & $47.97 \pm 0.58$ & $47.77 \pm 0.59$ & $47.65 \pm 0.60$ \\
    FSP-Patch   & $51.61 \pm 0.68$ & $51.61 \pm 0.67$ & $52.06 \pm 0.67$ & $51.74 \pm 0.66$ & $51.38 \pm 0.66$ \\
    FACILE-SupCon  & $\mathbf{54.02 \pm 0.66}$ & $\mathbf{55.51 \pm 0.69}$ & $\mathbf{55.12 \pm 0.69}$ & $\mathbf{54.43 \pm 0.69}$ & $\mathbf{53.14 \pm 0.68}$ \\
    FACILE-FSP   & $51.77 \pm 0.67$ & $53.16 \pm 0.66$ & $52.86 \pm 0.62$ & $52.50 \pm 0.68$ & $52.49 \pm 0.64$ \\
    \hline
    \multicolumn{6}{c}{5-shot 3-way test on PAIP dataset} \\
    \hline
    ImageNet (FSP)  & $62.46 \pm 0.52$ & $62.48 \pm 0.48$ & $63.14 \pm 0.50$ & $62.11 \pm 0.51$ & $60.52 \pm 0.49$ \\
    SimSiam  & $63.05 \pm 0.52$ & $64.44 \pm 0.49$ & $64.66 \pm 0.50$ & $65.44 \pm 0.53$ & $64.64 \pm 0.55$ \\
    SimCLR  & $61.48 \pm 0.52$ & $61.84 \pm 0.53$ & $62.75 \pm 0.51$ & $63.03 \pm 0.52$ & $61.70 \pm 0.52$ \\
    FSP-Patch   & $65.29 \pm 0.49$ & $65.81 \pm 0.51$ & $65.98 \pm 0.48$ & $65.70 \pm 0.50$ & $64.01 \pm 0.52$ \\
    FACILE-SupCon & $67.12 \pm 0.48$ & $67.20 \pm 0.47$ & $67.44 \pm 0.50$ & $68.18 \pm 0.52$ & $67.27 \pm 0.51$ \\
    FACILE-FSP   & $\mathbf{68.23 \pm 0.52}$ & $\mathbf{68.10 \pm 0.51}$ & $\mathbf{68.50 \pm 0.49}$ & $\mathbf{68.60 \pm 0.50}$ & $\mathbf{67.81 \pm 0.54}$ \\
    \hline
        pretraining method & NC & LR & RC & LR+LA & RC+LA \\
    \hline
    \multicolumn{6}{c}{1-shot 9-way test on NCT dataset } \\
    \hline
    ImageNet (FSP) & $58.75 \pm 0.35$ & $58.66 \pm 0.36$ & $58.48 \pm 0.34$ & $58.83 \pm 0.36$ & $57.32 \pm 0.36$ \\
    SimSiam  & $64.76 \pm 0.40$ & $66.09 \pm 0.39$ & $66.09 \pm 0.39$ & $66.54 \pm 0.40$ & $67.05 \pm 0.41$ \\
    SimCLR  & $60.47 \pm 0.41$ & $61.17 \pm 0.38$ & $61.43 \pm 0.39$ & $61.65 \pm 0.40$ & $62.48 \pm 0.38$ \\
    FSP-Patch   & $61.03 \pm 0.42$ & $63.53 \pm 0.40$ & $63.26 \pm 0.42$ & $62.75 \pm 0.43$ & $61.57 \pm 0.42$ \\
    FACILE-SupCon  & $65.63 \pm 0.39$ & $65.46 \pm 0.38$ & $65.56 \pm 0.38$ & $66.24 \pm 0.38$ & $65.60 \pm 0.41$ \\
    FACILE-FSP   & $\mathbf{69.24 \pm 0.44}$ & $\mathbf{70.32 \pm 0.43}$ & $\mathbf{70.36 \pm 0.41}$ & $\mathbf{69.56 \pm 0.41}$ & $\mathbf{69.06 \pm 0.42}$ \\
    \hline
    \multicolumn{6}{c}{5-shot 9-way test on NCT dataset} \\
    \hline
    ImageNet (FSP)  & $74.82 \pm 0.26$ & $74.35 \pm 0.26$ & $75.20 \pm 0.26$ & $77.11 \pm 0.23$ & $74.89 \pm 0.26$ \\
    SimSiam   & $80.59 \pm 0.23$ & $80.51 \pm 0.23$ & $81.54 \pm 0.21$ & $83.68 \pm 0.22$ & $83.85 \pm 0.22$ \\
    SimCLR  & $77.30 \pm 0.25$ & $77.64 \pm 0.24$ & $79.17 \pm 0.24$ & $80.99 \pm 0.24$ & $81.71 \pm 0.23$ \\
    FSP-Patch   & $79.61 \pm 0.25$ & $79.89 \pm 0.24$ & $81.71 \pm 0.23$ & $82.92 \pm 0.24$ & $81.67 \pm 0.24$ \\
    FACILE-SupCon   & $80.14 \pm 0.25$ & $80.11 \pm 0.22$ & $81.41 \pm 0.24$ & $84.12 \pm 0.22$ &  $84.27 \pm 0.22$ \\
    FACILE-FSP   & $\mathbf{86.87 \pm 0.22}$ & $\mathbf{87.70 \pm 0.20}$ & $\mathbf{88.53 \pm 0.19}$ & $\mathbf{88.21 \pm 0.20}$ & $\mathbf{87.24 \pm 0.22}$ \\
    \hline
    \end{tabular}
    }
    \caption{Models tested on LC, PAIP, and NCT dataset; average ACC and CI are reported.}
    \label{tab:lc_paip_nct_acc}
\end{table}

\subsubsection{Test with Large Shot Number}
\label{app:larger_shot}
We further test the trained models with a larger shot number $k$. The result is shown in \tabref{tab:lc_paip_nct_f1_k10}

\begin{table}[ht]
    \centering
    \scalebox{0.85}{
    \begin{tabular}{c|c|c|c|c|c}
    \hline
        pretraining method & NC & LR & RC & LR+LA & RC+LA \\
    \hline
    \multicolumn{6}{c}{10-shot 5-way on LC } \\
    \hline
    ImageNet (FSP) & $78.76 \pm 0.94$  & $78.92 \pm 0.92$ & $80.45 \pm 0.87$ & $82.25 \pm 0.83$  & $80.20 \pm 0.89$ \\
    SimSiam  & $88.52 \pm 0.55$ & $87.20 \pm 0.58$ & $87.73 \pm 0.56$ & $91.62 \pm 0.46$ & $91.88 \pm 0.47$ \\
    SimCLR  & $87.02 \pm 0.64$ & $86.26 \pm 0.64$ & $85.61 \pm 0.72$ & $90.28 \pm 0.52$ & $89.60 \pm 0.58$ \\
    FSP-Patch  & $88.41 \pm 0.53$ & $88.64 \pm 0.52$ & $89.15 \pm 0.51$ & $90.49 \pm 0.50$ & $89.88 \pm 0.54$ \\
    FACILE-SupCon  & $87.58 \pm 0.56$ & $87.00 \pm 0.61$ & $87.16 \pm 0.63$ & $90.75 \pm 0.49$ & $91.02 \pm 0.49$ \\
    FACILE-FSP   & $\mathbf{92.81 \pm 0.42}$ & $\mathbf{92.33 \pm 0.44}$ & $\mathbf{92.44 \pm 0.43}$ & $\mathbf{93.07 \pm 0.43}$ & $\mathbf{92.35 \pm 0.48}$ \\ 
    \hline
    \multicolumn{6}{c}{10-shot 3-way on PAIP } \\
    \hline
    ImageNet (FSP)  & $65.36 \pm 0.91$ & $65.17 \pm 1.00$ & $65.40 \pm 0.99$ & $66.52 \pm 0.81$ & $64.45 \pm 0.81$ \\
    SimSiam    & $67.19 \pm 0.88$ & $67.35 \pm 0.98$ & $68.55 \pm 0.94$ & $70.88 \pm 0.77$ & $70.62 \pm 0.77$ \\
    SimCLR   & $65.77 \pm 0.85$ & $66.70 \pm 0.91$ & $67.01 \pm 0.91$ & $68.41 \pm 0.79$ & $66.96 \pm 0.82$ \\
    FSP-Patch  & $68.50 \pm 0.82$ & $69.12 \pm 0.85$ & $69.39 \pm 0.85$ & $70.13 \pm 0.75$ &  $68.25 \pm 0.76$ \\
    FACILE-SupCon  & $69.79 \pm 0.85$ & $69.78 \pm 0.90$ & $70.36 \pm 0.88$ & $73.53 \pm 0.73$ & $72.53 \pm 0.73$  \\
    FACILE-FSP   & $\mathbf{71.93 \pm 0.82}$ & $\mathbf{73.02 \pm 0.84}$ & $\mathbf{73.54 \pm 0.80}$ & $\mathbf{74.19 \pm 0.71}$ & $\mathbf{72.70 \pm 0.72}$ \\
    \hline
        \multicolumn{6}{c}{10-shot 9-way on NCT } \\
    \hline
    ImageNet (FSP)  & $78.76 \pm 0.94$ & $78.92 \pm 0.92$ & $80.45 \pm 0.87$ & $82.25 \pm 0.83$ & $80.20 \pm 0.89$ \\
    SimSiam    & $82.92 \pm 0.91$ & $83.42 \pm 0.89$ & $84.76 \pm 0.81$ & $87.66 \pm 0.72$ & $88.12 \pm 0.69$ \\
    SimCLR   & $80.34 \pm 0.96$ & $81.67 \pm 0.90$ & $83.09 \pm 0.84$ & $85.96 \pm 0.76$ & $86.82 \pm 0.72$ \\
    FSP-Patch   & $83.36 \pm 0.77$ & $84.05 \pm 0.74$ & $85.93 \pm 0.65$ & $87.15 \pm 0.62$ & $86.05 \pm 0.63$ \\
    FACILE-SupCon   & $83.05 \pm 0.83$ & $83.84 \pm 0.80$ & $84.95 \pm 0.73$ & $88.29 \pm 0.60$ & $88.58 \pm 0.61$ \\
    FACILE-FSP   & $\mathbf{89.55 \pm 0.50}$ & $\mathbf{90.57 \pm 0.47}$ & $\mathbf{91.28 \pm 0.45}$ &  $\mathbf{91.43 \pm 0.45}$ & $\mathbf{90.77 \pm 0.47}$ \\
    \hline
    \end{tabular}
    }
    \caption{Test result on LC, PAIP, and NCT dataset with shot number 10; average F1 and CI are reported.}
    \label{tab:lc_paip_nct_f1_k10}
\end{table}

% \begin{figure}
%     \centering
%     \includegraphics[width=0.5\textwidth]{Figs/appendix/LearningCurve/FACILE_FSP_WSI.pdf}
%     \caption{Learning curves of FACILE-FSP model.}
%     \label{fig:facile_fsp_wsi_learning_curve}
% \end{figure}

% \begin{figure}
%     \centering
%     \includegraphics[width=0.5\textwidth]{Figs/appendix/LearningCurve/SupCE_tile.pdf}
%     \caption{Learning curves of FSP-Patch model.}
%     \label{fig:facile_fsp_tile_learning_curve}
% \end{figure}

% \begin{figure}
% \centering
% \begin{subfigure}{.5\textwidth}
%   \centering
%   \includegraphics[width=.4\linewidth]{Figs/appendix/LearningCurve/FACILE_FSP_WSI.pdf}
%   \caption{A subfigure}
%   \label{fig:sub1}
% \end{subfigure}%
% \begin{subfigure}{.5\textwidth}
%   \centering
%   \includegraphics[width=.4\linewidth]{Figs/appendix/LearningCurve/SupCE_tile.pdf}
%   \caption{A subfigure}
%   \label{fig:sub2}
% \end{subfigure}
% \caption{Learning curves of FACILE-FSP model and FSP-Patch model. The mean F1 score and CI of 5 few-shot models are shown with curves.}
% \label{fig:test}
% \end{figure}

\subsection{Finetune ViT-B/14 of DINO V2 on TCGA Dataset}
\label{app:finetune_dino_v2}
Similar to \secref{app:finetune_clip}, we finetune a fully-connected layer that is appended after DINO V2 \citet{oquab2023dinov2} ViT-B/14. This methodology is applied across various models to assess their performance on histopathology image datasets. By adopting the DINO V2 architecture, known for its robustness and effectiveness in visual representation learning, we aim to harness its potential for the specialized domain of histopathology. We refer interested readers to \secref{app:finetune_dino_v2_training} for details of pretraining.

Notably, our methods, FACILE-SupCon and FACILE-FSP, demonstrated markedly superior results in comparison to other baseline models when applied to histopathology image datasets as shown in \tabref{tab:dino_v2_lc_paip_nct}. This outcome highlights the effectiveness of these methods in leveraging coarse-grained labels specific to histopathology, thereby greatly enhancing the model performance of downstream tasks. Another critical insight emerged from our research: the current foundation model, DINO V2, exhibits limitations in its generalization performance on histopathology images. This suggests that while DINO V2 provides a strong starting point due to its robust visual representation capabilities, there is a clear need for further finetuning or prompt learning to optimize its performance for the unique challenges presented by histopathology datasets. This finding underscores the importance of specialized adaptation in the application of foundation models to specific domains like medical imaging.
 
\begin{table}[ht]
    \centering
    \scalebox{0.8}{
    \begin{tabular}{c|c|c|c|c|c}
    \hline
        pretraining method & NC & LR & RC & LR+LA & RC+LA \\
    \hline
    \multicolumn{6}{c}{1-shot 5-way test on LC dataset} \\
    \hline
    DINO V2 (ViT-B/14) & $ 44.82\pm 1.41$ & $ 47.51\pm 1.39$ & $ 47.63\pm 1.38$ & $ 47.36 \pm 1.39$ & $ 48.88\pm 1.44$ \\
    SimSiam & $48.79 \pm 1.37$ & $49.43 \pm 1.35$ & $48.43 \pm 1.36$ & $49.38 \pm 1.34$ & $49.50 \pm 1.34$ \\
    SimCLR  & $50.47 \pm 1.31$ & $50.52 \pm 1.33$ & $50.44 \pm 1.32$ & $51.66 \pm 1.32$ & $51.78 \pm 1.38$ \\
    FSP-Patch  & $49.73 \pm 1.41$ & $53.59 \pm 1.38$ & $53.07 \pm 1.41$ & $51.79 \pm 1.40$ & $51.27 \pm 1.43$ \\
    FACILE-SupCon & $\mathbf{56.24 \pm 1.43}$ & $\mathbf{56.51 \pm 1.41}$ & $\mathbf{55.95 \pm 1.42}$ & $\mathbf{56.29 \pm 1.43}$ & $54.07 \pm 1.44$ \\
    FACILE-FSP  & $55.67 \pm 1.40$ & $56.26 \pm 1.36$ & $55.83 \pm 1.35$ & $56.01 \pm 1.38$ & $\mathbf{55.35 \pm 1.40}$ \\
    \hline
    \multicolumn{6}{c}{5-shot 5-way test on LC dataset} \\
    \hline
    DINO V2 (ViT-B/14) & $66.12 \pm 0.98$ & $64.71 \pm 1.12$ & $66.36 \pm 1.10$ & $72.95 \pm 0.93$ & $75.11 \pm 0.91$ \\
    SimSiam  & $67.51 \pm 0.96$ & $64.99 \pm 1.05$  & $65.39 \pm 1.05$  & $70.30 \pm 0.93$  & $71.19 \pm 0.93$ \\
    SimCLR & $70.10 \pm 0.92$  & $69.28 \pm 0.96$  & $69.18 \pm 0.97$  & $72.99 \pm 0.92$  & $72.91 \pm 0.94$ \\ 
    FSP-Patch  &  $71.97 \pm 0.96$ & $71.11 \pm 1.04$  & $71.19 \pm 1.03$  &  $73.96 \pm 0.94$ &  $73.20 \pm 0.96$ \\
    FACILE-SupCon & $75.58 \pm 0.88$ & $74.26 \pm 0.94$ & $73.20 \pm 0.95$ & $75.81 \pm 0.90$ & $74.34 \pm 0.96$ \\
    FACILE-FSP  & $\mathbf{75.86 \pm 0.86}$  & $\mathbf{74.64 \pm 0.89}$  & $\mathbf{74.12 \pm 0.93}$  &  $\mathbf{76.17 \pm 0.88}$  & $\mathbf{75.08 \pm 0.95}$  \\
    \hline
        \multicolumn{6}{c}{1-shot 3-way test on PAIP dataset} \\
    \hline
    DINO V2 (ViT-B/14) & $41.51 \pm 1.27$ & $44.37 \pm 1.26$ & $44.28 \pm 1.25$ & $42.43 \pm 1.27$ & $42.78 \pm 1.27$ \\
    SimSiam & $49.42 \pm 1.28$  & $48.07 \pm 1.35$  & $48.44 \pm 1.36$  & $48.76 \pm 1.33$  &  $46.48 \pm 1.37$ \\
    SimCLR  & $48.60 \pm 1.19$ & $48.76 \pm 1.25$ & $47.98 \pm 1.26$ & $48.94 \pm 1.23$ &  $47.20 \pm 1.26$ \\
    FSP-Patch  &  $46.09 \pm 1.17$ & $47.44 \pm 1.18$  &  $48.09 \pm 1.19$ &  $46.76 \pm 1.18$ &  $43.68 \pm 1.22$ \\
    FACILE-SupCon & $\mathbf{51.97 \pm 1.18}$ & $\mathbf{52.25 \pm 1.22}$ & $\mathbf{51.80 \pm 1.22}$ & $51.36 \pm 1.22$ & $\mathbf{50.24 \pm 1.23}$ \\
    FACILE-FSP  & $51.34 \pm 1.16$ &  $51.18 \pm 1.19$ & $51.51 \pm 1.19$ & $\mathbf{51.50 \pm 1.16}$  &  $49.77 \pm 1.22$ \\
    \hline
    \multicolumn{6}{c}{5-shot 3-way test on PAIP dataset} \\
    \hline
    DINO V2 (ViT-B/14) & $57.59 \pm 1.07$ & $58.19 \pm 1.10$ & $59.37 \pm 1.07$ & $61.84 \pm 0.85$ & $60.81 \pm 0.86$ \\
    SimSiam  & $61.56 \pm 0.97$  & $62.52 \pm 1.01$ & $62.81 \pm 1.01$ & $64.40 \pm 0.86$  &  $62.44 \pm 0.93$  \\
    SimCLR & $62.20 \pm 0.93$ & $61.78 \pm 0.99$ & $63.20 \pm 0.97$ & $63.38 \pm 0.86$ & $63.03 \pm 0.88$ \\
    FSP-Patch  &  $63.77 \pm 0.88$ & $63.85 \pm 0.94$  &  $63.85 \pm 0.93$ &  $63.61 \pm 0.85$ &  $60.91 \pm 0.87$ \\
    FACILE-SupCon & $\mathbf{67.16 \pm 0.84}$ & $67.29 \pm 0.89$ & $66.88 \pm 0.90$ & $\mathbf{67.61 \pm 0.85}$ & $\mathbf{66.34 \pm 0.84}$ \\
    FACILE-FSP  & $67.14 \pm 0.85$  & $\mathbf{67.67 \pm 0.84}$  & $\mathbf{67.54 \pm 0.86}$ &  $67.12 \pm 0.81$ &  $66.05 \pm 0.83$ \\
    \hline
        \multicolumn{6}{c}{1-shot 9-way test on NCT dataset} \\
    \hline
    DINO V2 (ViT-B/14) & $56.03 \pm 1.62$ & $59.11 \pm 1.57$ & $60.13 \pm 1.55$ & $58.71 \pm 1.57$ & $59.06 \pm 1.55$ \\
    SimSiam & $62.60 \pm 1.45$ & $61.89 \pm 1.50 $ & $61.90 \pm 1.51$ & $ 62.27 \pm 1.47 $ & $ 61.05 \pm 1.44 $ \\
    SimCLR & $65.43 \pm 1.43$ & $64.18 \pm 1.44$ & $64.15 \pm 1.46$ & $64.83 \pm 1.43$ & $62.69 \pm 1.38$ \\
    FSP-Patch  & $65.22 \pm 1.49$ & $65.93 \pm 1.41$ & $65.94 \pm 1.40$  & $65.26 \pm 1.45$ &  $62.66 \pm 1.46$ \\
    FACILE-SupCon & $71.55 \pm 1.36$ & $70.36 \pm 1.37$ & $70.52 \pm 1.35$ & $71.05 \pm 1.35$ & $\mathbf{68.85 \pm 1.40}$ \\
    FACILE-FSP  &  $\mathbf{72.05 \pm 1.34}$ & $\mathbf{70.70 \pm 1.35}$ & $\mathbf{70.77 \pm 1.34}$ & $\mathbf{71.14 \pm 1.34}$ &  $68.03 \pm 1.40$\\
    \hline
    \multicolumn{6}{c}{5-shot 9-way test on NCT dataset} \\
    \hline
    DINO V2 (ViT-B/14) & $76.85 \pm 0.98$ & $76.51 \pm 1.02$ & $78.67 \pm 0.94$ & $82.20 \pm 0.82$ & $82.75 \pm 0.83$ \\
    SimSiam  & $80.81 \pm 0.85 $ & $80.06 \pm 0.87$ & $81.55 \pm 0.85$ & $83.18 \pm 0.80$ & $82.39 \pm 0.83$ \\
    SimCLR & $82.87 \pm 0.80$ & $81.91 \pm 0.82$ & $82.86 \pm 0.80$ & $83.92 \pm 0.77$ & $82.89 \pm 0.79$ \\
    FSP-Patch  & $83.63 \pm 0.83$ & $83.49 \pm 0.80$ & $84.34 \pm 0.78$ & $85.32 \pm 0.75$ & $83.03 \pm 0.79$ \\
    FACILE-SupCon  & $87.74 \pm 0.64$ & $87.00 \pm 0.64$ & $87.38 \pm 0.62$ & $87.82 \pm 0.63$ & $86.15 \pm 0.69$ \\
    FACILE-FSP  & $ \mathbf{87.93 \pm 0.65}$ & $\mathbf{87.52 \pm 0.65}$ & $\mathbf{87.72 \pm 0.62}$ & $\mathbf{88.01 \pm 0.64}$ & $\mathbf{86.46 \pm 0.70}$ \\
    \hline
    \end{tabular}
    }
    \caption{Test result on LC, PAIP, and NCT dataset with ViT-B/14 from DINO V2; average F1 and CI are reported.}
    \label{tab:dino_v2_lc_paip_nct}
    % \vspace{-2mm}
\end{table}

\subsection{Benefits of Pretraining on Large Pathology Datasets}
\label{app:benefits_of_large_dataset}
In order to demonstrate the advantages of pretraining on large pathology datasets, we compare the performance of models pretrained on TCGA datasets with those pretrained on NCT datast, which are also studied in \citet{yang2022towards}. 

The SimSiam model is trained for 100 epochs. SGD optimizer is used with learning rate of 0.01, weight decay of 0.0001, momentum of 0.9, and cosine learning rate decay. The batch size is 55. 

For MoCo v3, similar to \citep{chen2021empirical,yang2022towards}, LARS optimizer \citep{you2017large} was used with an initial learning rate of $0.3$, weight decay of $1.5e-6$, the momentum of $0.9$, and cosine decay schedule. MoCo v3 was trained with a batch size of 256 for 200 epochs. 

The FSP model with simple augmentation follows the setting of \citet{yang2022towards}. SGD optimizer with learning rate of 0.5, momentum of 0.9 and weight decay of 0 are used. A large batch size is used 512. The model is trained for 100 epochs with ``step decay'' schedule. The learning rate multiplied by 0.1 at 30, 60, and 90 epochs respectively. 
The FSP model with strong augmentation was trained for 50 epochs. The batch size is set to 64. The SGD is used with a learning rate of 0.03, momentum of 0.9, weight decay of 0.0001, and the cosine schedule. The model is trained for 50 epochs. 

The SupCon model is trained with trained for 100 epochs. The batch size is set to 64. The SGD optimizer is used with a learning rate of 0.01, momentum of 0.9, weight decay of 0.0001, and the cosine schedule.

\tabref{tab:pretrain_nct_lc_paip} shows the performance of the pretrained models on the LC and PAIP dataset with shot numbers 1 or 5. Notably, the best-performing models on the two test datasets exhibit a significant performance gap compared to the best models pretrained on TCGA datasets as depicted in \tabref{tab:lc_paip_nct}. 

\begin{table}[ht]
    \centering
    \scalebox{0.75}{
    \begin{tabular}{c|c|c|c|c|c}
    \hline
        pretraining method & NC & LR & RC & LR+LA & RC+LA \\
    \hline
     \multicolumn{6}{c}{1-shot 5-way test on LC dataset} \\
     \hline
   SimSiam  & $59.30 \pm 1.31$ & $58.67 \pm 1.41$ & $58.58 \pm 1.40$ & $59.66 \pm 1.35$ & $59.85 \pm 1.35$ \\
     MoCo v3 (\citep{yang2022towards}) & $59.38 \pm 1.62$ & $59.39 \pm 1.68$ & $59.46 \pm 1.68$ & $60.15 \pm 1.59$ & $60.54 \pm 1.58$ \\
     FSP (simple aug; \citep{yang2022towards}) & $51.42 \pm 1.59$ & $46.06 \pm 1.88$ & $46.33 \pm 1.86$ & $50.53 \pm 1.65$ & $51.00 \pm 1.65$ \\
    FSP (strong aug) & $\mathbf{68.00 \pm 1.29}$ & $\mathbf{66.17 \pm 1.41}$ & $\mathbf{66.18 \pm 1.46}$ &  $\mathbf{68.39 \pm 1.34}$  & $\mathbf{68.02 \pm 1.40}$ \\
     SupCon & $64.48 \pm 1.33$ & $63.52 \pm 1.42$ & $63.84 \pm 1.40$ & $65.43 \pm 1.33$ & $65.98 \pm 1.38$ \\
    \hline
    \multicolumn{6}{c}{5-shot 5-way test on LC dataset} \\
    \hline
       SimSiam  & $76.21\pm 0.87$ & $74.05\pm 1.10$ & $74.59\pm 1.10$ & $77.87\pm 0.87$ & $76.03\pm 0.94$\\
       %SimCLR & $82.09 \pm 0.78$ & $78.54 \pm 1.07$ & $78.19 \pm 1.13$ & $83.36 \pm 0.76$ & $82.35 \pm 0.88$ \\
        MoCo v3 (\citep{yang2022towards}) & $72.82\pm 1.25$ & $70.29\pm 1.43$ & $71.31\pm 1.40$ & $78.72\pm 1.00$ & $79.71\pm 0.95$\\ 
        FSP (simple aug; \citep{yang2022towards}) & $56.44\pm 1.50$ & $52.27\pm 1.81$ & $55.62 \pm 1.74$ & $63.47\pm 1.37$ & $63.47\pm 1.46$ \\
        FSP (strong aug) & $\mathbf{83.53 \pm 0.79}$ & $\mathbf{80.81 \pm 1.01}$ & $\mathbf{80.27 \pm 1.08}$ & $\mathbf{85.57 \pm 0.77}$ & $\mathbf{84.06 \pm 0.89}$ \\
        SupCon & $81.51 \pm 0.85$ & $78.77 \pm 1.03$ & $78.65 \pm 1.08$ & $83.51 \pm 0.84$ & $83.31 \pm 0.91$ \\
        \hline
             \multicolumn{6}{c}{1-shot 3-way test on PAIP dataset} \\
     \hline
   SimSiam  & $37.13 \pm 1.14$ & $38.26 \pm 1.13$ & $37.93 \pm 1.15$ & $38.00 \pm 1.12$ & $38.67 \pm 1.12$ \\
     MoCo v3 (\citep{yang2022towards})  & $43.17 \pm 1.26$ & $42.48 \pm 1.30$ & $43.02 \pm 1.31$ & $43.55 \pm 1.28$ & $44.57 \pm 1.28$ \\
     FSP (simple aug; \citep{yang2022towards})  & $37.15 \pm 1.07$ & $36.69 \pm 1.13$ & $37.39 \pm 1.08$ & $37.40 \pm 1.07$ & $35.28 \pm 1.09$ \\
    FSP (strong aug)  & $47.67 \pm 1.18$ & $48.44 \pm 1.19$ & $48.16 \pm 1.21$ & $48.27 \pm 1.17$ & $\mathbf{49.38 \pm 1.19}$ \\
     SupCon  & $\mathbf{48.45 \pm 1.19}$ & $\mathbf{49.29 \pm 1.20}$ & $\mathbf{48.97 \pm 1.22}$ & $\mathbf{49.47 \pm 1.20}$ & $48.53 \pm 1.20$ \\
    \hline
    \multicolumn{6}{c}{5-shot 3-way test on PAIP dataset} \\
    \hline
       SimSiam   & $47.52 \pm 1.00$ & $48.12 \pm 1.10$ & $47.04 \pm 1.11$ & $52.70 \pm 0.95$ & $54.51 \pm 1.00$ \\
       %SimCLR  &  &  &  &  &  \\
        MoCo v3 (\citep{yang2022towards})  & $55.43 \pm 1.00$ & $54.23 \pm 1.09$ & $54.05 \pm 1.09$ & $56.07 \pm 0.92$ & $55.73 \pm 0.93$ \\
        FSP (simple aug; \citep{yang2022towards})  & $44.98 \pm 0.95$ & $45.13 \pm 0.96$ & $45.30 \pm 0.96$ & $44.34 \pm 0.87$ & $44.03 \pm 0.88$ \\
        FSP (strong aug)  & $62.00 \pm 0.88$ & $62.48 \pm 0.97$ & $62.04 \pm 0.98$ & $\mathbf{64.82 \pm 0.86}$ &  $\mathbf{64.60 \pm 0.87}$ \\
        SupCon  & $\mathbf{63.62 \pm 0.91}$ & $\mathbf{64.38 \pm 0.96}$ & $\mathbf{63.61 \pm 1.00}$ & $64.37 \pm 0.87$ & $64.28 \pm 0.88$ \\
        \hline
    \end{tabular}
    }
    \caption{pretraining on NCT dataset and testing on LC and PAIP dataset; average F1 and CI are reported.}
    \label{tab:pretrain_nct_lc_paip}
\end{table}

\subsection{Pretrain on TCGA and GTEx with Patch Size 1,000X1,000}
\label{app:pdac_result}
We train the models using TCGA patches of size $1,000\times 1,000$, which are extracted from 20X magnification and resized to $224\times 224$. Subsequently, the pretrained models are evaluated on PDAC datasets, and the corresponding test performance is presented in Figure \ref{tab:tcga_pdac}. Notably, for shot number of 1 and 5, our model significantly outperforms other models, demonstrating a substantial performance margin.

% \begin{table}[ht]
%     \centering
% \begin{tabular}{ p{3.5cm}||p{2cm}|p{3.5cm}|p{2cm}  }
%  \hline
%  Few-shot Learning & SimSiam & FACILE-FSP & FSP-Patch\\
%  \hline
%    \multicolumn{4}{c}{1-shot 5-way test on LC25000 dataset} \\
%  \hline
%   NearestCentroid &   &   &    \\
%  LogisticRegression &   &   &  \\
%  RidgeClassiﬁer &   &   &   \\
%  LogisticRegression+LA &   &   &    \\
%  RidgeClassiﬁer+LA &   &   &   \\
%  \hline
%   \multicolumn{4}{c}{5-shot 5-way test on LC25000 dataset} \\
%  \hline
%  NearestCentroid &  &  &   \\
%  LogisticRegression &  &  & \\
%  RidgeClassiﬁer &  &  &  \\
%  LogisticRegression+LA &  &  &   \\
%  RidgeClassiﬁer+LA &  &  &  \\
%  \hline
% \end{tabular}
%     \caption{Models pretrained on TCGA; tested on PDAC dataset}
%     \label{tab:tcga_pdac}
% \end{table}

\begin{table}[ht]
    \centering
    \scalebox{0.8}{
    \begin{tabular}{c|c|c|c|c|c}
    \hline
        pretraining method & NC & LR & RC & LR+LA & RC+LA \\
    \hline
    \multicolumn{6}{c}{1-shot 5-way test } \\
    \hline
    ImageNet (FSP) & $29.57 \pm 1.07$ & $31.32 \pm 1.09$ & $31.16 \pm 1.07$ & $30.88 \pm 1.08$ & $30.14 \pm 1.08$ \\
    SimSiam & $30.48 \pm 1.08$  & $30.18 \pm 1.12$  & $30.19 \pm 1.13$  &  $30.41 \pm 1.08$ &  $31.13 \pm 1.10$ \\
    SimCLR & $30.79 \pm 1.08$  &  $30.93 \pm 1.13$ &  $30.78 \pm 1.12$ & $31.33 \pm 1.08$  & $31.22 \pm 1.07$  \\
    FSP-Patch  & $34.04 \pm 1.16$  & $33.99 \pm 1.20$  &  $33.69 \pm 1.20$ & $34.29 \pm 1.15$  &  $34.99 \pm 1.16$ \\
    FACILE-SupCon & $33.72 \pm 1.18$ & $32.13 \pm 1.26$ & $31.93 \pm 1.26$ & $34.54 \pm 1.17$ & $34.30 \pm 1.20$ \\
    FACILE-FSP  & $\mathbf{37.36 \pm 1.16}$  & $\mathbf{36.07 \pm 1.23}$  &  $\mathbf{36.93 \pm 1.21}$ & $\mathbf{36.79 \pm 1.19}$  & $\mathbf{36.81 \pm 1.18}$ \\
    \hline
    \multicolumn{6}{c}{5-shot 5-way test } \\
    \hline
    ImageNet (FSP) & $41.83 \pm 0.96$ & $41.30 \pm 1.10$ & $41.08 \pm 1.08$ & $42.38 \pm 0.94$ & $41.29 \pm 0.93$ \\
    SimSiam  & $40.15 \pm 1.03$ & $37.29 \pm 1.21$ & $37.43 \pm 1.21$ & $41.87 \pm 1.00$ & $42.70 \pm 1.01$ \\
    SimCLR & $40.30 \pm 1.04$ & $38.74 \pm 1.19$ & $39.02 \pm 1.16$ & $40.98 \pm 0.96$ &  $40.90 \pm 0.98$ \\
    FSP-Patch  & $44.26 \pm 1.10$ & $42.99 \pm 1.20$ & $43.69 \pm 1.12$ & $46.32 \pm 0.97$ & $46.69 \pm 0.96$ \\
    FACILE-SupCon  & $42.51 \pm 1.12$ & $40.27 \pm 1.25$ & $40.75 \pm 1.24$ & $45.10 \pm 0.99$ & $45.73 \pm 0.99$ \\
    FACILE-FSP  & $\mathbf{48.21 \pm 1.04}$ & $\mathbf{47.62 \pm 1.12}$ & $\mathbf{47.94 \pm 1.08}$ & $\mathbf{48.84 \pm 0.95}$ &  $\mathbf{48.37 \pm 0.95}$ \\
    \hline
    \end{tabular}
    }
     \caption{Models pretrained on TCGA and tested on PDAC dataset; average F1 and CI are reported.}
     \label{tab:tcga_pdac}
     %\vspace{-3mm}
\end{table}

%\subsubsection{pretrain on GTEx with patch size 1000X1000}
Similarly, we train the models using GTEx patches with dimensions of $1,000\times 1,000$. The patches are extracted from 20X magnification and resized to $224\times 224$. The pretrained models are tested on PDAC datasets, revealing similar outcomes, as illustrated in Table \ref{tab:gtex_pdac}.

% \begin{table}[ht]
%     \centering
% \begin{tabular}{ p{3.5cm}||p{2cm}|p{3.5cm}|p{2cm}  }
%  \hline
%  Few-shot Learning & SimSiam & FACILE-FSP & FSP-Patch\\
%  \hline
%    \multicolumn{4}{c}{1-shot 5-way test on LC25000 dataset} \\
%  \hline
%   NearestCentroid &   &   &    \\
%  LogisticRegression &   &   &  \\
%  RidgeClassiﬁer &   &   &   \\
%  LogisticRegression+LA &   &   &    \\
%  RidgeClassiﬁer+LA &   &   &   \\
%  \hline
%   \multicolumn{4}{c}{5-shot 5-way test on LC25000 dataset} \\
%  \hline
%  NearestCentroid &  &  &   \\
%  LogisticRegression &  &  & \\
%  RidgeClassiﬁer &  &  &  \\
%  LogisticRegression+LA &  &  &   \\
%  RidgeClassiﬁer+LA &  &  &  \\
%  \hline
% \end{tabular}
%     \caption{Models pretrained on GTEx; tested on PDAC dataset}
%     \label{tab:gtex_pdac}
% \end{table}

\begin{table}[ht]
    \centering
    \scalebox{0.8}{
    \begin{tabular}{c|c|c|c|c|c}
    \hline
        pretraining method & NC & LR & RC & LR+LA & RC+LA \\
    \hline
    \multicolumn{6}{c}{1-shot 5-way test } \\
    \hline
    SimSiam & $34.78 \pm 1.18$  &  $34.57 \pm 1.25$ & $35.30 \pm 1.25$  & $35.13 \pm 1.19$  & $35.27 \pm 1.19$  \\
    SimCLR &  $33.68 \pm 1.14$ & $33.74 \pm 1.18$  & $33.69 \pm 1.17$  & $34.28 \pm 1.14$  &  $33.84 \pm 1.12$ \\
    FSP-Patch  & $31.87 \pm 1.09$  & $32.90 \pm 1.13$  & $32.53 \pm 1.11$  & $32.55 \pm 1.09$  &  $32.10 \pm 1.07$ \\
    FACILE-SupCon & $34.36 \pm 1.06$ & $34.35 \pm 1.13$ & $34.39 \pm 1.14$ & $34.70 \pm 1.07$ & $34.35 \pm 1.07$ \\
    FACILE-FSP  & $\mathbf{35.62 \pm 1.10}$  &  $\mathbf{35.51 \pm 1.15}$ & $\mathbf{35.40 \pm 1.13}$  &  $\mathbf{35.87 \pm 1.10}$ & $\mathbf{36.16 \pm 1.09}$ \\
    \hline
    \multicolumn{6}{c}{5-shot 5-way test } \\
    \hline
    SimSiam  & $46.00 \pm 1.10$ & $43.26 \pm 1.30$ & $44.19 \pm 1.26$ & $47.24 \pm 1.00$ & $\mathbf{47.85 \pm 1.00}$ \\
    SimCLR & $44.44 \pm 1.08$ & $43.40 \pm 1.19$ & $43.58 \pm 1.15$ & $44.60 \pm 0.98$ & $44.17 \pm 0.96$ \\
    FSP-Patch  & $42.09 \pm 0.99$ & $40.15 \pm 1.15$ & $40.69 \pm 1.09$ & $42.71 \pm 0.92$ & $42.66 \pm 0.90$ \\
    FACILE-SupCon  & $44.85 \pm 1.02$ & $43.65 \pm 1.15$ & $44.01 \pm 1.13$ & $46.37 \pm 0.93$ & $45.10 \pm 0.92$ \\
    FACILE-FSP  & $\mathbf{46.91 \pm 0.97}$ & $\mathbf{46.32 \pm 1.07}$ & $\mathbf{47.10 \pm 1.02}$ & $\mathbf{48.01 \pm 0.90}$ & $47.70 \pm 0.89$ \\
    \hline
    \end{tabular}
    }
    \caption{Models pretrained on GTEx and tested on PDAC dataset; average F1 and CI are reported.}
    \label{tab:gtex_pdac}
    %\vspace{-8mm}
\end{table}

\section{Datasets}
\label{app:datasets}
\subsection{GTEx Dataset}
The Genotype-Tissue Expression (GTEx) project is a pioneering initiative aimed at constructing an extensive public repository to investigate tissue-specific gene expression and regulation. The GTEx project collected samples from 54 non-diseased tissue sites across nearly 1000 individuals, with an emphasis on molecular assays such as Whole Genome Sequencing (WGS), Whole Exome Sequencing (WES), and RNA-sequencing. Additionally, the GTEx Biobank contains a plethora of unutilized samples. The GTEx portal (\url{https://gtexportal.org/home/}) provides unrestricted access to a plethora of data, including gene expression levels, quantitative trait loci (QTLs), and histology images, to aid the research community in advancing our understanding of human gene expression and its regulation.

We downloaded all the slides from the GTEx portal. The organs from which the slides are extracted are used for coarse-grained labels. We extract all the non-overlapping patches with size $1,000\times 1,000$ and only keep those with intensity in $[0.1,0.85]$ to filter out backgrounds.

\begin{figure}[ht]
    \centering
    \includegraphics[width=0.9\textwidth]{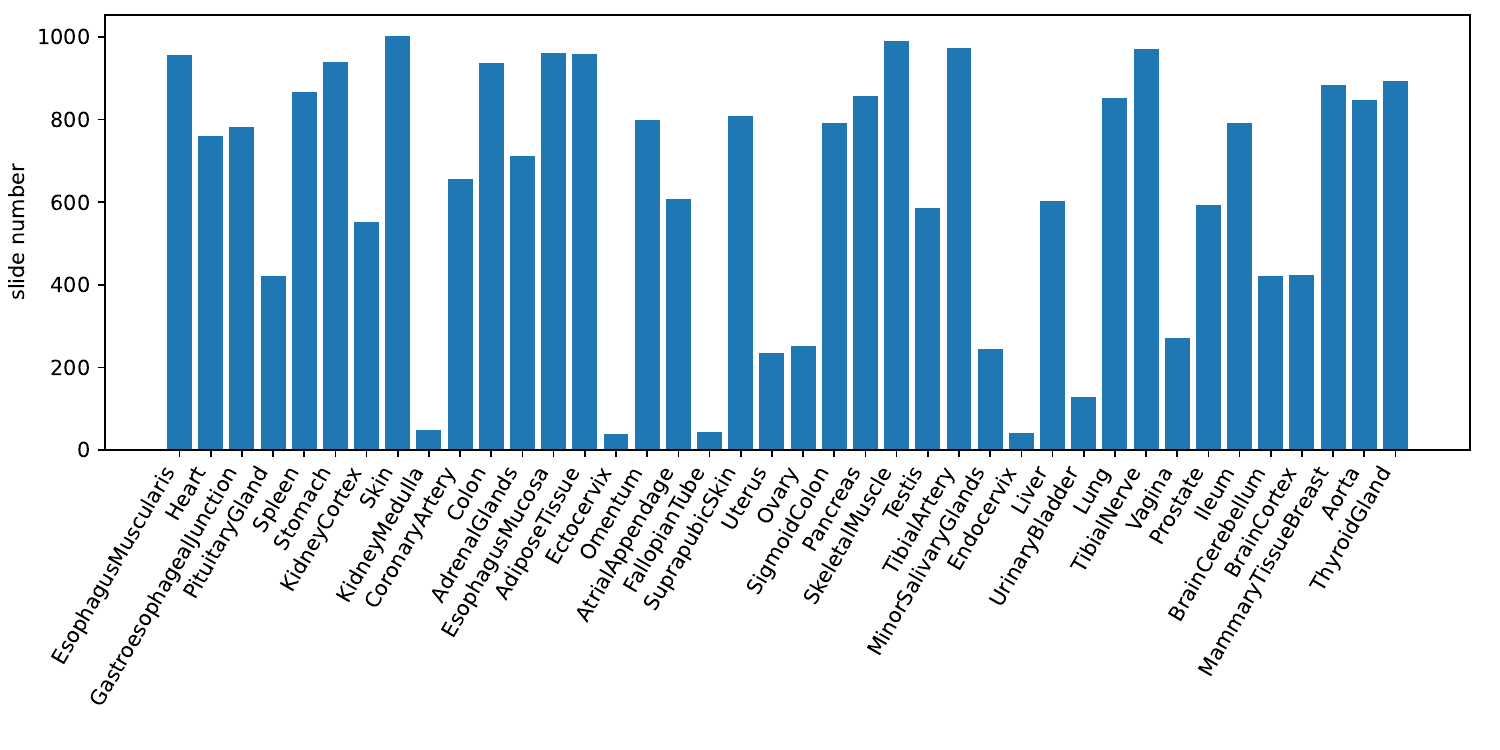}
    \caption{Slide number for each organ in GTEx}
    \label{fig:gtex_slide_num}
\end{figure}

\begin{figure}
    \centering
    \includegraphics[width=0.95\textwidth]{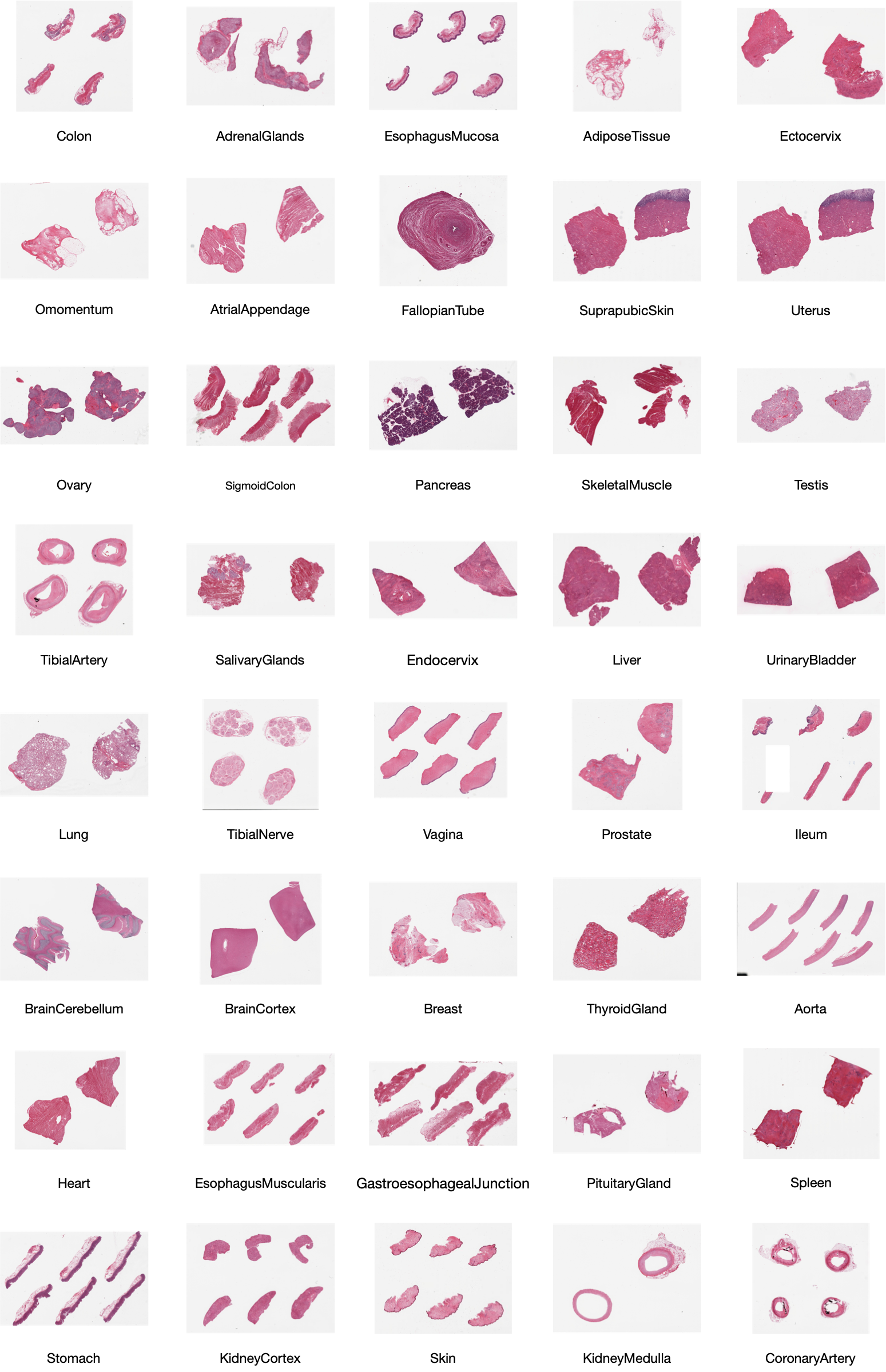}
    \caption{Randomly deleted examples from GTEx dataset}
    \label{fig:gtex_examples}
\end{figure}

The number of slides from each organ for GTEx can be found in \figref{fig:gtex_slide_num}. Thumbnails of WSI examples from the GTEx dataset can be found in \figref{fig:gtex_examples}. 

\subsection{TCGA Dataset}
The Cancer Genome Atlas (TCGA; \url{https://www.cancer.gov/ccg/research/genome-sequencing/tcga}) is a project that aims to comprehensively characterize genetic mutations responsible for cancer using genome sequencing and bioinformatics. The TCGA dataset consists of 10,825 patient samples, including gene expression, DNA methylation, copy number variation, and mutation data, histopathology data, among others \citep{cancer2008comprehensive,cancer2012comprehensive}. This large-scale dataset has enabled researchers to identify numerous genomic alterations associated with cancer and has contributed to the development of new diagnostic and therapeutic approaches. 

We downloaded all the diagnostic slides from GDC portal \url{https://portal.gdc.cancer.gov/}. The project names of the slides are used for coarse-grained labels. We extract patches at two different scales, i.e., $224\times 224$ and $1,000\times 1,000$ at 20X magnification, from all the slides.

\begin{figure}[ht]
    \centering
    \includegraphics[width=0.9\textwidth]{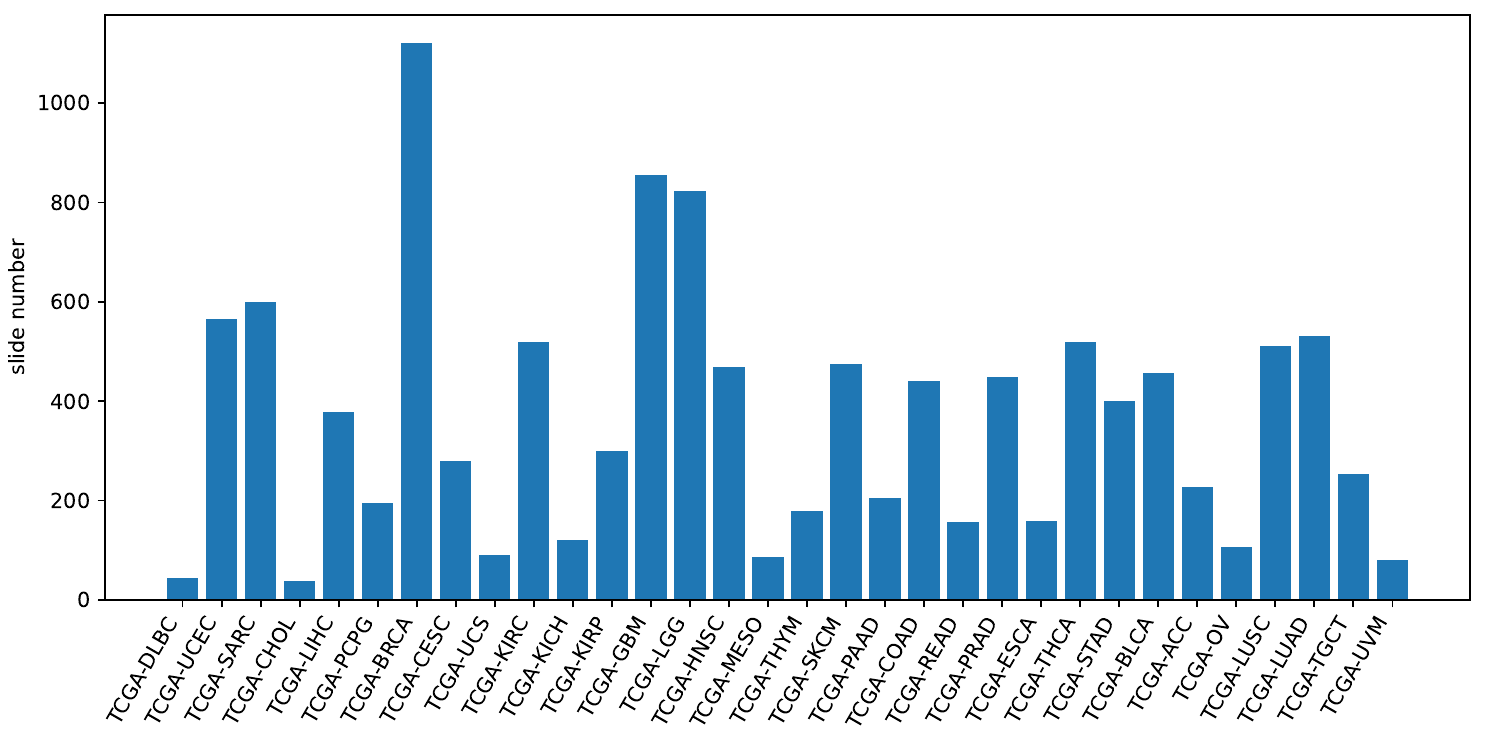}
    \caption{Slide number for each tumor in TCGA}
    \label{fig:tcga_slide_num}
\end{figure}

\begin{figure}
    \centering
    \includegraphics[width=0.95\textwidth]{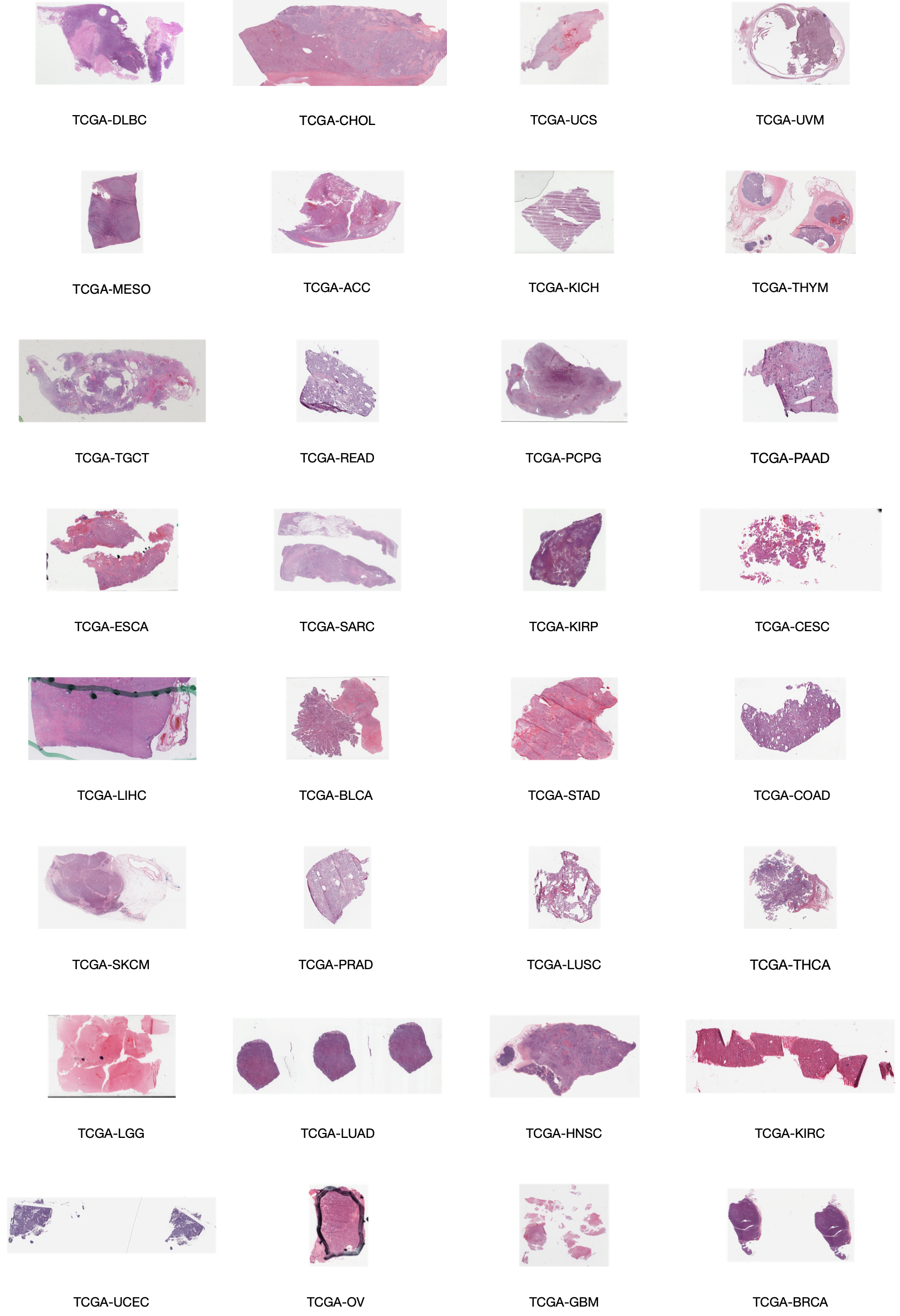}
    \caption{Randomly selected examples from TCGA dataset}
    \label{fig:tcga_examples}
\end{figure}

The number of slides from each project for TCGA can be found in \figref{fig:tcga_slide_num}. 
Thumbnails of WSI examples from TCGA dataset can be found in \figref{fig:tcga_examples}. 

\subsection{PDAC Dataset}
\begin{figure}[ht]
    \centering
    \includegraphics[width=0.5\textwidth]{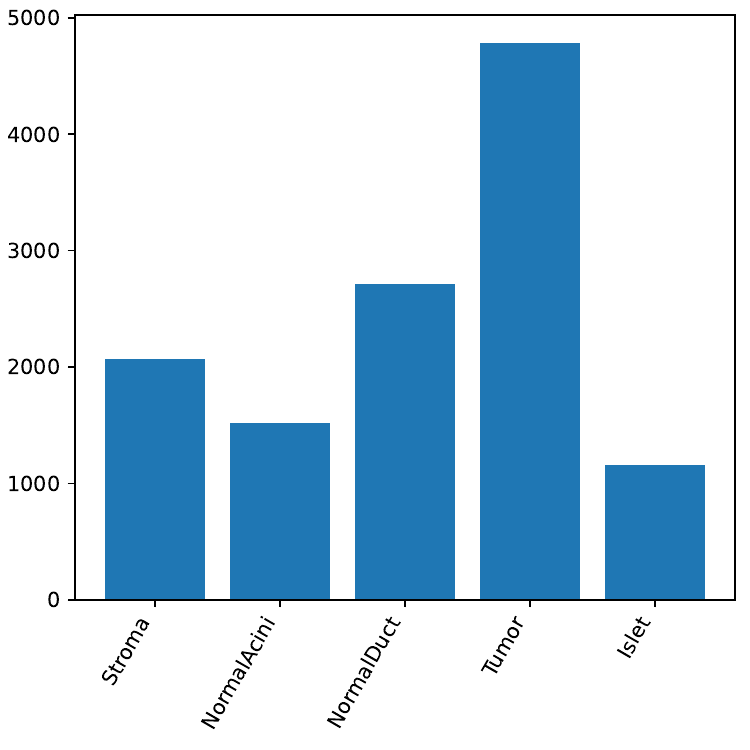}
    \caption{Patch number for each tissue for PDAC}
    \label{fig:pdac_tile_num}
\end{figure}

\begin{figure}
    \centering
    \includegraphics[width=0.9\textwidth]{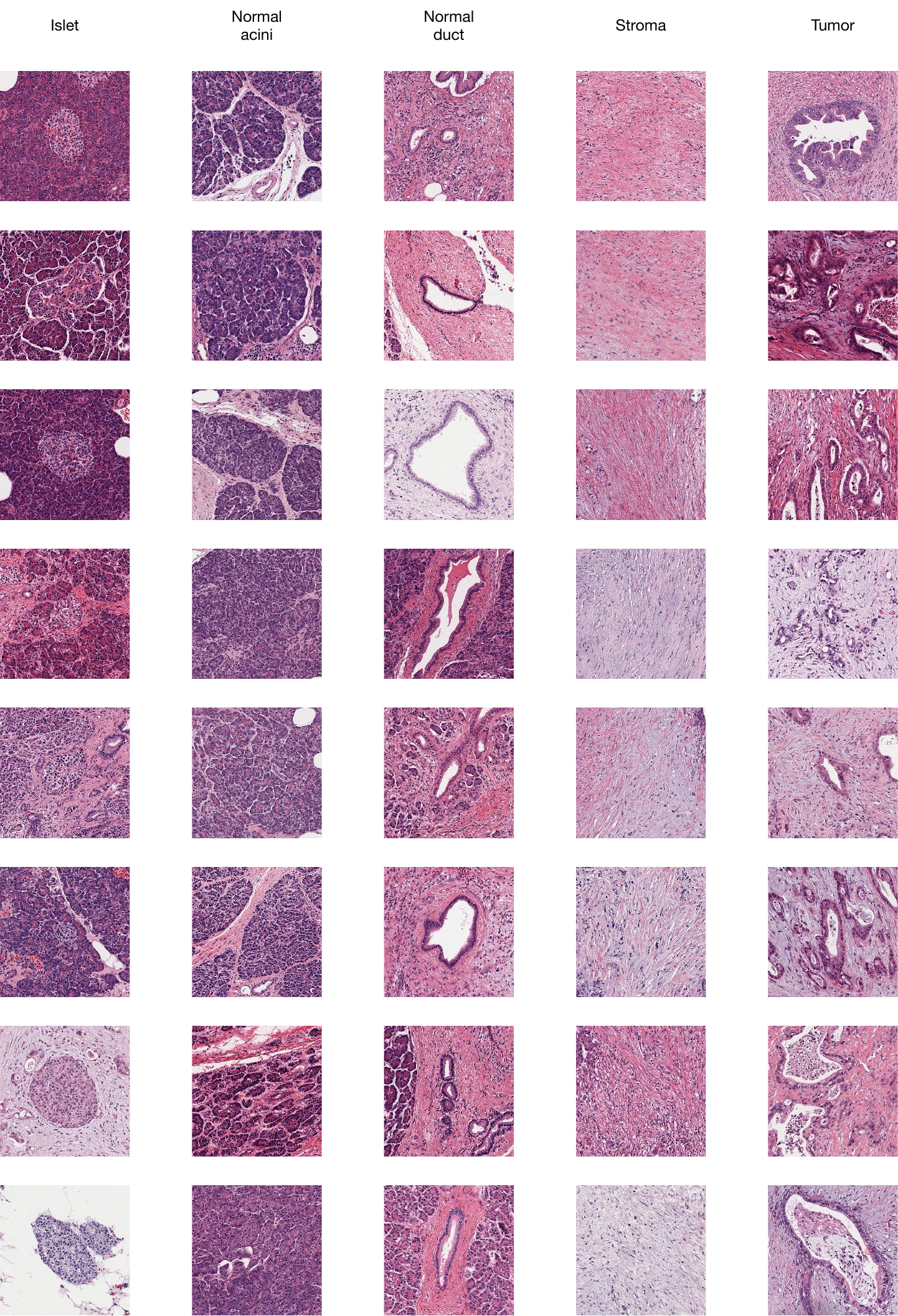}
    \caption{Randomly selected examples from each class of PDAC dataset.}
    \label{fig:pdac_examples}
\end{figure}

To address the presence of multiple tissues within certain patches, we employ a labeling strategy that involves identifying and labeling the centered tissues within these patches. To ensure annotation accuracy, each patch undergoes labeling by a minimum of two pathologists, thereby maintaining the quality of the annotations. For the specific patch numbers corresponding to each tissue in the PDAC dataset, please refer to Figure \ref{fig:pdac_tile_num}. Furthermore, examples of patches from the PDAC dataset are provided in Figure \ref{fig:pdac_examples}, offering visual illustrations of the dataset.

\begin{table}[ht]
    \centering
    \scalebox{0.85}{
    \begin{tabular}{c|c|c|c}
    \hline
        Coarse-Grained Dataset & Data type and annotation & WSI number & Extracted patch number \\
        \hline
        GTEx & slides; organs & 25,501 & 9,465,689 \\
        \multirow{2}{*}{TCGA} & \multirow{2}{*}{slides; tumors} & \multirow{2}{*}{11,638} & 10,321,273 \\
        & & & (11,588,226 w/ size 224)  \\
        \hline
        Fine-Grained Dataset & Data type and annotation & WSI number & Extracted patch number \\
    \hline
        PDAC & patches; tissues & 194 & 12,250 \\
        LC25000 & patches; tissues & 1,250 & 25,000 \\
        PAIP19 & patches; tissues & 60 & 75,000 \\
        NCT-CRC-HE-100K & patches; tissues & 86 & 100,000 \\
        \hline
    \end{tabular}
    }
    \caption{Dataset statistics}
    \label{tab:dataset}
    %\vspace{-3mm}
\end{table}

In order to validate our model on a real-world dataset, we generated WSIs of Pancreatic Ductal Adenocarcinoma (PDAC)\footnote{We will make data publicly available upon acceptance of our paper}. PDAC, a particularly aggressive and lethal form of cancer originating in the pancreatic duct cells, presents various subtypes, each with distinct morphological characteristics. These variations underscore the need for advanced automated tools to accurately characterize and differentiate between these subtypes, thereby aiding disease studies and potentially informing treatment strategies. Examples of PDAC and class distribution are detailed in \secref{app:datasets}. There are in total 12,250 annotated patches extracted from 194 slides. The patch size used for this analysis is $1,000 \times 1,000$ at a 20X magnification. Each patch was annotated into one of 5 classes (i.e., Stroma, Normal Acini, Normal Duct, Tumor, and Islet) and confirmed by at least two pathologists.

\subsection{NCT, PAIP, and LC}

We test our models on 4 datasets with fine-grained labels. These datasets are from diverse body sites. Statistics of these datasets can be found in \tabref{tab:dataset}.

NCT-CRC-HE-100K (NCT) is collected from colon \citep{kather_jakob_nikolas_2018_1214456}. It consists of 9 classes with 100K non-overlapping patches. The patch size is $224\times224$. LC25000 (LC) is collected from lung and colon sites \citep{lc25000}. It has 5 classes and each class has 5,000 patches. The patch size is $768\times 768$. We resize the patches to $224 \times 224$. PAIP19 (PAIP) is collected from liver site \citep{kim2021paip}. There are in total 50 WSIs. The WSIs are cropped into patches with size $224 \times 224$. We only keep those patches with masks and assign labels with majority voting similar to \citet{yang2022towards}. We downsample these patches to 75K patches, with 25K in each class. 

\section{Data Augmentation}\label{app:data_aug}
Two data augmentation strategies are used in this paper.
\paragraph{Simple augmentation}
Following \citet{yang2022towards}, we also used a simple augmentation policy which includes random resized cropping and horizontal flipping. In our paper, this simple augmentation policy is only used for FSP-Patch model pretraining on the NCT dataset.

\paragraph{Strong augmentation}
Following previous work \citep{grill2020bootstrap,chen2021empirical,yang2022towards}, for SimCLR and SupCon models, we used similar strong data augmentation which contains random resized cropping, horizontal flipping, horizontal flipping, color jittering \citep{wu2018unsupervised} with (brightness=0.8, contrast=0.8, saturation=0.8, hue=0.2, probability=0.8), grayscale conversion \citep{wu2018unsupervised} with (probability=0.2), Gaussian blurring \citep{chen2020simple} with (kernel size=5, min=0.1, max=2.0, probability=0.5), and polarization \citep{grill2020bootstrap} with (threshold=128, probability=0.2).

In implementing the SimSiam model, we adopted a comparable augmentation strategy, utilizing robust data augmentation techniques. Specifically, we fine-tuned parameters for color jittering, setting brightness, contrast, and saturation adjustments to 0.4, and hue to 0.1. These modifications were applied with a probability of 0.8, as informed by \citet{chen2021exploring}.

\section{Latent Augmentation}\label{app:la}
Latent augmentation (LA) was originally proposed in \citet{yang2022towards} to improve the performance of the few-shot learning system in a simple unsupervised way. The pretrained feature extractor can only transfer parts of available knowledge in the pretraining datasets by the learned weights of the feature extractor. More transferable knowledge is inherent in the pretraining data representations. 

In order to fully exploit the pretraining data, possible semantic shifts of clustered representations of the pretraining dataset are transferred to downstream tasks besides the pretrained feature extractor weights. The k-means clustering method is performed on the representations of pretraining datasets, which are generated by the pretrained feature extractor $\hat{e}$. Assume we obtain $C$ clusters after clustering. The base dictionary $\cB=\{(c_i, \Sigma_i)\}_{i=1}^{C}$ is constructed, where $c_i$ is the $i$-th cluster prototype, i.e., mean representation of all samples in the cluster and $\Sigma_i$ is the covariance matrix of the cluster. During downstream task testing, LA uses the original representation $z$ to select the closest prototype from $\cB$. We can get additive augmentation $\Tilde{z}=z+\delta$, where $\delta$ is sampled from $\cN(0,\Sigma_{i^{*}})$ and $i^{*}$ is the index of closest prototype of $z$. The classifier of the downstream tasks is then trained on both the original representations and the augmented representations.

\section{Ablation Study}
\label{app:ablation}
\subsection{Set-input Models}
\label{app:set_input_model}
Pooling architectures have been used in various set-input problems, e.g, 3D shape recognition \citep{shi2015deeppano,su2015multi}, learning the statistics of a set \citep{edwards2016towards}. \citet{vinyals2015order,ilse2018attention} pool elements in a set by a weighted average with weights computed by the attention module. \citep{zaheer2017deep,edwards2016towards} proposed to aggregate embeddings of instances, extracted using a neural network, with pooling operations (e.g., mean, sum, max). This simple method satisfies the permutation invariant property and can work with any set size. \citet{santoro2017simple} used a relational network to model all pairwise interactions of elements in a given set. \citet{lee2019set} proposed to use the Transformer \citep{vaswani2017attention} to explicitly model higher-order interactions among the instances in a set.

We evaluate three set-input models for the FACILE-FSP model: attention-based MIL pooling \citep{ilse2018attention}, Deep Set \citep{zaheer2017deep}, and Set Transformer \citep{lee2019set}. Attention-based MIL pooling uses a weighted average of instance embeddings from a set where weights are determined by a neural network. The attention-based MIL pooling corresponds to a version of attention \citep{lin2017structured,raffel2015feed}. It has been adapted by \citet{zhang2020evaluating,zhang2020evaluation,pal2021deep} in the context of H\&E images. It uses a single fully connected layer and softmax with batch normalization and ReLU activation to predict the attention weights for instances. In the Deep Set model, each instance in a set is independently fed into a neural network that takes fixed-sized inputs. The extracted features are then aggregated using a pooling operation (i.e., mean, sum, or max). The final output is obtained by further non-linear operations. The simple architecture satisfies the permutation invariant property and can work with any set size. Set Transformer adapted the Transformer model for set data. It leverages the attention mechanism \citep{vaswani2017attention} to capture interactions between instances of the input set. It applies the idea of inducing points from the sparse Gaussian process literature to reduce quadratic complexity to linear in the size of the input set. 

We train FACILE-FSP with three set-input models. The set size $a$ is set to 5. In the attention-based MIL pooling model, we implemented the simple version, and use the single fc layer with softmax to predict attention weights from ResNet18 extracted features. For Deep Set model, we use two fc layers with ReLU activation functions in between to extract instance features before set pooling. In the Set Transformer, we use 4 attention heads and 3 inducing points. 

\begin{table}[ht]
    \centering
    \scalebox{0.8}{
    \begin{tabular}{c|c|c|c|c|c}
    \hline
        set-input model & NC & LR & RC & LR+LA & RC+LA \\
    \hline
    \multicolumn{6}{c}{1-shot 5-way test on LC dataset} \\
    \hline
    Attention-based MIL pooling  & $70.53 \pm 1.32$ & $69.86 \pm 1.39$ & $69.75 \pm 1.37$ & $71.15 \pm 1.31$ & $70.31 \pm 1.34$ \\
    Deep Set & $\mathbf{77.84 \pm 1.16}$ & $\mathbf{77.56 \pm 1.16}$ & $\mathbf{77.56 \pm 1.17}$ & $\mathbf{79.16 \pm 1.09}$ & $\mathbf{77.38 \pm 1.18}$ \\
    Set Transformer & $75.09 \pm 1.30$ & $73.57 \pm 1.29$ & $73.16 \pm 1.33$ & $74.03 \pm 1.28$ & $72.88 \pm 1.34$  \\
    \hline
    \multicolumn{6}{c}{5-shot 5-way test on LC dataset} \\
    \hline
    Attention-based MIL pooling  & $88.12 \pm 0.59$ & $81.60 \pm 1.04$ & $82.51 \pm 0.97$ & $89.18 \pm 0.57$ & $88.15 \pm 0.65$ \\
    Deep Set & $90.35 \pm 0.50$ & $\mathbf{90.91 \pm 0.47}$ & $\mathbf{91.54 \pm 0.46}$ & $\mathbf{91.68 \pm 0.50}$ & $\mathbf{90.97 \pm 0.54}$ \\
    Set Transformer & $\mathbf{90.67 \pm 0.54}$ & $89.18 \pm 0.61$ & $89.02 \pm 0.63$ & $90.03 \pm 0.59$ & $88.71 \pm 0.67$ \\
    \hline
        \multicolumn{6}{c}{1-shot 3-way test on PAIP dataset} \\
    \hline
    Attention-based MIL pooling & $50.98 \pm 1.37$ & $51.93 \pm 1.35$ & $51.91 \pm 1.36$ & $51.98 \pm 1.36$ & $52.39 \pm 1.35$ \\
    Deep Set & $\mathbf{52.04 \pm 1.25}$ & $\mathbf{53.27 \pm 1.25}$ & $\mathbf{54.19 \pm 1.26}$ & $\mathbf{52.66 \pm 1.25}$ & $\mathbf{52.79 \pm 1.23}$ \\
    Set Transformer & $48.81 \pm 1.21$ & $50.08 \pm 1.24$ & $50.75 \pm 1.23$ & $50.03 \pm 1.23$ & $49.41 \pm 1.20$ \\
    \hline
    \multicolumn{6}{c}{5-shot 3-way test on PAIP dataset} \\
    \hline
    Attention-based MIL pooling & $67.04 \pm 1.00$ & $66.06 \pm 1.17$ & $66.61 \pm 1.10$ & $\mathbf{70.19 \pm 0.87}$ & $\mathbf{70.54 \pm 0.81}$ \\
    Deep Set & $\mathbf{69.42 \pm 0.85}$ & $\mathbf{69.93 \pm 0.92}$ & $\mathbf{70.52 \pm 0.87}$ & $69.96 \pm 0.84$ & $68.39 \pm 0.84$ \\
    Set Transformer & $66.61 \pm 0.91$ & $67.57 \pm 0.95$ & $67.78 \pm 0.95$ & $68.24 \pm 0.85$ & $67.20 \pm 0.86$ \\
    \hline
    \multicolumn{6}{c}{1-shot 9-way test on NCT dataset} \\
    \hline
    Attention-based MIL pooling & $60.04 \pm 1.40$ & $64.53 \pm 1.29$ & $64.81 \pm 1.31$ & $64.00 \pm 1.34$ & $66.66 \pm 1.32$ \\
    Deep Set & $\mathbf{68.21 \pm 1.30}$ & $68.17 \pm 1.31$ & $\mathbf{68.69 \pm 1.30}$ & $\mathbf{69.24 \pm 1.28}$ & $\mathbf{68.18 \pm 1.33}$ \\
    Set Transformer & $67.76 \pm 1.31$ & $\mathbf{68.52 \pm 1.30}$ & $68.55 \pm 1.28$ & $68.33 \pm 1.28$ & $67.72 \pm 1.28$ \\
    \hline
    \multicolumn{6}{c}{5-shot 9-way test on NCT dataset} \\
    \hline
    Attention-based MIL pooling & $81.94 \pm 0.75$ & $82.40 \pm 0.72$ & $84.46 \pm 0.65$ & $86.49 \pm 0.62$ & $\mathbf{87.66 \pm 0.59}$ \\
    Deep Set & $85.18 \pm 0.60$ & $85.87 \pm 0.60$ & $87.11 \pm 0.56$ & $87.06 \pm 0.61$ & $85.81 \pm 0.66$ \\
    Set Transformer & $\mathbf{86.45 \pm 0.62}$ & $\mathbf{87.74 \pm 0.59}$ & $\mathbf{87.97 \pm 0.58}$ & $\mathbf{88.00 \pm 0.59}$ & $86.92 \pm 0.61$ \\
    \hline
    \end{tabular}
    }
    \caption{Performance of FACIEL-FSP-WSI with three different set-input models; average F1 and CI are reported.}
    \label{tab:set_input_models}
\end{table}

From \tabref{tab:set_input_models}, we conclude that none of the 3 set-input models used in FACILE-FSP is consistently better than the other set-input models. The Deep Set model achieves the highest average F1 score with more tasks.

% \begin{table}[ht]
% \begin{center}
% \begin{minipage}{174pt}
% \caption{Attention module in backbone}\label{tab_attention_ablation}%
% \begin{tabular}{@{}lll@{}}
% \toprule
% Method & Attention & Test Acc (3 times)\\
% \midrule
% \multirow{2}{*}{GTEx}  & Attention-based MIL pooling & $0.933\pm 0.014$  \\
%    & Set Transformer &  $0.937\pm 0.012$ \\
% \midrule
% \multirow{2}{*}{TCGA}  & Attention-based MIL pooling  &  $0.649\pm 0.012$ \\
%  & Set Transformer & $0.611\pm 0.025$ \\
% \midrule
% \end{tabular}
% %\footnotetext{Result on TCGA dataset}

% \end{minipage}
% \end{center}
% \end{table}

\subsection{Learning Curve}
\label{app:learning_curve}

\begin{figure}[!ht]%
    \centering
    \subfloat[\centering FACILE-FSP model]{{\includegraphics[width=.3\textwidth]{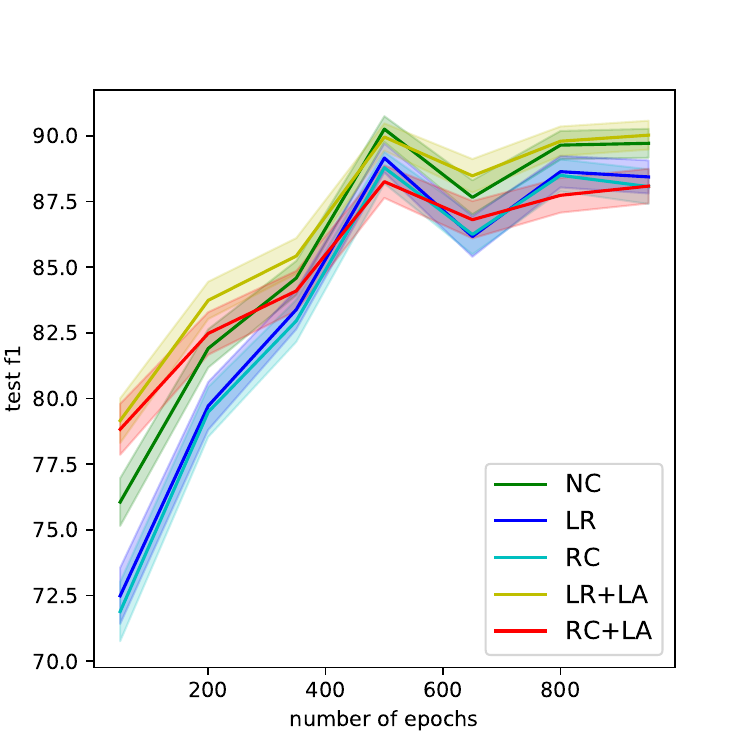} }}%
    \quad
    \subfloat[\centering FSP-Patch model]{{\includegraphics[width=.3\textwidth]{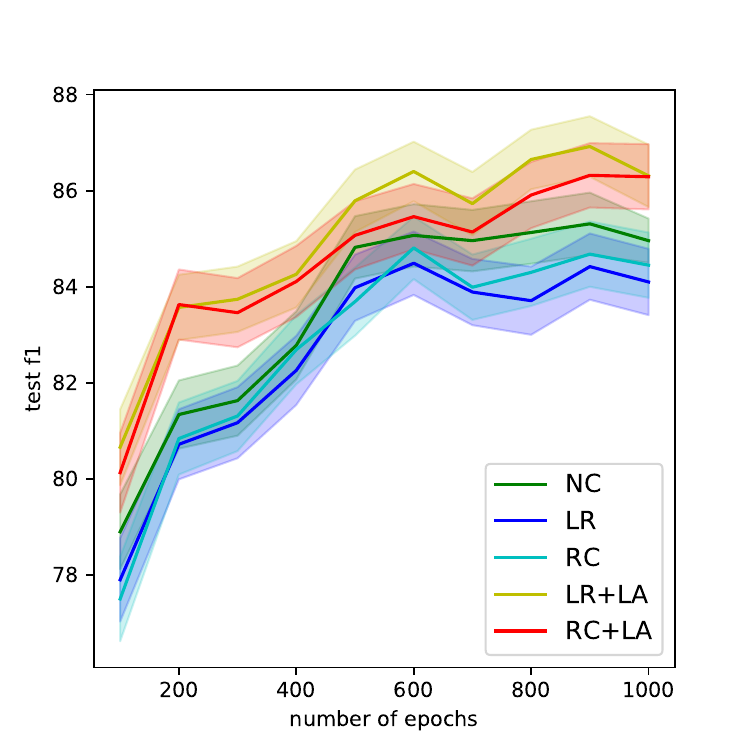} }}%
    \quad
    \subfloat[\centering SimSiam model]
    {{\includegraphics[width=0.3\textwidth]{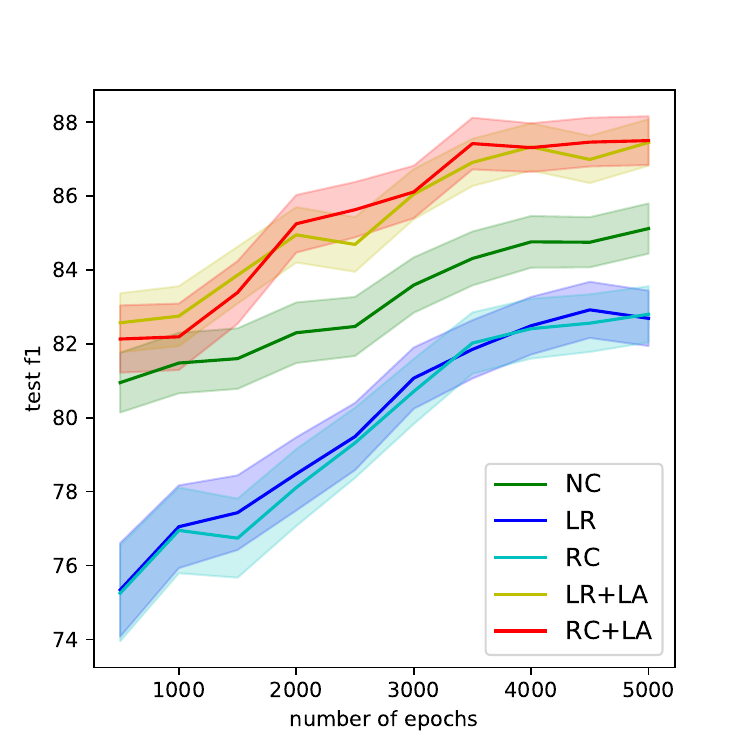}}}
    \caption{Learning curves of FACILE-FSP model, FSP-Patch model, and SimSiam. The mean F1 score and CI of 5 few-shot models tested on the LC dataset with 5-shot are shown with curves.}%
    \label{fig:learning_curve}%
\end{figure}
To validate the adequacy of training for all models, we assess the intermediate checkpoints of each pretraining model on the LC dataset. The learning curves and confidence intervals (CI) of FACILE-FSP, FSP-Patch, and SimSiam are displayed in \figref{fig:learning_curve}.
%\YC{check consistency. In the main paper, we used \\figref{} (first letter in small case) consistently through the paper. We are using Table X (capitalizing first letter) throughout the paper}. 
Upon careful examination of the learning curves in \figref{fig:learning_curve}, we observe conclusive evidence of complete training for all models, as they have reached convergence.

\subsection{Input Set Size}
To examine the impact of input set size on downstream tasks, we conduct pretraining experiments using FACILE-FSP on the TCGA dataset with varying input set sizes. The resulting feature map $e$ from the trained FACILE-FSP is then evaluated on LC, PAIP, and NCT datasets with shot numbers 1 and 5. The corresponding performances are reported in \tabref{tab:ablation_set_size_lc_paip_nct}.

Observing \tabref{tab:ablation_set_size_lc_paip_nct}, we find that models with an input set size of 5 consistently demonstrate superior performance for LC and PAIP datasets. While slight improvements are observed for larger input set sizes, they are not substantial. Conversely, for the NCT dataset, as presented in \tabref{tab:ablation_set_size_lc_paip_nct}, the best performance is attained when the input set size is 10.

\begin{table}[ht]
    \centering
        \scalebox{0.9}{
    \begin{tabular}{c|c|c|c|c}
        \hline
        set size & 2 & 5 & 10 & 15 \\
        \hline
        \multicolumn{4}{c}{1-shot 5-way test on LC dataset}\\
        \hline
        NC  & $75.29 \pm 1.33$ & $\mathbf{77.84 \pm 1.16}$ & $74.88 \pm 1.36$ & $75.25 \pm 1.29$\\
        LR   & $73.72 \pm 1.33$ & $\mathbf{77.56 \pm 1.16}$ & $73.84 \pm 1.29$ & $74.00 \pm 1.27$\\
        RC  & $74.10 \pm 1.34$ & $\mathbf{77.56 \pm 1.17}$ & $73.42 \pm 1.31$ & $73.42 \pm 1.29$\\
        LR+LA  & $75.27 \pm 1.28$ & $\mathbf{79.16 \pm 1.09}$ & $74.41 \pm 1.31$ & $74.92 \pm 1.26$ \\
        RC+LA  & $74.36 \pm 1.33$ & $\mathbf{77.38 \pm 1.18}$ & $72.60 \pm 1.34$ & $73.16 \pm 1.32$ \\
        \hline
        \multicolumn{4}{c}{5-shot 5-way test on LC dataset}\\
        \hline
        NC  & $90.62 \pm 0.56$ & $90.35 \pm 0.50$ & $90.62 \pm 0.57$ & $\mathbf{90.83 \pm 0.55}$ \\
        LR  & $89.41 \pm 0.63$ & $\mathbf{90.91 \pm 0.47}$ & $89.80 \pm 0.59$ & $89.63 \pm 0.60$ \\
        RC &  $89.11 \pm 0.63$ &  $\mathbf{91.54 \pm 0.46}$ & $89.26 \pm 0.61$ & $89.25 \pm 0.60$\\
        LR+LA & $90.46 \pm 0.58$  & $\mathbf{91.68 \pm 0.50}$ & $90.29 \pm 0.57$ & $90.46 \pm 0.56$ \\
        RC+LA & $89.64 \pm 0.63$  & $\mathbf{90.97 \pm 0.54}$ & $88.52 \pm 0.66$ & $89.00 \pm 0.64$ \\
        \hline
                \hline
        NC  & $48.95 \pm 1.24$ & $52.04 \pm 1.25$ & $51.72 \pm 1.22$ & $\mathbf{52.46 \pm 1.20}$ \\
        LR  & $50.55 \pm 1.22$ & $53.27 \pm 1.25$ & $52.33 \pm 1.25$ & $\mathbf{53.38 \pm 1.23}$  \\
        RC   & $50.14 \pm 1.25$  & $\mathbf{54.19 \pm 1.26}$ & $53.04 \pm 1.24$ & $52.68 \pm 1.25$\\
        LR+LA   & $50.12 \pm 1.22$ & $52.66 \pm 1.25$ & $52.96 \pm 1.21$ & $\mathbf{53.41 \pm 1.21}$ \\ 
        RC+LA    & $49.91 \pm 1.22$ & $\mathbf{52.79 \pm 1.23}$ & $51.67 \pm 1.17$ & $51.51 \pm 1.20$ \\ 
        \hline
        \multicolumn{4}{c}{5-shot 3-way test on PAIP dataset}\\
        \hline
        NC    & $66.99 \pm 0.93$ & $\mathbf{69.42 \pm 0.85}$ & $69.10 \pm 0.91$ & $69.08 \pm 0.87$ \\ 
        LR    & $68.11 \pm 0.94$ & $69.93 \pm 0.92$ & $\mathbf{70.30 \pm 0.90}$ & $69.28 \pm 0.90$\\
        RC     & $68.63 \pm 0.91$ & $\mathbf{70.52 \pm 0.87}$ & $70.45 \pm 0.87$ & $70.12 \pm 0.90$\\
        LR+LA   & $69.03 \pm 0.83$ & $69.96 \pm 0.84$ & $\mathbf{70.25 \pm 0.81}$ & $70.00 \pm 0.81$\\
        RC+LA   & $67.32 \pm 0.83$ & $\mathbf{68.39 \pm 0.84}$ & $68.35 \pm 0.83$ & $67.70 \pm 0.81$ \\
        \hline
                \hline
        NC   & $66.31 \pm 1.36$ & $68.21 \pm 1.30$ & $\mathbf{72.44 \pm 1.25}$ & $72.05 \pm 1.27$\\
        LR    & $68.55 \pm 1.32$ & $68.17 \pm 1.31$ & $\mathbf{72.62 \pm 1.25}$ & $72.14 \pm 1.27$\\
        RC    & $68.58 \pm 1.32$ & $68.69 \pm 1.30$ & $\mathbf{72.60 \pm 1.25}$ & $72.04 \pm 1.27$ \\
        LR+LA    & $67.42 \pm 1.33$ & $69.24 \pm 1.28$ & $\mathbf{72.18 \pm 1.26}$ & $71.92 \pm 1.27$ \\
        RC+LA    & $65.87 \pm 1.36$ & $68.18 \pm 1.33$ & $\mathbf{69.98 \pm 1.31}$ & $69.88 \pm 1.28$ \\
        \hline
        \multicolumn{4}{c}{5-shot 9-way test on NCT dataset}\\
        \hline
        NC    & $85.28 \pm 0.72$ & $85.18 \pm 0.60$ & $\mathbf{88.25 \pm 0.56}$ & $88.22 \pm 0.57$ \\
        LR    & $86.39 \pm 0.69$ & $85.87 \pm 0.60$ & $\mathbf{88.80 \pm 0.55}$ & $88.55 \pm 0.55$ \\
        RC   & $87.03 \pm 0.66$ & $87.11 \pm 0.56$ & $\mathbf{89.25 \pm 0.52}$ & $89.02 \pm 0.54$ \\
        LR+LA   & $86.85 \pm 0.65$ & $87.06 \pm 0.61$ & $88.52 \pm 0.59$ & $\mathbf{88.93 \pm 0.55}$ \\
        RC+LA   & $85.60 \pm 0.70$ & $85.81 \pm 0.66$ & $87.40 \pm 0.63$ & $\mathbf{87.74 \pm 0.59}$ \\
        \hline
    \end{tabular}
    }
    \caption{Abation on set size; models tested on LC, PAIP, and NCT dataset; average F1 and CI are reported.}
    \label{tab:ablation_set_size_lc_paip_nct}
\end{table}

\section{Contrastive and Non-contrastive Learning Models}
\label{contrastive_non_contrastive}
Self-supervised learning achieves promising results on multiple visual tasks \citep{bachman2019learning,he2020momentum,chen2020simple,grill2020bootstrap,caron2020unsupervised,chen2021exploring}. Contrastive learning method avoid collapse by encouraging the representations to be far apart for views from different images. \citet{henaff2020data,he2020momentum,misra2020self,chen2020simple} implemented instance discrimination, in which a pair of augmented views from the same image are positive and others are negative. \citet{caron2020unsupervised, caron2018deep} contrasted different cluster of positives. Non-contrastive models \citep{grill2020bootstrap,richemond2020byol,chen2021exploring} removed the reliance on negatives. These non-contrastive models achieved strong results in the ImageNet \citep{imagenet_cvpr09} pretraining setting. SimSiam \citep{chen2021exploring} works with typical batches and does not rely on large-batch training, which makes it preferable for academics and practitioners with low computation resources.

In this section, some contrastive learning and non-contrastive learning models, e.g., SimCLR, SupCon, and SimSiam,  that are used in this paper are explained. Details of implementation are provided. There are three main components in SimCLR and SupCon framework. We follow the notation of \citet{khosla2020supervised} in this section to explain SimCLR and SupCon.

\begin{itemize}
    \item Data augmentation $Aug(\cdot)$. For each input sample $x$, the augmentation module generates two random augmented views, i.e., $\tilde{x}\sim Aug(x)$. The augmentation schedules used in this paper are explained in \secref{app:data_aug}. 
    \item Encoder $Enc(\cdot)$. The encoder extracts a representation vector $r=Enc(\tilde{x})$. The pair of augmented views are separately fed to the same encoder and generate a pair of representations. The $r$ is normalized to the unit hypersphere.
    \item Projection head $Proj(\cdot)$. It maps $r$ to a vector $z=Proj(r)$. We instantiate $Proj(\cdot)$ as a multi-layer perceptron (MLP) with a single hidden layer of size 512 and output vector size of 512. We also normalize the output to the unit hypersphere. 
\end{itemize}

For a set of $N$ randomly sampled sample/label pairs, $\{(x_k, y_k)\}_{k=1}^{N}$. The corresponding batch used for training consists of $2N$ pairs, $\{(\tilde{x}_l, \tilde{y}_l)\}_{l=1}^{2N}$, where $\tilde{x}_{2k-1}$ and $\tilde{x}_{2k}$ are two random augmented views of $x_k$ and $\tilde{y}_{2k-1}=\tilde{y}_{2k}=y_k$. 

\subsection{SimCLR}
\begin{figure}
    \centering
    \includegraphics[width=0.9\textwidth]{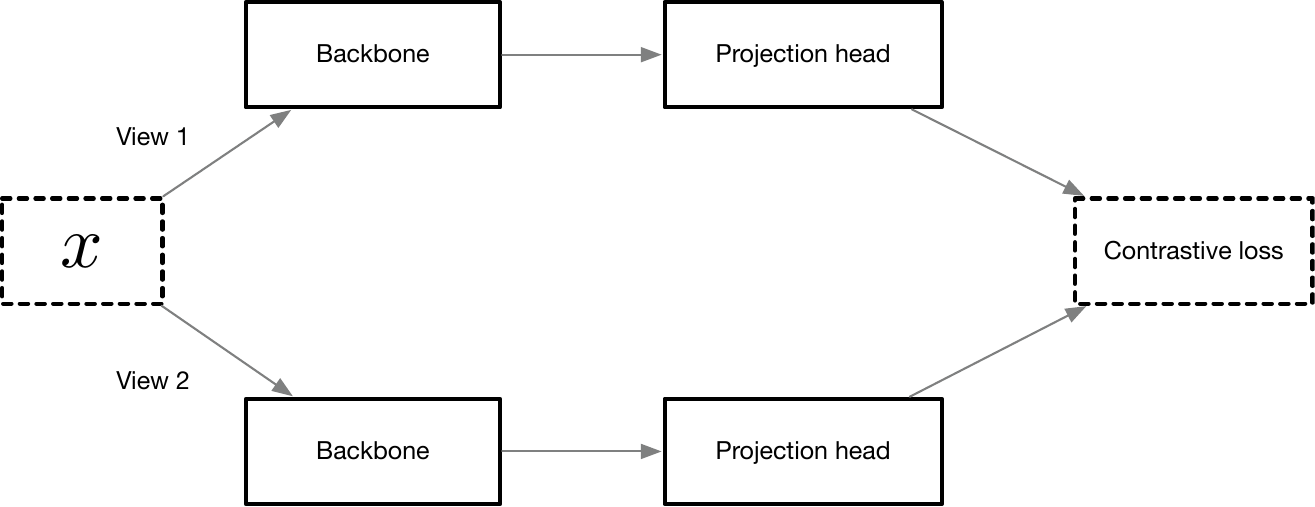}
    \caption{Abstraction of SimCLR structure}
    \label{app_fig:simclr}
\end{figure}

Let $i\in I \equiv\{1 \ldots 2 N\}$ be the index of an arbitrary augmented sample and let $j(i)$ be the index of the other augmented sample originating from the same source sample. The abstraction of SimCLR structure can be found in \figref{app_fig:simclr}. In SimCLR, the loss takes the following form.

\begin{equation}\label{app_eq:contrastive_loss}
    \mathcal{L}^{\text {self }}=\sum_{i \in I} \mathcal{L}_i^{\text {self }}=-\sum_{i \in I} \log \frac{\exp \left(z_i \cdot z_{j(i)} / \tau\right)}{\sum_{a \in A(i)} \exp \left(z_i \cdot z_a / \tau\right)}
\end{equation}

where $\tau$ is the temperature parameter. $A(i)\equiv I \backslash \{i\}$. The denominator has a total of $2N-1$ terms.

In this paper, the $\tau$ is always set to $0.07$. The patches are augmented randomly by the augmentation module described in \secref{app:data_aug}. We use an MLP as a projection head with two fully-connected layers, a hidden dimension of 512, and an output dimension of 512.

\subsection{SupCon}
For supervised learning, the contrastive loss in \eqref{app_eq:contrastive_loss} cannot handle class discrimination \citep{khosla2020supervised}. \citet{khosla2020supervised} proposed two straightforward ways, as shown in \eqref{app_eq:supcon_out} and \eqref{app_eq:supcon_in}, to generalize \eqref{app_eq:contrastive_loss} to incorporate supervison.

\begin{equation} \label{app_eq:supcon_out}
    \mathcal{L}_{\text {out }}^{\text {sup }}=\sum_{i \in I} \mathcal{L}_{\text {out }, i}^{\text {sup}}=\sum_{i \in I} \frac{-1}{|P(i)|} \sum_{p \in P(i)} \log \frac{\exp \left(z_i \cdot z_p / \tau\right)}{\sum_{a \in A(i)} \exp \left(z_i \cdot z_a / \tau\right)}
\end{equation}

\begin{equation} \label{app_eq:supcon_in}
\mathcal{L}_{i n}^{\text {sup }}=\sum_{i \in I} \mathcal{L}_{i n, i}^{\text {sup }}=\sum_{i \in I}-\log \left\{\frac{1}{|P(i)|} \sum_{p \in P(i)} \frac{\exp \left(z_i \cdot z_p / \tau\right)}{\sum_{a \in A(i)} \exp \left(z_i \cdot z_a / \tau\right)}\right\}
\end{equation}

Here $P(i)\equiv \{ p\in A(i):\tilde{y}_p=\tilde{y}_i \}$ is the set of indices of all positives in the batch distinct from $i$. The authors showed that $\mathcal{L}_{\text {in }}^{\text {sup }} \le \mathcal{L}_{\text {out }}^{\text {sup }}$ and $\mathcal{L}_{\text {out }}^{\text {sup }}$ is the superior supervised loss function. Thus, we use SupCon with \eqref{app_eq:supcon_out} as the default loss. The $\tau$ is also set to $0.07$.

In our model FACILE-SupCon, the input sample is a set of randomly sampled patches and labels are slide properties, i.e., organs or TCGA projects. Each patch is augmented randomly by the augmentation module described in \secref{app:data_aug}. The feature map $e$ and set function $g$ work as the encoder $Enc(\cdot)$. We also use an MLP as a projection head with two fully-connected layers, a hidden dimension of 512, and an output dimension of 512.

\subsection{SimSiam}
Non-contrastive models, e.g., SimSiam and BYOL, achieve strong results in typical ImageNet \citep{imagenet_cvpr09} pretraining setting \citep{chen2021exploring,grill2020bootstrap,li2022understanding}. Among the non-contrastive models, SimSiam removes the negatives and uses stop-grad to avoid collapse. Besides, it trains faster, requires less GPU memory, and works well with small batch size \citep{chen2021exploring, li2022understanding}, which makes it extremely appealing to academics.

\begin{figure}
    \centering
    \includegraphics[width=0.96\textwidth]{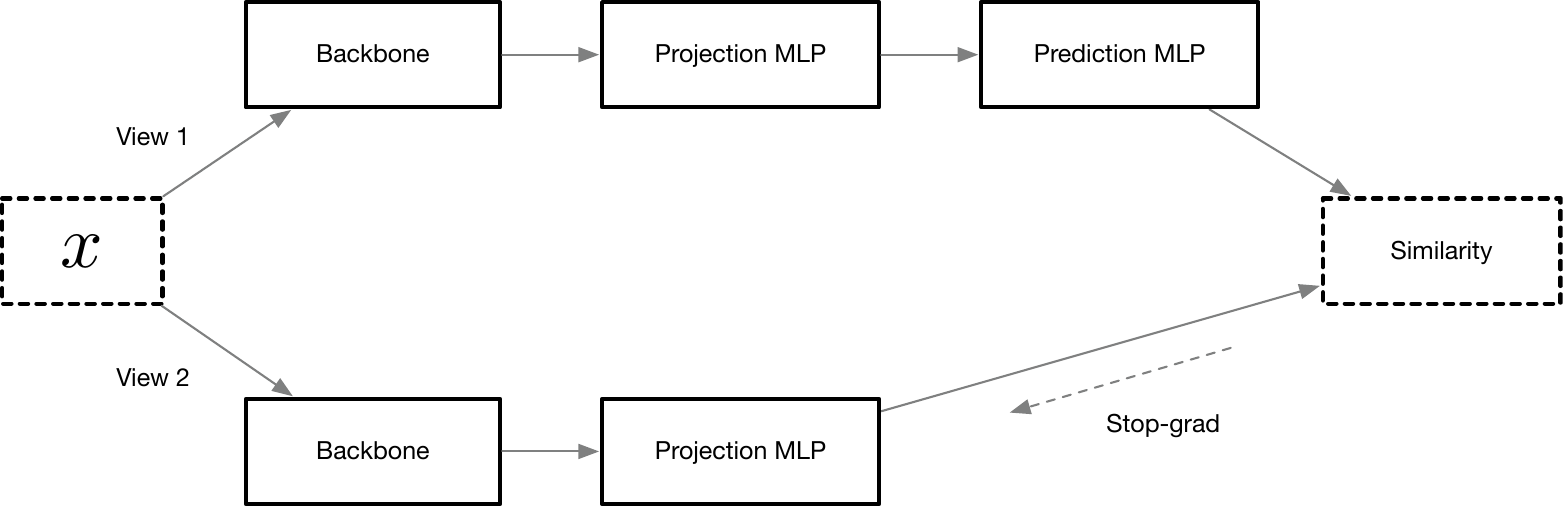}
    \caption{Abstraction of SimSiam structure}
    \label{app_fig:simsiam}
\end{figure}

The abstraction of SimSiam structure is shown in \figref{app_fig:simsiam}. Given two augmented views $\tilde{x}_1$ and $\tilde{x}_2$ of the same image $x$, SimSiam learns to use $\tilde{x}_1$ to predict the representation of $\tilde{x}_2$. Specifically, $\tilde{x}_1$ is passed into the online backbone network on the upper. The $\tilde{x}_2$ is passed into the target backbone network on the lower. The outputs of the two backbone networks are passed to the projection MLPs and then a prediction MLP is used to predict the projected representation of $\tilde{x}_2$ from the projected representation of $\tilde{x}_1$. SimSiam uses the same network for the online and target backbone and projection networks.

In our paper, the projection MLP has 3 fully-connected layers with a hidden dimension of 512 and an output dimension of 512. It has batch normalization (BN) applied to each fully-connected layer including its output fully-connected layer. The prediction MLP also has BN applied to its hidden fully-connected layer. Its output fully-connected layer does not have BN or ReLU. The prediction MLP has 2 layers.

\section{Excess Risk Bound of FACILE}
\label{app:excess_risk_proof}
Our proof framework follows closely the work of \citet{robinson2020strength}. We consider the setting where we have some coarse-grained labels of some sets, rather than instances and the downstream classifiers only use the learned embeddings to train and test on the downstream tasks. /\citet{robinson2020strength} considers a different setting where each instance has a weak label and a strong label, and the strong label predictor learns to predict the strong labels from the instances and their corresponding embeddings learned with weak labels. The diagram of only using trained embeddings for downstream tasks is more often used in self-supervised learning and representation learning for FSL literature \citep{du2020few,yang2021free,bachman2019learning,he2020momentum,chen2020simple,grill2020bootstrap,caron2020unsupervised,chen2021exploring}. The coarse-grained data contains useful information, which is characterized by our defined Lipschitzness, to pretrain a instance feature map that can be leveraged for downstream FSL. We include the full proof of our key result as follows.

In order to prove \thmref{theorem:main}, we first split the excess risk by the following proposition.
\begin{prop}\label{prop:excess_risk_split}
    Suppose that $f^{*}$ is $L$-Lipschitz relative to $\cE$. The excess risk $\sE\left[ \ell^{\mathrm{fg}}_{\hat{f}\circ \hat{e}}(X,Y)-\ell^{\mathrm{fg}}_{f^{*}\circ e^{*}}(X,Y) \right]$ is bounded by,
    \begin{equation*}
        2L\mathrm{Rate}_m(\ellcg, P_{S,W}, \cE) + \mathrm{Rate}_{n}(\ellfg, \hat{P}_{Z,Y}, \cF)
    \end{equation*}
\end{prop}

\paragraph{Proof.} We split the excess risk into three parts
\begin{equation*}
    \begin{split}
       & \sE_{P_{X,Y}}\left[ \ell^{\mathrm{fg}}_{\hat{f}\circ \hat{e}}(X,Y)-\ell^{\mathrm{fg}}_{f^{*}\circ e^{*}}(X,Y) \right]  \\
      =  & \sE_{P_{X,Y}}\left[ \ell^{\mathrm{fg}}_{\hat{f}\circ \hat{e}}(X,Y)-\ell^{\mathrm{fg}}_{f^{*}\circ \hat{e}}(X,Y) \right] +  \sE_{P_{X,Y}}\left[ \ell^{\mathrm{fg}}_{f^{*}\circ \hat{e}}(X,Y) - \ell^{\mathrm{fg}}_{f^{*}\circ e_0}(X,Y) \right] \\
        & \qquad \qquad + \sE_{P_{X,Y}}\left[ \ell^{\mathrm{fg}}_{f^{*}\circ e_0}(X,Y) - \ell^{\mathrm{fg}}_{f^{*}\circ e^{*}}(X,Y) \right]
    \end{split}
\end{equation*}

For the second term and third term, relative Lipschitzness of $f^{*}$ to $\cE$ delivers

\begin{equation*}
\begin{split}
        \sE_{P_{X,Y}}\left[ \ell^{\mathrm{fg}}_{f^{*}\circ \hat{e}}(X,Y)-\ell^{\mathrm{fg}}_{f^{*}\circ e_0}(X, Y)  \right] & =  \sE_{P_{X,Y,S,W}}\left[ \ell^{\mathrm{fg}}_{f^{*}\circ \hat{e}}(X,Y)-\ell^{\mathrm{fg}}_{f^{*}\circ e_0}(X, Y)  \right] \\
        & \le  L \sE_{P_{X,Y,S,W}}\ellcg\left(g_{\hat{e}}\circ \hat{e}(S), g_{e_0}\circ e_0(S)\right)\\
       & =  L\sE_{P_{S,W}}\ellcg\left(g_{\hat{e}}\circ \hat{e}(S), g_{e_0}\circ e_0(S)\right),\\
\end{split}
\end{equation*}

\begin{equation*}
    \begin{split}
    \sE_{P_{X,Y}}\left[ \ell^{\mathrm{fg}}_{f^{*}\circ e_0}(X,Y)-\ell^{\mathrm{fg}}_{f^{*}\circ e^{*}}(X,Y) \right] & =  \sE_{P_{X,Y,S,W}}\left[ \ell^{\mathrm{fg}}_{f^{*}\circ e_0}(X,Y)-\ell^{\mathrm{fg}}_{f^{*}\circ e^{*}}(X,Y) \right]\\
    & \le  L \sE_{P_{X,Y,S,W}} \ellcg\left( g_{e_0}\circ e_0(S),g_{e^{*}}\circ e^{*}(S) \right) \\
    & =  L \sE_{P_{S,W}} \ellcg\left( g_{e_0}\circ e_0(S),g_{e^{*}}\circ e^{*}(S) \right)
\end{split}
\end{equation*}

Since $e^{*}$ attains minimal risk and $W=g_{e_0}\circ e_0(S)$, the sum of the two terms can be bounded by,

\begin{equation*}
    \begin{split}
    & L\sE_{P_{S,W}}\ellcg\left(g_{\hat{e}}\circ \hat{e}(S), g_{e_0}\circ e_0(S)\right) + L \sE_{P_{S,W}} \ellcg \left( g_{e_0}\circ e_0(S),g_{e^{*}}\circ e^{*}(S) \right)\\
    \le & 2L\sE_{P_{S,W}}\ellcg \left( g_{\hat{e}}\circ \hat{e}(S), W  \right) \le 2L\mathrm{Rate}_m(\ellcg, P_{S,W}, \cE)
    \end{split}
\end{equation*}

By combining the bounds on the three terms we can get the claim.

%\paragraph{Relating pretraining and downstream tasks}

The central condition is well-known to yield fast rates for supervised learning \citep{van2015fast}. It directly implies that we could learn a map $Z\rightarrow Y$ with $\widetilde{\mathcal{O}}(1 / n)$ excess risk. The difficulty is that at test time we would need access to latent value $Z=e(X)$. To circumnavigate this hurdle, we replace $e_0$ with $\hat{e}$ and solve the supervised learning problem $(\ellfg,\hat{P}_{Z,Y},\cF)$. 

It is not clear whether this surrogate problem satisfies the central condition. We show that $(\ellfg, \hat{P}_{Z,Y}, \cF)$ indeed satisfies a weak central condition and shows weak central condition still enables strong excess risk guarantees.

Following \citet{robinson2020strength,van2015fast}, we define the central condition on $\cF$.
\begin{defn} \label{defn:central_condition}
(The central condition). A learning problem $(\ellfg, P_{Z,Y}, \cF)$ on $\cZ \times \cY$ is said to satisfy the $\epsilon$-weak $\eta$-central condition if there exists an $f^{*}\in \cF$ such that
\begin{equation*}
    \sE_{(Z,Y)\sim P_{Z,Y}}\left[ e^{\eta(\ell^{\mathrm{fg}}_{f^{*}}(Z,Y)-\ell^{\mathrm{fg}}_f(Z,Y))} \right] \le e^{\eta \epsilon} 
\end{equation*}
for all $f\in \cF$. The $0$-weak central condition is known as the strong central condition.
\end{defn}

\paragraph{Capturing relatedness of pretraining and downstream task with the central condition.} 

Intuitively, the strong central condition requires that the minimal risk model $f^{*}$ attains a higher loss than $f\in \cF$ on a set of $Z, Y$ with an exponentially small probability. This is likely to happen when $Z$ is highly predictive of $Y$ so that the probability of $P(Y|Z)$ concentrates in a single location for most $Z$. If $f^{*}$ in $\cF$ such that $f^{*}(Z)$ maps into this concentration, $\ell^{\mathrm{fg}}_{f^{*}}(Z,Y)$ will be close to zero most of the time.

We assume that the strong central condition holds for the learning problem $(\ellfg, P_{Z,Y}, \cF)$ where $Z=e_0(X)$. Similar to \citet{robinson2020strength}, we split the learning procedure into two supervised tasks as depicted in \algref{facile_alg}. In the algorithm, we replace $(\ellfg, P_{Z,Y}, \cF)$ with $(\ellfg, \hat{P}_{Z,Y}, \cF)$.

We will show that $(\ellfg, \hat{P}_{Z,Y},\cF)$ satisfies the weak central condition.

\begin{prop}\label{prop:surrogate_weak}
Assume that $\ellcg(w, w^{\prime})=\mathbbm{1}\left\{w \neq w^{\prime}\right\}$ and that $\ellfg$ is bounded by $B>0$, $\cF$ is $L$-Lipschitz relative to $\cE$, and that $(\ellfg,P_{Z,Y},\cF)$ satisfies $\epsilon$-weak central condition. Then $(\ellfg,\hat{P}_{Z,Y},\cF)$ satisfies the $\epsilon+\mathcal{O}\left(\frac{\exp(\eta B)}{\eta} \mathrm{ Rate }_m\left(\cE, P_{S, W}\right)\right)$-weak central condition with probability at least $1-\delta$.
\end{prop}

\paragraph{Proof.} Note that 
\begin{equation*}
    \frac{1}{\eta}\log \sE_{\hat{P}_{Z,Y}}\exp\left(\eta(\ell^{\mathrm{fg}}_{f^{*}}-\ell^{\mathrm{fg}}_{f})\right)=\frac{1}{\eta} \log \sE_{P_{X,Y}}\exp\left(\eta (\ell^{\mathrm{fg}}_{f^{*}\circ \hat{e}}-\ell^{\mathrm{fg}}_{f\circ \hat{e}})\right)
\end{equation*}

To prove that $(\ellfg, \hat{P}_{Z,Y}, \cF)$ satisfies the central condition we therefore need to bound $\frac{1}{\eta} \log \sE_{P_{X,Y}}\exp\left(\eta (\ell^{\mathrm{fg}}_{f^{*}\circ \hat{e}}-\ell^{\mathrm{fg}}_{f\circ \hat{e}})\right)$ by some constant. 

\begin{equation*}
    \begin{split}
        & \frac{1}{\eta} \log \sE_{P_{X,Y}}\exp\left(\eta (\ell^{\mathrm{fg}}_{f^{*}\circ \hat{e}}-\ell^{\mathrm{fg}}_{f\circ \hat{e}})\right) \\
        = & \frac{1}{\eta} \log \sE_{P_{X,Y,S,W}}\exp\left(\eta (\ell^{\mathrm{fg}}_{f^{*}\circ \hat{e}}-\ell^{\mathrm{fg}}_{f\circ \hat{e}})\right) \\
        = & \frac{1}{\eta} \log \sE_{P_{X,Y,S,W}}\left[ \exp(\eta (\ell^{\mathrm{fg}}_{f^{*}\circ \hat{e}}-\ell^{\mathrm{fg}}_{f\circ \hat{e}}))\mathbbm{1}\{\hat{g}_{\hat{e}}\circ \hat{e}(S)=W \} \right] + \\
        & \qquad \qquad \frac{1}{\eta} \log \sE_{P_{X,Y,S,W}}\left[ \exp(\eta (\ell^{\mathrm{fg}}_{f^{*}\circ \hat{e}}-\ell^{\mathrm{fg}}_{f\circ \hat{e}}))\mathbbm{1}\{\hat{g}_{\hat{e}}\circ \hat{e}(S)\ne W \} \right]\\
        = & \underbrace{\frac{1}{\eta} \log \sE_{P_{X,Y,S,W}}\left[ \exp(\eta (\ell^{\mathrm{fg}}_{f^{*}\circ e_0}-\ell^{\mathrm{fg}}_{f\circ e_0}))\mathbbm{1}\{\hat{g}_{\hat{e}}\circ \hat{e}(S)=W \} \right]}_{\text{first term}} + \\
        & \qquad \qquad \underbrace{\frac{1}{\eta} \log \sE_{P_{X,Y,S,W}}\left[ \exp(\eta (\ell^{\mathrm{fg}}_{f^{*}\circ \hat{e}}-\ell^{\mathrm{fg}}_{f\circ \hat{e}}))\mathbbm{1}\{\hat{g}_{\hat{e}}\circ \hat{e}(S)\ne W \} \right]}_{\text{second term}}
    \end{split}
\end{equation*}

The third line follows from the fact that for any $f$ in the event $\{\hat{g}_{\hat{e}}\circ \hat{e}(S)=W \}$ we have $\ell^{\mathrm{fg}}_{f\circ \hat{g}}=\ell^{\mathrm{fg}}_{f\circ g_0}$. 

This is because $|\ell^{\mathrm{fg}}_{f\circ \hat{e}}(X,Y)-\ell^{\mathrm{fg}}_{f\circ e_0}(X,Y)|\le L\ellcg(g_{\hat{e}}\circ \hat{e}(S), g_{e_0}\circ e_0(S))=L\ellcg(W,W)=0$.

We get $\frac{1}{\eta} \log \sE_{P_{X,Y,S,W}}\left[ \exp(\eta (\ell^{\mathrm{fg}}_{f^{*}\circ e_0}-\ell^{\mathrm{fg}}_{f\circ e_0}))\right]$ after we drop the $\mathbbm{1}\{\hat{g}_{\hat{e}}\circ \hat{e}(S)=W \}$. It is bounded by $\epsilon$ with the weak central condition. The second term is bounded by 

\begin{equation*}
    \begin{split}
        & \frac{1}{\eta} \log \sE_{P_{X,Y,S,W}}\left[ \exp(\eta (\ell^{\mathrm{fg}}_{f^{*}\circ \hat{e}}-\ell^{\mathrm{fg}}_{f\circ \hat{e}}))\mathbbm{1}\{\hat{g}_{\hat{e}}\circ \hat{e}(S)\ne W \} \right] \\
        \le & \frac{1}{\eta} \log \sE_{P_{X,Y,S,W}}\left[\exp(\eta B)\mathbbm{1}\{\hat{g}_{\hat{e}}\circ \hat{e}(S)\ne W  \} \right] \\
        = & \frac{1}{\eta} \log \sE_{P_{S,W}}\left[\exp(\eta B)\mathbbm{1}\{\hat{g}_{\hat{e}}\circ \hat{e}(S)\ne W  \} \right] \\
        %= & \frac{1}{\eta} \left[ \eta B+ \log \sE_{P} \left[ \mathbbm{1}\{\hat{g}_{\hat{e}}\circ \hat{e}(S)\ne W  \}  \right]  \right] \\
        %\le & B + \frac{1}{\eta}(\mathbb{P}_{P}(\hat{g}_{\hat{e}}\circ \hat{e}(S)\ne W)-1) \\
        %= & \frac{1}{\eta}\mathrm{Rate}_{m}(\ellcg, P_{S,W}, \cE) + B - \frac{1}{\eta}
        < & \frac{1}{\eta} \sE_{P_{S,W}}\left[\exp(\eta B)\mathbbm{1}\{\hat{g}_{\hat{e}}\circ \hat{e}(S)\ne W  \} \right] \\
        = & \frac{\exp(\eta B)}{\eta} P_{S,W}(\hat{g}_{\hat{e}}\circ \hat{e}(S)\ne W  )\\
        = & \frac{\exp(\eta B)}{\eta} \mathrm{Rate}_{m}(\ellcg,P_{S,W}, \cE)
    \end{split}
\end{equation*}

% do we need to say $g_{e*{*}}\circ e*{*}$ need to have zero loss?? 

The first inequality uses the fact that $\ellfg$ is bounded by $B$. The forth line is because that $\log x < x$. By combining this bound with $\epsilon$ bound on the first term we can get the claimed result of \propref{prop:surrogate_weak}. 

% On the other hand, the second term is bounded by
% \begin{equation*}
%     \begin{split}
%      & \frac{1}{\eta} \log \sE_{P_{X,Y,S,W}}\left[ \exp(\eta (\ell^{\mathrm{fg}}_{f^{*}\circ \hat{e}}-\ell^{\mathrm{fg}}_{f\circ \hat{e}}))\mathbbm{1}\{\hat{g}_{\hat{e}}\circ \hat{e}(S)\ne W \} \right] \\
%      \le & \frac{1}{\eta} \log \sE_{P_{X,Y,S,W}}\left[\exp(\eta B)\mathbbm{1}\{\hat{g}_{\hat{e}}\circ \hat{e}(S)\ne W  \} \right] \\
%      = & \frac{1}{\eta} \log \sE_{P_{S,W}}\left[\exp(\eta B)\mathbbm{1}\{\hat{g}_{\hat{e}}\circ \hat{e}(S)\ne W  \} \right] \\
%      = & \frac{1}{\eta}\left[ \log(\exp(\eta B)) + \log(\sE_{P_{S,W}}\mathbbm{1}\{\hat{g}_{\hat{e}}\circ \hat{e}(S)\ne W  \}) \right] \\
%      = & B + \frac{1}{\eta}\log(\mathrm{Rate}_{m}(\ellcg,P_{S,W}, \cE))
%     \end{split}
% \end{equation*}

% \begin{theorem}\label{theorem:main}
% Suppose that $(\ellfg, P, \cF)$ satisfies the central condition and that $\mathrm{Rate}_{m}(\ellcg,P_{S,W}, \cE)=\mathcal{O}\left(1 / m^\alpha\right)$. Then when $\alg_{n}(\ellfg, \hat{P}, \cF)$ is ERM we obtain excess risk $\sE_P\left[ \ell^{\mathrm{fg}}_{\hat{f}\circ \hat{e}}(X, Y) -\ell^{\mathrm{fg}}_{f^{*}\circ e^{*}}(X,Y) \right]$ bound with probability at least $1-\delta$ as follows,
% \begin{equation*}
%     \mathcal{O}\left(\frac{d\alpha \beta \log R L^{\prime}n + \log \frac{1}{\delta}}{n} + \frac{L}{n^{\alpha \beta}}\right)
% \end{equation*}
% if $m=\Omega (n^{\beta})$ and $\ellcg(w, w^{\prime})=\mathbbm{1}\{w\ne w^{\prime}\}$.
% \end{theorem}

%It is proved in Proposition 9 of \citet{robinson2020strength} that

The proof of the main theorem further relies on a proposition provided by \citet{robinson2020strength}, as we show below:
\begin{prop}{\citet{robinson2020strength}}\label{prop:central_condition4fast_rate}
    Suppose $(\ellfg, Q_{Z,Y}, \cF)$ satisfies the $\epsilon$-weak central condition, $\ellfg$ is bounded by $B > 0$, $\cF$ is $L^{\prime}$-Lipschitz in its $d$-dimensional parameters in the $l_2$ norm, $\cF$ is contained in Euclidean ball of radius $R$, and $\cY$ is compact. Then when $\alg_n(\ellfg, Q_{Z,Y}, \cF)$ is ERM, the excess risk $\sE_{Z,Y\sim Q_{Z,Y}}\left[ \ell^{\mathrm{fg}}_{\hat{f}}(Z,Y) - \ell^{\mathrm{fg}}_{f^{*}}(Z,Y) \right]$ is bounded by,
    \begin{equation*}
        \mathcal{O}\left( V\frac{d\log \frac{RL^{\prime}}{\epsilon} + \log \frac{1}{\delta}}{n} + V\epsilon \right)
    \end{equation*}
    with probability at least $1-\delta$, where $V=B+\epsilon$.
\end{prop}

% Q is consistent with former definititon of hat{P}_{Z,Y} and P_{Z,Y}

\paragraph{Proof of the main theorem:} 
If $m=\Omega(n^{\beta})$, the $\mathrm{Rate}_{m}(\ellcg,P_{S,W}, \cE)=\cO (\frac{1}{m^{\alpha}})=\cO(\frac{1}{n^{\alpha \beta}})$. \propref{prop:surrogate_weak} concludes that $(\ellfg, \hat{P}_{Z,Y},\cF)$ satisfies the $\mathcal{O}(\frac{1}{n^{\alpha \beta}})$-weak central condition with probability at least $1-\delta$.
Thus by \propref{prop:central_condition4fast_rate}, we can get $\mathrm{Rate}_n(\ellfg, \hat{P}_{Z,Y}, \cF)=\cO\left(\frac{d \alpha \beta \log R L^{\prime} n+\log \frac{1}{\delta}}{n}+\frac{B}{n^{\alpha \beta}}\right)$.
Combining bounds with \propref{prop:excess_risk_split} we conclude that

\begin{equation*}
\begin{split}
        \sE\left[ \ell^{\mathrm{fg}}_{\hat{f}\circ \hat{e}}(X,Y)-\ell^{\mathrm{fg}}_{f^{*}\circ e^{*}}(X,Y) \right] \le &  2L\mathrm{Rate}_m(\ellcg, P_{S,W}, \cE) + \mathrm{Rate}_{n}(\ellfg, \hat{P}_{Z,Y}, \cF) \\
        \le & \cO\left(\frac{d \alpha \beta \log R L^{\prime} n+\log \frac{1}{\delta}}{n}+\frac{B}{n^{\alpha \beta}} + \frac{2L}{n^{\alpha \beta}}\right)
\end{split}
\end{equation*}

\end{document}